\documentclass[preprint,12pt]{elsarticle}

\usepackage{amssymb}
\usepackage{amsmath,amssymb,amsfonts}
\usepackage{algorithmic}
\usepackage{algorithm}
\usepackage{graphicx}
\usepackage{textcomp}
\usepackage{verbatim}
\usepackage{multirow}
\usepackage{tabu}
\usepackage{makecell}
\usepackage{color}
\usepackage{xcolor}
\usepackage{wrapfig}
\usepackage{hyperref}

\journal{Engineering Applications of Artificial Intelligence}

\begin{document}

\begin{frontmatter}



\title{IoT Data Analytics in Dynamic Environments: From An Automated Machine Learning Perspective}


\author{Li Yang}
\ead{lyang339@uwo.ca}

\author{Abdallah Shami}
\ead{abdallah.shami@uwo.ca}

\address{Department of Electrical and Computer Engineering, University of Western Ontario, 1151 Richmond St, London, Ontario, Canada N6A 3K7}

\date{}

\let\today\relax
\makeatletter
\def\ps@pprintTitle{%
    \let\@oddhead\@empty
    \let\@evenhead\@empty
    \def\@oddfoot{\footnotesize\itshape
         {Preprint published in Engineering Applications of Artificial Intelligence} \hfill\today}%
    \let\@evenfoot\@oddfoot
    }

\begin{abstract}
With the wide spread of sensors and smart devices in recent years, the data generation speed of the Internet of Things (IoT) systems has increased dramatically. In IoT systems, massive volumes of data must be processed, transformed, and analyzed on a frequent basis to enable various IoT services and functionalities. Machine Learning (ML) approaches have shown their capacity for IoT data analytics. However, applying ML models to IoT data analytics tasks still faces many difficulties and challenges, specifically, effective model selection, design/tuning, and updating, which have brought massive demand for experienced data scientists. Additionally, the dynamic nature of IoT data may introduce concept drift issues, causing model performance degradation. To reduce human efforts, Automated Machine Learning (AutoML) has become a popular field that aims to automatically select, construct, tune, and update machine learning models to achieve the best performance on specified tasks. In this paper, we conduct a review of existing methods in the model selection, tuning, and updating procedures in the area of AutoML in order to identify and summarize the optimal solutions for every step of applying ML algorithms to IoT data analytics. To justify our findings and help industrial users and researchers better implement AutoML approaches, a case study of applying AutoML to IoT anomaly detection problems is conducted in this work. Lastly, we discuss and classify the challenges and research directions for this domain.
\end{abstract}


\end{frontmatter}


\section{Introduction}
\footnote{DOI: \url{https://doi.org/10.1016/j.engappai.2022.105366}\\
Code Implementation and General AutoML Tutorials: \url{https://github.com/Western-OC2-Lab/AutoML-Implementation-for-Static-and-Dynamic-Data-Analytics}}By leveraging rapidly evolving communications technologies, the Internet of Things (IoT) systems permit the exchange of meaningful information and knowledge across IoT devices and systems to create value for humans \cite{iot1}. IoT is defined as a network of connected devices and end systems that interact directly to collect, exchange, and analyze critical data through the cloud. Typical IoT applications include smart grids, intelligent vehicles, smart homes, smart agriculture, smart healthcare, and so on \cite{iotr21}. IoT nodes and devices can form many subnets of IoT networks. For example, as a common type of IoT network, a smart city network managed by a municipal government usually consists of many subsets, such as smart homes, smart grids, smart factories, intelligent transportation systems, smart healthcare systems, etc. \cite{iotr21}. 

 IoT is expanding at a rapid speed. According to the Cisco report, about 18.4 billion IoT sensors and devices were connected by 2018, and over 2.5 quintillion bytes of IoT data are generated daily \cite{iotsta1} \cite{iotsta2}.Thus, each real-world IoT device generates an average of approximately 135.9 million bytes of IoT data every day, which is an extremely high volume and generation speed. It is also estimated that more than 30 billion IoT devices would be connected by 2023 \cite{iotsta1}. For example, in intelligent transportation systems, as a critical IoT application, a single connected vehicle generates about 1 gigabyte (GB) of data per second, and Boeing 787 generates about 5 GB of data per second \cite{iotsta3} \cite{iotsta4}.
The massive number of IoT end devices that generate an enormous volume of IoT data on a continuous basis, is posing challenges to IoT systems in terms of providing reliable services and making trustworthy decisions \cite{iot1}. This is because IoT services and functionalities often require fast and accurate data analytics. Effective and efficient data analytics enables IoT systems to make fast decisions, gain rapid insights, discover hidden patterns, and interact with users and other systems \cite{iot4}. 

For IoT data analytics, although human experts can effectively recognize simple data patterns when data dimensionalities are smaller than 3, the dimensions of most real-world IoT data are much larger than 3, making it extremely difficult to analyze data manually \cite{hitl1}. According to comprehensive research \cite{humanmachine} that fairly compares machine analysis and human analysis performance in a research abstract classification study, ML classifiers achieve better classification performance than human classifiers overall. Specifically, most ML classifiers achieve over 0.7 F1-scores, while only 3 out of 63 human classifiers have F1-scores over 0.6. ML models are also robust to changes and variations between the training and test data. Additionally, it requires extensive time and resources for human classifiers to achieve a relatively high F-score. Specifically, the ML classifiers classified 247 abstracts in less than 5 seconds, while the fastest human classifiers spent more than 2 hours classifying them in this study. Additionally, ML models can achieve higher accuracy when trained on larger datasets (e.g., hundreds of thousands of data samples), while it is infeasible for humans to memorize many complex patterns in large datasets. This is primarily due to the limited memory and sensing delays of human brains \cite{humanmachine2}. To summarize, ML models outperform human classifiers in terms of accuracy, time, and reliability \cite{humanmachine}.
Thus, Machine Learning (ML) algorithms have become critical contributors to IoT data analytics, enabling the rapid and accurate processing of massive volumes of data produced by IoT systems to identify patterns required by IoT services \cite{iot2} \cite{Inj_ML}.

Although ML techniques have been widely applied to IoT data analytics applications, deploying ML algorithms often requires intensive domain knowledge and human efforts \cite{auto1}. Therefore, many data analysts and ML researchers have been conducting research on Automated Machine Learning (AutoML) technology, which aims to complete data analytics tasks using ML algorithms with minimal human intervention. 
AutoML techniques are state-of-the-art solutions to automate IoT data analytics processes and reduce human efforts \cite{auto1}. AutoML enables people to save valuable resources, including time, financial, and human resources, by automatically making accurate decisions.

Combined Algorithm Selection and Hyperparameter tuning (CASH) is the essential procedure of general AutoML solutions and data analytics pipelines because the suitable ML algorithms and their hyperparameter configurations have a substantial impact on the data learning performance \cite{auto2}. Other components in AutoML pipelines, like data pre-processing and feature engineering, also significantly affect the outcomes of data analytics, but their automation still faces many challenges and usually requires human intervention. On the other hand, since certain IoT data generated in dynamic IoT environments is dynamic streaming data that changes over time, concept drift issues often occur in IoT data analytics \cite{oasw}. Effective AutoML solutions for IoT dynamic data analytics should also incorporate automated model updating and concept drift-adaptive learning techniques.

To leverage human expertise and knowledge, Human-In-The-Loop (HITL) is introduced to develop ML models by combining machine intelligence with human intelligence \cite{hitl1}. HITL indicates the process that human experts supervise the ML process and help reduce prediction errors. Thus, human experts can still participate in ML pipelines to make creative decisions, while AutoML techniques can help with tedious, repetitive, and laborious data analytics tasks with higher precision and less human effort \cite{hitl1} \cite{auto2}. In the application of AutoML in IoT data analytics, certain procedures in AutoML pipelines, like data sampling and feature extraction, can be interfered with by HITL to deal with the high volumes of IoT data and make creative decisions. Human experts can also help determine the ML model candidates and their hyperparameter ranges for initial model selection and faster hyperparameter tuning to prevent AutoML techniques from doing many unnecessary evaluations, thus enabling more efficient analytics of IoT data with high-volume and high-velocity. Additionally, human experts can help with data analytics result validation and data labeling for model updating and model effectiveness maintenance in high variability data analytics tasks \cite{hitl1}. Despite that certain ML procedures require HITL, most ML procedures can be completely automated by machines to allow human experts to intervene in creative ML procedures.



In this paper, we discuss how AutoML techniques can be applied to IoT data analytics problems to address the existing challenges. To apply ML models to IoT data analytics tasks, there are several major difficulties and challenges \cite{auto1} \cite{auto2}:
\begin{enumerate}
\item It is usually difficult to select the most appropriate ML algorithm among a large number of existing ML models for each specific task.
\item It is usually time-consuming and requires expert knowledge to manually tune ML models to fit each specific task.
\item Many IoT data streams have concept drift issues, causing ML model performance degradation. 
\item The community lacks public and comprehensive benchmarks or practical examples in the AutoML field for IoT data analytics. 
\end{enumerate}

To overcome the above difficulties/challenges, the primary achievements and contributions of this paper are as follows:
\begin{enumerate}
\item Review common ML algorithms, summarizes their pros and cons, and discuss their important hyperparameters for better selection in IoT data analytics.
\item Review existing optimization or  AutoML methods for automated model selection and hyperparameter tuning, summarizes their pros and cons, and recommends the most appropriate method for specific situations.
\item Provide a comprehensive review of automated model updating, a novel  AutoML procedure, by discussing concept drift detection and adaptation methods.   
\item Conduct a comprehensive case study analysis by applying AutoML to practical IoT data analytics tasks; the implementation code is publicly available on GitHub\footnote{
Code for this paper is available at: https://github.com/Western-OC2-Lab/AutoML-Implementation-for-Static-and-Dynamic-Data-Analytics}.
\end{enumerate}

Moreover, this paper makes the following secondary contributions:
\begin{enumerate}
\item Define the overall framework and tasks for IoT data analytics. 
\item Review existing techniques for other important procedures of general AutoML pipelines, including data pre-processing, feature engineering, and performance metric selection.
\item Introduce many existing tools and libraries designed for AutoML and IoT data analytics.
\item Discuss the open challenges and research directions in the field of IoT data analytics and AutoML.
\end{enumerate}

Additionally, as a state-of-the-art research topic, the review on the application of AutoML techniques in IoT data analytics has brought out the following novelties: 
\begin{enumerate}
\item Deploying AutoML techniques at both edge and cloud servers can effectively perform different types of data analytics tasks on IoT data with high volume and high velocity. This is discussed in Section 2. 
\item Different AutoML methods have different suitabilities and performance when applied to specific use cases according to theoretical analysis and practical experiments. This is discussed in Sections 3, 4, and 10. 
\item Concept drift is a distinctive issue in certain IoT data analytics tasks due to the high variability of certain IoT data streams. The occurrence of concept drift will cause ML model performance degradation, which can be addressed by automated model updating and concept drift adaptation techniques. This is discussed in detail in Sections 7 and 10. 

\end{enumerate}

Compared with other earlier review papers \cite{iot1} \cite{auto1} \cite{auto2} \cite{iotdata} \cite{auto4} \cite{cash1} \cite{gra3} on the IoT data analytics or AutoML research topic, this paper has the following differences and improvements:
\begin{enumerate}
\item This paper is the first paper that discusses the AutoML application in the IoT data analytics field. Other existing review or survey papers only discuss either AutoML techniques or IoT data analytics methods. 
\item This paper is the first paper that includes the automated model updating process in the review of AutoML techniques. Other existing review papers ignore automated model updating as they focus on batch/static learning.
\item This paper is the first paper that conducts a comprehensive case study of applying AutoML to practical IoT data analytics tasks with code available. To the best of our knowledge, there is no other paper that publishes a comprehensive sample code for IoT data analytics using multiple AutoML techniques.
\end{enumerate}

Lastly, compared with our previously published review paper in the hyperparameter optimization and AutoML field \cite{hpome}, this paper further discusses the hyperparameters of common ML algorithms and their suitable optimization methods in Sections 3 \& 4. Despite that, there are still many differences and improvements in this paper. In this paper, besides hyperparameter tuning, we have discussed all other procedures of general AutoML pipelines, including data pre-processing, feature engineering, automated model selection, and automated model updating.

This paper is organized as follows: Section 2 presents the properties of IoT time-series data, as well as the layers and tasks of IoT data analytics. Section 3 reviews the ML algorithms that are usually used for IoT data analytics. Section 4 provides an overview of the AutoML pipeline and discusses the optimization methods available for AutoML systems. Sections 5 \& 6 discuss the automated data pre-processing and feature engineering procedures, respectively. Section 7 discusses the automated model updating process by introducing concept drift detection and adaptation methods. Section 8 describes the selection of appropriate performance metrics and validation methods for ML tasks. Section 9 introduces the tools and libraries for AutoML and time-series analytics. Section 10 presents a case study of applying AutoML to IoT data analytics problems and discusses the experimental results. Section 11 discusses the challenges and research directions of AutoML and IoT data analytics. Section 12 concludes the paper.

\section{IoT Data Analytics}

\subsection{IoT Data Characteristics}

Although IoT data is similar to data from many other fields, there are several factors affecting the efficacy of IoT data analytics, such as dynamic IoT environments and time series characteristics. Overall, IoT data has the following characteristics \cite{data2}:

\begin{enumerate}
\item \textbf{High volume}: With the development of large-scale IoT systems, a massive amount of data is continuously generated from a large number of IoT devices. As described in Section 1, real-world IoT devices generate an average of approximately 135.9 million bytes of IoT data daily \cite{iotsta1} \cite{iotsta2}. Both real-time and historical IoT data must be saved and processed to analyze previous patterns and future trends, resulting in a large volume of IoT data requiring analytics. 
\item \textbf{High velocity}: IoT data is often generated at high speed by a large number of IoT devices. To achieve real-time analytics, the processing speed of IoT data should be higher than the generation speed of the data. Thus, efficient data analytics techniques should be developed to process the IoT data with high generation speed.
\item \textbf{High variability}: Due to the dynamic nature of IoT environments, IoT data is often dynamic data with varying distributions over time. For example, when specific events occur, such as the occurrence of Coronavirus Disease 2019 (COVID-19), the generated IoT data would significantly change. 
\item \textbf{Time-series/Temporal correlation}: As IoT devices collect data over time, the IoT data is often created with timestamps or time-related information. Due to environmental factors, IoT data is often time-series data with strong temporal correlations. Thus, time-series analysis is often beneficial for IoT data analytics.
\end{enumerate}

In IoT systems, the majority of data is created in the form of time-series data that has temporal correlations (i.e., the sample collected at time $t$ is related to the data samples collected at previous times [$t-1$ to $t-n$]) \cite{iot3}. A time-series dataset is a collection of measurements or observations collected in chronological order. In time-series data, time is a dependent variable of the target variable. Time-series prediction is the process of predicting future trends using past observations. Global temperature prediction, energy consumption prediction, and IoT device failure detection are typical examples of IoT time-series data analytics tasks.

On the other hand, real-world IoT data is often non-stationary time-series data with varying mean, variance, or autocorrelation \cite{iot3}. Due to the dynamic nature of IoT settings and environments, IoT streaming data is subject to a range of data distribution adjustments. For instance, the physical events observed by IoT sensors may evolve over time, rendering sensing components outdated or necessitating periodic updates. As such, changes in the distribution of IoT data over time, referred to as concept drift, are often inevitable \cite{oasw}.

Concept drift may hamper the decision-making capabilities of IoT data analytics models, which might have a negative impact on IoT systems \cite{oasw}. For example, the misleading decision-making process performed by an IoT anomaly detection model with concept drift issues may significantly impair detection accuracy, leaving the IoT system vulnerable to a range of hostile cyber-attacks. When a concept or distribution in IoT data changes, it should be properly handled. Thus, proper analytics approaches should be used to address concept drift issues in dynamic or online IoT data analytics tasks.

\subsection{IoT Data Analytics Layers}

\begin{figure}
     \centering
     \includegraphics[width=12cm]{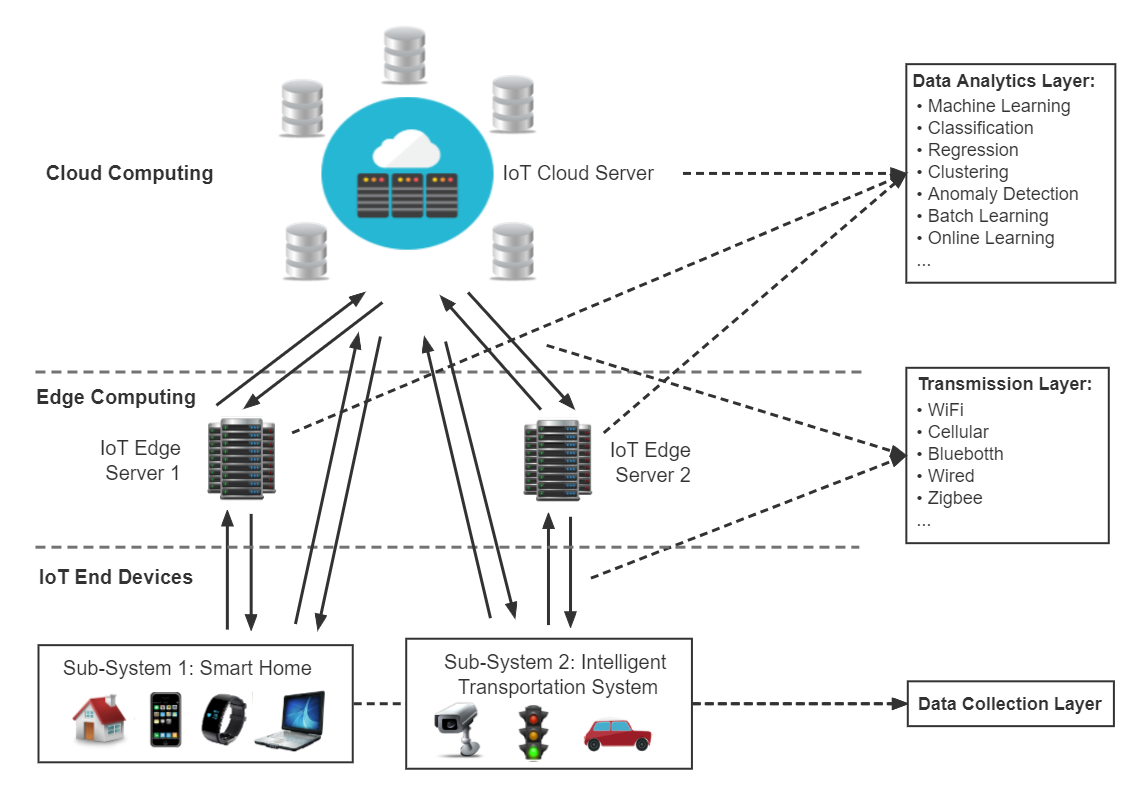}
     \caption{An overview of IoT data analytics architecture.} 
     \label{iot}
\end{figure}

Typical IoT systems consist of three major layers: data collection, transmission, and analytics layers, as shown in Fig. \ref{iot} \cite{data2}. The data collection layer is comprised of IoT end devices that are used to detect, collect, and store sensor data. IoT devices can form sub-systems, like smart homes and Intelligent Transportation Systems (ITSs). IoT end devices and sub-systems are the basic and core components of IoT systems, which directly interact with their physical IoT environments through sensors and actuators. The transmission layer is enabled by gateways to transmit data between IoT end devices and edge/cloud servers. Common transmission strategies include cellular networks, Wireless Fidelity (WiFi), Bluetooth, Zigbee, and so on \cite{oasw}. The analytics layer is responsible for processing and analyzing data from IoT devices, which can be completed in both cloud and edge servers. 

Edge and cloud computing are modern computing and storage paradigms for intelligent IoT data analytics \cite{iot2}. Firstly, IoT data can be processed locally on IoT end devices or edge servers through edge computing. Edge computing enables fundamental IoT data analytics inside the local network of the data source to avoid long-distance transmissions and return real-time processing responses. However, to reduce IoT system costs, IoT end devices are usually constructed with limited computational power and resources. This is because financial costs are critical factors that are considered by IoT manufacturers and designers. Low cost is a core requirement of real-world IoT end devices, and they can collaborate with cloud servers to complete complex tasks \cite{iot1}.
Hence, IoT end devices or edge servers can only perform basic and initial data processing due to their limited resources. It is usually difficult for these resource-constrained IoT devices to perform computational-intensive data analytics tasks. For example, due to the lack of a Graphics Processing Unit (GPU), it is challenging to train Deep Learning (DL) on many IoT end devices; hence, they may have to use simple ML models that may cause under-fitting.

On the other hand, cloud servers can be used to handle large-scale IoT data and perform complex data analytics tasks. Cloud computing is a paradigm in which data is stored, gathered, managed, and processed on remotely placed computing servers connected over the Internet \cite{iot1}. Cloud computing support many services, including Software as a Service (SaaS), Infrastructure as a Service (IaaS), and Platform as a Service (PaaS). As IoT systems can generate a massive amount of data, cloud servers are required for the storage and analytics of large and complex IoT data. One major limitation of cloud computing is the high overhead of processing huge amounts of data, since transmitting IoT data streams between cloud and edge devices would require additional costs, bandwidth, and power. Thus, it is usually difficult to achieve real-time analytics by only using cloud servers.

Thus, collaborative computing, including both edge and cloud computing, is often used in modern IoT systems for large IoT data analytics tasks \cite{edgecloud}. AutoML methods can be operated at both central servers and local edge servers for IoT data analytics applications to improve learning performance and reduce human effort. In IoT systems, any device with computational, communication, and storage capabilities can be used as edge computing devices, such as smart gateways and lightweight base stations. AutoML methods can be implemented in these edge computing devices to perform fundamental and preliminary analytics on the local sensor data. Many preliminary data analytics procedures, such as data pre-processing, feature engineering, storage, and initial analytics, can be conducted at edge devices, which reduces the burden of communication and resource consumption on the cloud servers. 

Although many IoT edge devices enable local processing of IoT data, only basic functionalities are feasible in local processing due to the limited computing capability of many IoT end devices \cite{iot1}. Therefore, large IoT data is often transmitted to the cloud server for comprehensive analytics. AutoML methods can be implemented in central machines with strong computational power on the cloud. After receiving the pre-processed data transferred from the edge devices, the central machines on the cloud servers will comprehensively process and analyze the data using AutoML techniques and return the analytics results to IoT end devices. Although certain IoT systems have a large number of nodes and each node has limited data, it is still necessary to integrate data from multiple data sources or IoT nodes to provide a comprehensive analysis of important events for reliable functionalities and services \cite{iot1} \cite{nodes}. Cloud servers can process data faster and more accurately than edge devices due to their powerful computational capabilities, but they can incur additional latency owing to the data transmission process \cite{edgecloud}. Thus, it is required to strike a balance between the data analytics time and transmission time.

To summarize, cloud computing is suitable for complicated, large-scale, and delay-tolerant data analytics tasks due to its high computational power, while edge computing is suitable for low-latency and real-time data analytics tasks owing to its ability to process data locally. For example, in IoT anomaly detection applications, if multiple edge devices or nodes are under different types of attacks, each edge server can collect and analyze local IoT traffic data using AutoML methods to detect the type of attack that it suffered as a fundamental attack detection, while the cloud server can collect the attack data transmitted from multiple edge servers using AutoML methods to detect various types of attacks as a comprehensive attack detection. The local and fundamental attack detection results can protect each local device from the types of attacks it has suffered, while the comprehensive attack detection results from the cloud server can prevent each local device from both the old and new types of attacks. Deploying AutoML techniques using collaborative computing can provide IoT devices and users with rapid and reliable services.

\subsection{IoT Data Analytics Tasks}
In the analytics layers of the IoT architecture, many IoT data analytics tasks can be completed to provide reliable services and functionalities. IoT data analytics tasks can be classified into four categories: classification, regression, clustering, and anomaly detection \cite{task1}.
\begin{enumerate}
\item \textbf{Classification}: Given a collection of labeled time-series data, the objective of a classification task is to train a classifier capable of assigning correct labels to new time-series data samples. Training on labeled samples enables a classifier to identify distinctive characteristics that can be used to distinguish different classes.
\item \textbf{Regression}: Regression tasks aim to determine the relationship between a series of input features and a continuous target variable. For IoT time-series data analytics, regression tasks are often prediction tasks that aim to predict future trends and measurements using historical data samples with temporal dependencies. 
\item \textbf{Clustering}: Clustering is the process of dividing data samples into a number of groups according to their natural characteristics and patterns. The purpose of clustering is to group similar data samples into the same group and separate the different data samples into different groups. 
\item \textbf{Anomaly detection}: In IoT time-series analytics, anomaly detection is the process of detecting abnormal sequences in a time series to analyze abnormal events. The conventional anomaly detection process is to model normal patterns and then identify sequences that deviate from them.
\end{enumerate}

Moreover, IoT data analytics algorithms can also be classified as batch learning and online learning algorithms, depending on the type of IoT data to be processed \cite{oasw}. 
\begin{enumerate}
\item \textbf{Batch learning}: Batch learning methods analyze static IoT data in batches and often need access to the entire dataset prior to model training. Traditional ML algorithms can effectively solve batch learning tasks \cite{alex}. Although batch learning models often achieve high performance due to their ability to learn diverse data patterns, it is often difficult to update these models once created. Therefore, batch learning faces two significant challenges: model degradation and data unavailability. 
\item \textbf{Online learning}: Online learning techniques are able to train models using continuously incoming online IoT data streams in dynamic IoT environments \cite{online1}. By learning a single data sample at a time, online learning models can reduce memory requirements for data storage and learn new data patterns. Additionally, online learning models can often achieve real-time processing and address concept drift issues. Thus, when applied to dynamic data streams or when inadequate data is available, online learning is often more effective than batch learning.
\end{enumerate}

\section{Model Learning}

ML algorithms have been widely employed in IoT data analytics applications to analyze IoT data and make decisions \cite{ml1}. This Section discusses the commonly used ML algorithms for IoT data analytics in detail. Firstly, it discusses six basic ML algorithms, including K-Nearest Neighbors (KNN), Naïve Bayes (NB), Support Vector Machine (SVM), K-means, Density-Based Spatial Clustering of Applications with Noise (DBSCAN), and Principal Component Analysis (PCA). Secondly, two popular and robust sets of ML algorithms, Decision Tree (DT) based algorithms and Deep Learning (DL) algorithms, are discussed in detail. Tree-based algorithms include DT, Random Forest (RF), eXtreme Gradient Boosting (XGBoost), and Light Gradient Boosting Machine (LightGBM), while DL algorithms include Multi-Layer Perceptron (MLP), Vanilla Recurrent Neural Networks (RNNs), Long Short Term Memory (LSTM) Convolution Neural Networks (CNNs), and Autoencoders (AEs). Lastly, Reinforcement Learning (RL) techniques, as an advanced type of ML algorithms, are discussed.

\subsection{K-Nearest Neighbors (KNN)}
K-Nearest Neighbors (KNN) is a basic ML algorithm that can be used to solve both classification and regression problems \cite{ml1}. KNNs identify the nearest k data points to each test sample in order to estimate its value or category. The average distances between a test sample and its neighbor samples are calculated using a distance metric, such as Euclidean or Mahalanobis distance \cite{ml1}. The majority label or the average observation value of nearby samples will be assigned to each test sample \cite{ml2}. 

The key hyperparameter for KNN is $k$, the number of nearest neighbors, since it has a direct effect on the performance of KNN models \cite{hpome}. To prevent both under-fitting and over-fitting, an appropriate $k$ should be determined. 

While KNN is simple to implement, it often under-performs on complex datasets due to its simplexity \cite{ml2}. Nevertheless, KNN models have been utilized in a variety of IoT applications, such as smart healthcare for disease diagnosis \cite{knna1} and elderly behavior detection \cite{knna2}, as well as Botnet detection in IoT networks \cite{knna3}.

\subsection{Naïve Bayes (NB)}
Naïve Bayes (NB) is a ML algorithm that can be used to solve binary and multi-class classification problems \cite{nb1}. The Bayes Theorem is the core premise of NB; it assumes that there is no statistical relationship between the data points and utilizes the notion of conditional probability to learn data patterns. NB classifiers predict the class label  $\hat{y}$ of each observation $x_{i}$ with n features by \cite{hpome}:
\begin{equation}
\hat{y}=\arg \max _{y} P(y) \prod_{i=1}^{n} P\left(x_{i} | y\right),
\end{equation}
where $P(y)$ is the probability of a target variable $y$, and $P\left(x_{i} | y\right)$ is the posterior probability. 

Traditional NB models do not have any hyperparameter that requires tuning, but multinomial NB \cite{nb2} and complement NB \cite{nb3} are two special types of NB models that add the additive smoothing parameter $\alpha$ to smooth the maximum likelihood. The smoothing parameter $\alpha$ is a continuous hyperparameter that needs to be tuned. 

NB is highly interpretable and computationally efficient. Additionally, when compared to other ML algorithms, the primary benefit of NB is that it does not need a large number of training samples \cite{ml1}. However, a major limitation of NB is that it requires prior knowledge to calculate Bayesian probabilities and make predictions. Additionally, as NB treats all features as independent, certain important correlations and dependencies among different features and observations may be lost \cite{ml3}.

NB is used in a variety of IoT data analytics applications, such as intrusion detection \cite{nba1} and medical diagnosis \cite{nba2}. 

\subsection{Support Vector Machine (SVM)}

SVM is a non-probabilistic supervised learning technique that can be used for classification and regression problems \cite{svmme}. It was initially developed for binary classification problems by establishing a proper hyperplane as the decision boundary that can clearly separate and classify the data. Despite the fact that there are many potential hyperplanes, SVM aims to identify the hyperplane that maximizes the margin between samples of different classes while minimizing the error rate \cite{ml3}. In addition to being able to perform linear classification, SVM can process nonlinear data by identifying a nonlinear hyperplane using kernel functions, which transform the input variable into higher-dimensional feature spaces. A critical hyperparameter of SVM is the kernel function type, including linear, polynomial, Gaussian, and sigmoid kernels. SVM models can also be utilized to solve multi-class classification and regression problems.

SVM is a powerful algorithm that is capable of handling nonlinear and high-dimensional data with effective regularization and generalization. It is also excessively efficient in terms of memory consumption \cite{ml1}. One significant drawback of SVM is that it does not use explicit probability estimations, making it challenging to interpret the model. 

SVM is well-performing in many IoT data analysis applications, including anomaly detection, localization, traffic data classification, and IoT intrusion detection \cite{ml1} \cite{ml2}.

\subsection{Tree-Based Algorithms}
\subsubsection{Decision Tree (DT)}
Decision Tree (DT) \cite{dt1} is a tree-structured ML model that uses a series of if-then conditions to construct a tree and make decisions. A DT is built with multiple decision nodes and leaf nodes that denote a decision test over one of the characteristics and the outcome classes, respectively. The Classification And Regression Trees (CART) model is a common type of DT in which the Gini impurity is used as the split measure function. Assuming $S$ denotes the sub-trees, CART aims to minimize \cite{dt1}:
\begin{equation}
C(S) = {\hat L}_n(S)+{\alpha}|S|,
\end{equation}
where ${\alpha}$ is a constant, $|S|$ is the cardinality of the sub-trees, and ${\hat L}_n(S)$ denotes the empirical risk of using the tree $S$. The complexity of CART mainly depends on the tree depth, a crucial hyperparameter of CART. If the tree depth is too high, over-fitting issues will occur. Overfitting issues can also be addressed via the use of stopping criteria or pruning methods \cite{ml1}.

The primary strengths of DT include its interpretability, capacity for large-scale data processing, and high efficiency \cite{ml1}. However, single DTs' learning performance is often insufficient, as they may get stuck in a local minimum. 

For IoT applications, DTs have shown to be effective in smart citizen behavior classification \cite{ml1} and IoT anomaly detection \cite{mjdt} problems.

\subsubsection{Random Forest (RF)}
Random Forest (RF) \cite{mjrf} is an ensemble learning technique that constructs multiple DTs rather than a single DT to improve prediction accuracy. By integrating multiple decision trees during model learning, RF has become a competent method for both classification and regression tasks. The output of a RF model is the majority class for classification problems, whereas for regression problems, the model output is the average prediction value of base DTs \cite{ml1}. 
RFs are regarded as resilient and reliable ML algorithms capable of handling non-linear and complex datasets. However, they are more prone to overfitting than DTs and have a lower degree of human interpretability. 
RFs have a broad range of applications in IoT smart cities, such as healthcare monitoring systems and intrusion detection systems \cite{ml1} \cite{ml2}.

\subsubsection{XGBoost}
The eXtreme Gradient Boosting (XGBoost) model is a strong ensemble ML model based on DTs \cite{treeme}. In XGBoost, a tree ensemble model with $k$ additive functions is constructed for a data set of $n$ instances and $m$ features. Each function corresponds to an independent tree structure. The decision or regression trees are used, depending on the type of the target variable. For time-series forecasting, the leaves of regression trees are summed up to predict the output. A regularised object is also utilized to learn the functions, which selects a model with simple and predictive functions. Due to the inability of standard approaches to optimize the functions, the model is trained in an additive greedy way. This approach is also known as gradient tree boosting. Apart from the regularized objective, two additional strategies are utilized to minimize overfitting. The first technique is shrinkage, which reduces the influence of each individual tree to make room for new ones. The second technique is feature subsampling, which reduces the impact of noisy features.

XGBoost is a strong tree-based ML model that often achieves high performance in many tasks. Additionally, it works well with high-dimensional data. However, due to the model complexity, XGBoost may encounter over-fitting issues. XGBoost has shown effectiveness in many IoT analytics applications, such as IoT intrusion detection \cite{treeme}, smart home monitoring \cite{xga1}, and human gesture recognition \cite{xga2}. 

\subsubsection{LightGBM}

Light Gradient Boosting Machine (LightGBM) \cite{oasw} is a high-performance tree-based model that is constructed from an ensemble of DTs. In comparison to other ML techniques, the primary strength of LightGBM is its ability to deal with large-scale and high-dimensional data. It is achieved via the use of two strategies: gradient-based one-side sampling (GOSS) and exclusive feature bundling (EFB). GOSS is a downsampling technique that maintains only data samples with high gradients during model training to save time and memory. By combining mutually exclusive characteristics, the EFB approach significantly reduces training time without compromising crucial information. By including GOSS and EFB, LightGBM's time and space complexity has been significantly lowered from $O(NF)$ to $O(N'F')$, where $N$ and $N'$ denote the original and the reduced number of instances, respectively; and $F$ and $F'$ denote the original and bundled number of features, respectively \cite{oasw}.

LightGBM outperforms many other ML approaches in terms of generalizability and robustness \cite{oasw}. Additionally, LightGBM allows multithreading for parallel execution, which significantly increases model efficiency.

Due to its high performance and high efficiency, LightGBM has been applied to many IoT data analytics tasks, such as cyber-attack and malware detection for IoT systems \cite{oasw} \cite{lighta1}.

\subsection{K-means}

In unsupervised learning tasks, there is no ground-truth label for the given input data. Thus, learning models need to discover meaningful patterns within the provided dataset. Clustering algorithms are an important set of unsupervised models that aim to group data samples based on their similarities \cite{ml3}. 

K-means is a common clustering technique that divides an unlabeled dataset into $k$ divisions or clusters depending on the degree of similarity between data points. Typically, the similarity metric is stated in terms of the distance between data points. The objective of k-means is to minimize the sum of squared errors \cite{km}:
\begin{equation}
\sum_{i=0}^{n_{k}} \min _{u_{j} \in C_{k}}\left(\mathbf{x}_{i}-u_{j}\right)^{2},\end{equation}
where $\mathbf{x}_{i}$ denotes the input data samples, $u_j$ represents the centroid of each cluster $C_k$, and $n_k$ denotes the total number of instances in each cluster $C_k$. 

K-means is a scalable, flexible, and efficient unsupervised algorithm. However, it is prone to outliers and ineffective at processing clusters of non-convex shapes.

K-means has been employed in many IoT applications, especially for IoT data that is difficult to label. For example, k-means is used in smart city applications to find suitable living areas \cite{ml3}.

\subsection{DBSCAN}
Density-Based Spatial Clustering of Applications with Noise (DBSCAN) \cite{db1} is another popular clustering algorithm that groups data using the concept of density. In DBSCAN, clusters are defined as dense areas of data points that are separated from low-density regions in the data space. Unlike k-means, which requires configuring the number of clusters, DBSCAN has two hyperparameters to be tuned: the scan radius and the minimum included points. They jointly define the density of clusters \cite{db2}.

DBSCAN is robust to outliers and works well on huge datasets. It is, however, often slower than k-means. 
DBSCAN is a frequently used clustering method that has been used in a variety of real-world IoT applications, such as fraud detection and data labeling \cite{ml1}.

\subsection{PCA}
Principal Component Analysis (PCA) \cite{pca} is a popular unsupervised learning algorithm for dimensionality reduction. PCA is built on the notion of mapping the original $n$-dimensional characteristics to $k$-dimensional orthogonal features $(n>k)$, referred to as the principal components. PCA works by computing the covariance matrix of the data matrix in order to acquire the covariance matrix's eigenvectors. The matrix contains the eigenvectors of the $k$ greatest eigenvalued features (i.e., the largest variance). As a result, the data matrix may be translated into a reduced-dimensional space. Singular Value Decomposition (SVD) is a widely used technique for obtaining the eigenvalues and eigenvectors of a PCA covariance matrix.

PCA has been utilized for feature extraction in a variety of IoT applications, such as IoT anomaly detection \cite{pcaa1} and data fault detection \cite{pcaa2}.

\subsection{Deep Learning (DL) Algorithms}
In recent years, Deep Learning (DL) models have received more attention than traditional ML models, owing to their ability to solve non-linear and difficult data analytics problems \cite{data2}. The use of DL models in IoT data analytics overcomes the limitation of traditional ML models, which are incapable of retaining temporal correlations. DL models are constructed based on Artificial Neural Networks (ANNs) with multiple hidden layers. An ANN is a network of artificial neurons that mimics biological neurons by linking layers of neurons \cite{ml2}. The neurons in ANNs map input data to an appropriate output to predict target values. Many DL models have shown success in IoT applications, including Multi-Layer Perceptrons (MLPs), Recurrent Neural Networks (RNNs), Convolution Neural Networks (CNNs), and AutoEncoders (AEs) \cite{data2}.

\subsubsection{Multi-Layer Perceptron (MLP)}
An ANN consists of an input layer, multiple hidden layers, and an output layer \cite{data2}. The number of hidden layers determines the depth and complexity of a DL model. Each layer is composed of a set of neurons connected through parameterized parameters. ANN models alter the weights of the neuron connections to map the relationship between the input and output. The most fundamental type of ANN is MLP. A MLP is a Feed-Forward Neural Network (FFNN) in which all connections between neurons are forward connections \cite{dl2}. 

MLPs have shown competent performance in many basic classification and regression problems, comparable to that of other well-performing ML algorithms, such as SVM and RF. However, they are often under-performing in time-series data processing since they treat each observation independently and cannot capture the temporal sequence of datasets \cite{dl2}. Therefore, other DL models, such as RNNs and CNNs, have been used more for IoT time-series data analytics, since they can convert time-series tasks into a spatial architecture capable of encoding the time dimension and capturing the dynamical patterns of time-series data. Nevertheless, MLP can still be used in many non-time series data analytics problems, like temperature distribution estimation \cite{r192}.

\subsubsection{Vanilla Recurrent Neural Networks (RNNs)}
RNNs are another set of DL models that can be used to analyze and discover patterns in time-series and sequential data \cite{dl3}. In many real-world applications, only processing individual instances is insufficient, and input sequences must be analyzed to provide reliable predictions \cite{data2}. In contrast to MLPs, which ignore temporal correlations within the data, RNNs link each time step to previous steps in order to capture temporal correlations for time-series analysis. Each neuron in RNNs has a feedback loop that uses the current output as an input for the next time step, allowing the RNN to maintain an internal memory for previous calculations.

However, RNNs suffer from two major issues: exploding gradient and vanishing gradient \cite{dl2}. The exploding gradient problem occurs when the weights start to oscillate, while the vanishing gradient occurs when it takes an excessive amount of time to update the model for long-term pattern learning. Additionally, RNN is designed to tackle short-term correlations and is incapable of capturing long-term dependence accurately \cite{dl4}.

\subsubsection{Long Short Term Memory (LSTM)}
LSTM is an upgraded version of traditional RNNs. It is effective at solving a broad range of problems and is currently extensively utilized for time-series analysis. LSTMs are capable of modeling long-term temporal correlations without discarding short-term trends \cite{dl2}. In contrast to traditional RNNs, LSTMs have memory units that are controlled by a multiplicative input gate, preventing them from being altered by irrelevant disturbances. Additionally, forget gates are included in LSTMs to learn to delete unnecessary memory contents. Through the use of memory units and forget gates, LSTM can overcome the exploding gradient and vanishing gradient issues of traditional RNNs. 

LSTM models often outperform RNN models when input data has a long temporal dependence, but they require a large amount of fixed-sized input data for training. Moreover, similar to other DL models, LSTMs have high computational complexity \cite{stat2}.

RNN and LSTM models have been used in many IoT applications, like weather forecasting \cite{r193}, due to their strong capacity for dealing with time-series data. Liu et al. \cite{rnna1} proposed a water quality monitoring system based on LSTM as an IoT application. The proposed system can effectively process water quality data for time-series prediction. Wu et al. \cite{rnna2} proposed an LSTM and Gaussian Bayes-based model for anomaly detection in industrial IoT systems.

\subsubsection{Convolution Neural Networks (CNNs)}
CNNs are a set of DL models that are initially developed for computer vision applications. CNNs are capable of automatically extracting features from high-dimensional input, such as picture pixels, without the need for additional feature engineering \cite{dl2}. This automatic feature extraction is implemented by performing convolutional operations, which is a sliding filter that generates feature maps to capture repeated characteristics throughout the data. Convolutional processes provide distortion invariance on CNNs, which means that features are retrieved regardless of their locations in the input data. Thus, CNNs are well-suited for sequence data, such as IoT time-series data. One-dimensional CNN (1D-CNN) models are used to process time-series datasets by treating them as one-dimensional images and extracting features with temporal correlations \cite{dl2}.

While CNNs are often more efficient than RNNs due to their local connectivity feature, their prediction performance is often lower than that of RNN models, particularly LSTMs, for time series analysis \cite{dl2}. Another disadvantage of CNNs is that many IoT devices are low-cost and have limited resources, while CNNs require high computational power to achieve high accuracy \cite{dl3}. 

Although CNNs are mainly used for image-related tasks, such as traffic sign detection and visibility estimation in ITS applications \cite{svmme} \cite{thesisme}, as well as fish species identification in smart agriculture \cite{r191}, they can also be used for time-series analysis through data transformation \cite{data2}. Ullah et al. \cite{cnna1} utilized a CNN model to identify malware and infected files on IoT devices by analyzing source code. Roopak et al. \cite{cnna2} proposed a hybrid CNN and LSTM model to detect Distributed Denial-of-Service (DDoS) attacks in IoT systems through IoT data analytics. Yang et al. \cite{cnnme} proposed an optimized CNN and transfer learning-based intrusion detection model in ITS systems to protect autonomous vehicles.

\subsubsection{Autoencoders (AEs)}
AEs are a set of powerful unsupervised DL models for extracting patterns from unlabeled input data \cite{dl3}. AE is primarily composed of an encoder, a decoder, and hidden layers in between \cite{dl4}. The encoder compresses data by translating it to a specified hidden layer, and the decoder reconstructs the approximate input information when the input data needs to be transformed into higher dimensional data. Additionally, the encoder is capable of reconstructing the approximation in relation to the input data. When the input data is non-linear, it is necessary to add more hidden layers to create a more sophisticated AE. After that, the decoder can transform the encoded data into a dataset with fewer dimensions but more meaningful features. 

The primary advantage of AEs for IoT applications is their ability to decrease the dimensionality of input datasets and learn the features of unlabeled datasets \cite{dl4}. Additionally, AE can enhance the security of IoT data by encoding it. As a consequence, AE is an effective IoT data analytics model for security applications. AEs have been utilized to address unsupervised anomaly detection problems in IoT systems \cite{dl3}. Moreover, AEs can be integrated with other DL models to improve prediction performance. Hwang et al. \cite{aea1} proposed an unsupervised DL model for cyber-attack detection in IoT systems by integrating AE and CNN models.  

\subsubsection{DL Conclusion}
DL models have shown promising performance in a variety of IoT data analytics applications due to their strong prediction power and time-series analysis capabilities \cite{r194}. However, DL methods have two serious drawbacks: 1) Large-scale DL models are computationally expensive and energy-intensive; 
2) Designing an optimal DL architecture is time-consuming and labor-intensive. For the first drawback, DL models must strike a balance between prediction accuracy and computational costs, especially in low-cost IoT devices. For the second drawback, AutoML technologies are potential solutions to automate the DL model architecture design process and reduce human efforts \cite{dl3}.

\subsection{Reinforcement Learning}
Reinforcement Learning (RL) techniques have been widely used to solve problems that lack prior knowledge about outputs and inputs \cite{ml2}. Due to the lack of defined outcomes in RL, agents must learn from feedback obtained after interacting with the environment \cite{rl1}. Rewards are given to the agent based on its previous actions to assist it in determining future actions. The agents perform actions and make decisions based on their earned rewards. Through trial and error, agents in RL models can identify the best actions that can gain the highest accumulated reward based on their experience \cite{ml3}. Selecting an appropriate reward function is a critical stage in RL, since it has a direct impact on the learning performance. 

Q-learning is a widely used RL paradigm. The main procedures of Q-learning are as follows \cite{ml2}:
\begin{enumerate}
\item Initialize a Q table.
\item Perform a random action and measure the corresponding reward.
\item Update the Q table using the reward information.
\item Repeat steps 2-3 until the complete Q-table is constructed.
\item Learn the action-value function $Q(S,a)$ based on the Q-table to determine the optimal action $a$ at a state $S$.
\end{enumerate}

RL models have been used in many IoT applications, such as routing protocol design and network performance enhancement \cite{ml2}.  Q-learning strategies have mainly been used in IoT security applications, such as authentication and malware detection \cite{ml3}. Additionally, many IoT devices, such as sensors and smart appliances, have used RL models to adapt to their environments automatically \cite{drl}.

On the other hand, the Deep Reinforcement Learning (DRL) technique, which combines RL with DL models, has also been proposed to process high-dimensional data in non-stationary environments \cite{rl1}. Its objective is to develop self-learning software agents that are capable of establishing effective rules for maximizing long-term benefits. In DRL models, the RL algorithm finds the best policy of actions in an environment based on the output of a DL model. RL’s strong capacity to automatically learn from the environment without requiring feature construction also assists DL in performing effective predictions. In the field of IoT, DRL has been used in semi-supervised learning tasks, such as localization problems in smart campuses \cite{data2}. DL models enable RL models to gain more rewards and produce more accurate predictions.

\subsection{Model Selection Conclusion}
The important hyperparameters, advantages and limitations, as well as the suitable IoT tasks for each ML algorithm, are summarized in Tables \ref{t1} and \ref{t2}. The hyperparameter names in Tables \ref{t1} and \ref{t2} are defined based on the Scikit-learn library \cite{sklearn}.
Different ML algorithms have their own suitabilities for specific IoT tasks. Specifically, KNN, NB, SVM, and DT are suitable for small and simple IoT data analytics tasks, as they are easy to implement and have less over-fitting risks than other complex ML algorithms. Among them, DT has high learning efficiency, so it is suitable for simple IoT data analytics tasks with high-efficiency requirements.

K-means, DBSCAN, and PCA are suitable for unsupervised or unlabeled IoT data analytics tasks, as they are unsupervised learning algorithms and are able to process unlabeled datasets. Among them, K-means is suitable for edge computing and simple IoT tasks due to its low computational complexity, DBSCAN is suitable for cloud computing and complex IoT tasks due to its capacity for processing various types of data distributions, and PCA is mainly used for feature extraction or dimensionality reduction problems. 

Tree-based ensemble ML algorithms, including RF, XGBoost, and LightGBM, are suitable for high-dimensional, complex, and imbalanced IoT data analytics tasks, especially IoT anomaly detection tasks. This is primarily due to their strong prediction power and generalization ability. Among them, LightGBM is more suitable for IoT data analytics tasks with high-efficiency requirements due to its low computational complexity.

\begin{table*}[htbp]
\caption{A comprehensive overview of traditional ML algorithms, their hyperparameters, their advantages and limitations, and suitable IoT tasks.}
\setlength\extrarowheight{1pt}
\centering
\scriptsize
\begin{tabular}{p{1.8cm}|p{2.5cm}|p{4.5cm}|p{3.5cm}}
\Xhline{1.2pt}
\textbf{ML Algorithm}                                                         & \textbf{Main Hyperparameters}                                                                                                                                                                                              & \textbf{Advantages and Limitations}                                                                                                                                                                                                                                                                                                   & \textbf{IoT Task Suitability}                                                                                                               \\ 
\hline
KNN                                                                           & n\_neighbors                                                                                                                                                                                                               & \begin{tabular}[t]{@{}p{4.5cm}}· Easy to implement. \\· Slow on large or high-dimensional datasets. \\· Sensitive to noise.\end{tabular}                                                                                                                                                                                         & Simple IoT data analytics tasks with little noise.                                                                                          \\ 
\hline
NB                                                                            & alpha                                                                                                                                                                                                                      & \begin{tabular}[t]{@{}p{4.5cm}}· Can only be used in classification problems. \\· Treat features independently. \\· Work better with discrete datasets than continuous datasets.\end{tabular}                                                                                                                                   & Classification IoT data analytics tasks with only
  discrete features.                                                                      \\ 
\hline
SVM                                                                           & \begin{tabular}[t]{@{}p{4.5cm}}C, \\kernel\end{tabular}                                                                                                                                                                        & \begin{tabular}[t]{@{}p{4.5cm}}· Can work with non-linear datasets through kernel functions. \\· Robust against noise. \\· Unsuitable for large datasets.\end{tabular}                                                                                                                                                           & Small IoT data analytics tasks.                                                                                                             \\ 
\hline
DT                                                                            & \begin{tabular}[t]{@{}p{4.5cm}}criterion, \\max\_depth, \\min\_samples\_split, \\min\_samples\_leaf, \\max\_features\end{tabular}                                                                                              & \begin{tabular}[t]{@{}p{4.5cm}}· Have good interpretability.  \\· Can handle non-linear data. \\· High efficiency. \\· Single DTs have limited prediction power.\end{tabular}                                                                                                                                                 & Simple IoT data analytics tasks with high-efficiency
  requirements.                                                                        \\ 
\hline
RF                                                                            & \begin{tabular}[t]{@{}p{4.5cm}}n\_estimators \\max\_depth, \\criterion, \\min\_samples\_split, \\min\_samples\_leaf \\max\_features\end{tabular}                                                                               & \begin{tabular}[t]{@{}p{4.5cm}}· Work well with high-dimensional datasets. \\· Can handle imbalanced datasets.\end{tabular}                                                                                                                                                                                                         & High-dimensional, complex, and imbalanced IoT data analytics
  tasks, especially IoT anomaly detection.                                    \\ 
\hline
XGBoost                                                                       & \begin{tabular}[t]{@{}p{4.5cm}}n\_estimators, \\max\_depth, \\learning\_rate, \\subsample, \\colsample\_bytree,\end{tabular}                                                                                                   & \begin{tabular}[t]{@{}p{4.5cm}}· Work well with high-dimensional datasets. \\· Strong prediction power. \\· High computational complexity. \\· May have over-fitting issues.\end{tabular}                                                                                                                                     & High-dimensional, complex, and imbalanced IoT data analytics
  tasks, especially IoT anomaly detection.                                    \\ 
\hline
LightGBM                                                                      & \begin{tabular}[t]{@{}p{4.5cm}}n\_estimators, \\max\_depth, \\learning\_rate, \\num\_leaves\end{tabular}                                                                                                                       & \begin{tabular}[t]{@{}p{4.5cm}}· Work well with high-dimensional datasets. \\· Strong prediction power. \\· Low computational complexity.  \end{tabular}                                                                                                                                                                  & High-dimensional, complex, and imbalanced IoT data analytics
  tasks with high-efficiency requirements, especially IoT anomaly detection.  \\ 
\hline
K-means                                                                       & n\_clusters                                                                                                                                                                                                                & \begin{tabular}[t]{@{}p{4.5cm}}· Easy to implement. \\· Low computational complexity. \\· Only work well with globular-shape data.\end{tabular}                                                                                                                                                                                  & Simple unsupervised IoT data analytics tasks with
  high-efficiency requirements.                                                           \\ 
\hline
DBSCAN                                                                        & \begin{tabular}[t]{@{}p{4.5cm}}eps, \\min\_samples\end{tabular}                                                                                                                                                                & \begin{tabular}[t]{@{}p{4.5cm}}· Suitable for density-based datasets. \\· Work well with more types of data distributions than k-means. \\· Low convergence speed.\end{tabular}                                                                                                                                                  & Complex unsupervised IoT data analytics tasks without
  high-efficiency requirements.                                                       \\ 
\hline
PCA                                                                           & n\_components                                                                                                                                                                                                              & \begin{tabular}[t]{@{}p{4.5cm}}· Effective for feature extraction. \\· May lose important feature patterns.\end{tabular}                                                                                                                                                                                                            & Feature extraction or dimensionality reduction tasks
  for IoT data analytics.                                                              \\ 

\Xhline{1.2pt}
\end{tabular}
\label{t1}%
\end{table*}

\begin{table*}[htbp]
\caption{A comprehensive overview of DL and RL models, their hyperparameters, their advantages and limitations, and suitable IoT tasks.}
\setlength\extrarowheight{1pt}
\centering
\scriptsize
\begin{tabular}{p{1.8cm}|p{2.5cm}|p{4.5cm}|p{3.5cm}}
\Xhline{1.2pt}
\textbf{ML Algorithm}                                                & \textbf{Main Hyperparameters}                                                                                                                                                                                              & \textbf{Advantages and Limitations}                                                                                                                                                                                                                                                                                                   & \textbf{IoT Task Suitability}                                      \\ 
\hline
\begin{tabular}[t]{@{}p{1.8cm}}General Deep Learning (e.g., MLP) \\\end{tabular} & \multirow{5}{*}{\begin{tabular}[t]{@{}p{2.5cm}}number of hidden layers, \\‘units’ per layer, \\loss, \\optimizer, \\activation, \\learning\_rate, \\dropout rate, \\epochs, \\batch\_size, \\early stop patience\end{tabular}} & \begin{tabular}[t]{@{}p{4.5cm}}· Can work with non-linear and complex datasets. \\· Work well with various types of datasets and tasks. \\· Not require feature engineering. \\· Prone to over-fitting. \\· Require a large number of data samples. \\· Require high computational power.\end{tabular}                  & Complex IoT data analytics tasks without high-efficiency requirements, preferably on cloud servers.                                       \\ 
\cline{1-1}\cline{3-4}
RNN                                                                           &                                                                                                                                                                                                                            & \begin{tabular}[t]{@{}p{4.5cm}}· Can learn complex time-series patterns. \\· Have exploding gradient and vanishing gradient problems.\end{tabular}                                                                                                                                                                                  & Complex time-series IoT data analytics tasks without
  high-efficiency requirements, preferably on cloud servers.                           \\ 
\cline{1-1}\cline{3-4}
LSTM                                                                          &                                                                                                                                                                                                                            & \begin{tabular}[t]{@{}p{4.5cm}}· Can learn complex time-series patterns. \\· Can keep long-term dependencies and address gradient and vanishing gradient problems.\end{tabular}                                                                                                                                                     & Complex time-series IoT data analytics tasks without
  high-efficiency requirements, preferably on cloud servers.                           \\ 
\cline{1-1}\cline{3-4}
CNN                                                                           &                                                                                                                                                                                                                            & \begin{tabular}[t]{@{}p{4.5cm}}· Can learn sequence patterns by data transformation. \\· Have many transfer learning models for efficiency and accuracy improvement. \\· Require complex computations for convolution and pooling operations.\end{tabular}                                                                       & Image-based IoT data analytics tasks.                                                                                                       \\ 
\cline{1-1}\cline{3-4}
AE                                                                            &                                                                                                                                                                                                                            & \begin{tabular}[t]{@{}p{4.5cm}}· Can work with unlabeled data. \\· Can preserve only useful patterns and eliminate irrelevant patterns.  \\· Have low complexity. \\· May have a sparse illustration and layer-by-layer errors.\end{tabular}                                                                                    & Complex unsupervised IoT data analytics tasks without
  high-efficiency requirements.                                                       \\ 
\hline
RL                                                                            & \begin{tabular}[t]{@{}p{4.5cm}}number of epochs, \\batch\_size \\learning\_rate, \\decay\_rate, \\gamma,\end{tabular}                                                                                                          & \begin{tabular}[t]{@{}p{4.5cm}}· Suitable when there is no training set available, as learning can be completed through interaction with the environment. \\· Can be used with DL models to construct effective DRL models. \\· Suffer from the curse of dimensionality. \\· Have high computational complexity.\end{tabular} & Complex IoT data analytics tasks without high-efficiency requirements, preferably on cloud servers.                                       \\ 

\Xhline{1.2pt}
\end{tabular}
\label{t2}%
\end{table*}

For DL and RL algorithms, owing to their high computational power, they are suitable for complex IoT data analytics tasks. However, they are usually only suitable for IoT cloud computing tasks due to their high computational complexity and Graphics Processing Unit (GPU) requirements. There are two special types of DL algorithms. RNNs are suitable for time-series problems due to their time-series support, while CNNs are designed for image-based IoT data analytics tasks. 

In conclusion, different ML algorithms have their own advantages and limitations, and should be selected based on specific types of IoT data analytics tasks. For automated model selection, as described in Section 1, human experts can also help determine the optimal ML model by providing an initial list of potential candidate models based on their knowledge and experience, which is a common HITL process. This can significantly reduce the AutoML execution time. On the other hand, using AutoML enables human experts to get rid of manually evaluating, tuning, and selecting from multiple candidate models, as this can be automatically completed by machines.

\section{AutoML Overview \& Optimization Techniques}

\subsection{AutoML Overview}
Although there are many existing ML algorithms that are commonly used in IoT data analytics applications to analyze IoT data and make decisions, as described in Section 3, a randomly-selected ML model with default architecture or hyperparameter configuration usually cannot achieve the optimal analytics results and make accurate decisions. Thus, experienced data scientists are required in many procedures of ML pipelines, including preparing appropriate and clean data, selecting the most suitable ML algorithm, tuning hyperparameters, and determining whether the model needs to be updated. Data scientists often conduct experiments using a variety of ML algorithms and hyperparameter values in order to determine the most efficient combination. These procedures are labor-intensive, time-demanding, and require specialized expertise in ML and data analysis \cite{auto3}. The process of automating this ML design and tuning process is referred to as AutoML. Thus, AutoML refers to the fully automated process of applying machine learning to real-world and practical applications. AutoML can be used by both beginners and experts to apply ML models efficiently. It has the potential to significantly improve the performance and effectiveness of ML models by shortening work cycles, enhancing model performance, and even possibly eliminating the necessity for data scientists. Hence, AutoML is a promising solution to data scientist shortage and high labor costs. In this Section, the basic concept and common optimization techniques of AutoML technology are discussed to automatically optimize the ML learning models introduced in Section 3.

There are three significant benefits of using AutoML:
\begin{enumerate}
\item It increases efficiency and reduces computational costs by automating repetitive ML processes. 
\item It assists in avoiding mistakes caused by human labor.
\item It lowers the threshold of implementing ML models by requiring less ML expertise and experience. 
\end{enumerate}

Moreover, manually selecting, designing, and tuning ML algorithms for IoT data analytics is usually more time-consuming than AutoML \cite{manual}. The objective of AutoML is to enable the automation of the tedious and time-consuming ML model selecting, designing, and tuning procedures by machines instead of humans. Compared with manually crafted ML models, AutoML-based models can achieve higher performance in terms of both accuracy and efficiency/execution speed. This has been stated and proved in many scientific papers, such as \cite{eff1} - \cite{eff3}. Thus, using AutoML techniques will not increase the IoT data analytics time compared with using traditional ML models, as they automate and simplify the selecting, designing, and tuning procedures of traditional data analytics models.

\begin{figure}
     \centering
     \includegraphics[width=7.3cm]{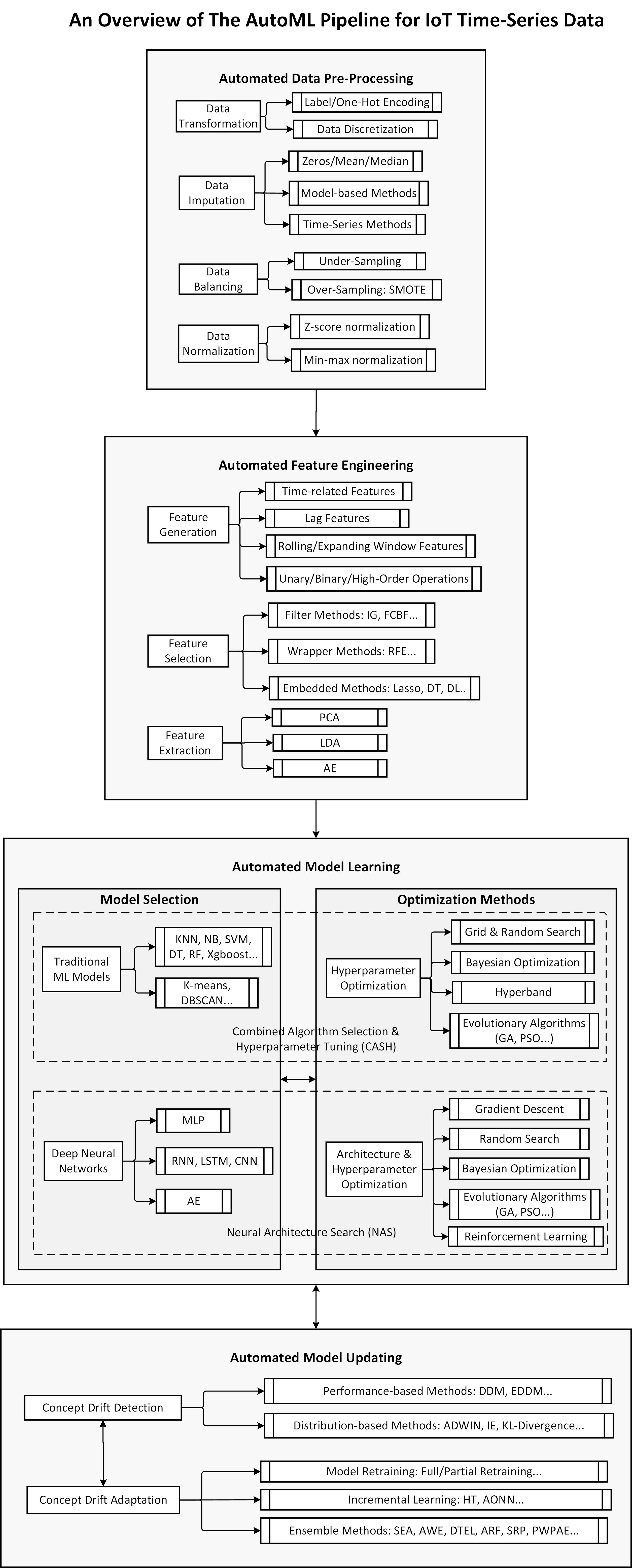}
     \caption{The overview of an AutoML pipeline for IoT data analytics. } 
     \label{overview}
\end{figure}

An overview of the AutoML pipeline for IoT time-series data analytics is illustrated in Fig. \ref{overview}. It consists of four stages: automated data pre-processing, automated feature engineering, automated model learning, and automated model updating \cite{auto2}. Automated model learning can be further divided into automated model selection and Hyper-Parameter Optimization (HPO). AutoML begins with data pre-processing, which aims to transform the original data into a sanitized version. It is a time-consuming and important procedure that has a significant effect on learning performance. The next step is feature engineering, which includes feature extraction and selection. This stage preserves important patterns of datasets while enhancing learning generalization. The following step is model selection, which uses an optimization technique to identify the optimal ML algorithm that produces the most accurate predictions. The process of tuning hyperparameters, referred to as hyper-parameter optimization, aims to further enhance the model learning performance. AutoML systems often need to use a range of optimization algorithms to carry out these phases in the pipeline. This Section discusses the process of automated model learning using optimization techniques. The following subsections and Sections 5-8 will explain the remaining steps of the AutoML pipeline in depth.

\subsection{Combined Algorithm Selection and Hyperparameter Optimization \\(CASH)}
To apply ML algorithms to real-world problems, the primary challenge is selecting and configuring ML. In a prediction task, the accuracy of different models with different configurations might vary significantly \cite{auto4}. Therefore, it is critical to determine the most appropriate ML model with the optimal hyperparameter configuration.

Hyperparameters are the parameters of ML algorithms that determine the architecture of ML models and must be specified prior to model learning \cite{hpome}. Hyperparameters can be classified into three types based on their domains: continuous hyperparameters (e.g., the learning rate of neural networks), discrete hyperparameters (e.g., the number of clusters in k-means), and categorical hyperparameters (e.g., the kernel type in support vector machines). Additionally, certain hyperparameter configurations have conditionality. For example, the two important hyperparameters in DBSCAN, the scan radius and the minimum included points, have strong correlations to determine the data density together \cite{hpome}. Conditional hyperparameters must be tuned together in order to identify the optimal configuration.

In ML model learning, selecting appropriate ML algorithms and hyperparameter values can be seen as a search problem. All the potential ML models and their hyperparameter combinations define a search space, and a single model with a hyperparameter configuration can be seen as a point in the search space. Detecting the optimal point in the search space is a global optimization problem.

Therefore, using optimization techniques to automatically detect the best ML algorithm with the optimal hyperparameter configuration is defined as a Combined Algorithm Selection and Hyperparameter (CASH) optimization problem \cite{weka}. CASH is a core component of current AutoML systems. CASH systems are divided into two stages: model selection and HPO. At the first stage, suitable ML models are selected with their default hyperparameters. At the second stage, model-specific hyperparameters are tuned to obtain the optimal final model \cite{auto4}.

In general, the CASH problem is defined as finding the ML algorithm and hyperparameter configuration that minimizes the loss function. It can be described as follows \cite{weka}:
 \begin{equation}
A^{\star}, \boldsymbol{\lambda}_{\star} \in \underset{A(j) \in \mathcal{A}, \boldsymbol{\lambda} \in \Lambda^{(j)}}{\operatorname{argmin}} \frac{1}{K} \sum_{i=1}^{K} \mathcal{L}\left(A_{\boldsymbol{\lambda}}^{(j)}, D_{\text {train}}^{(i)}, D_{\text {valid}}^{(i)}\right)
\end{equation}
where $A^{\star} \in A$ is the algorithm to be chosen, $\lambda$ is the hyperparameters of the algorithms to be tuned, $D_{train}$ and $D_{valid}$ denote the training and validation sets, $K$ denotes k-fold cross-validation.  

To summarize, CASH is the process of applying optimization methods to select ML models and tune their hyperparameters in order to achieve optimal or near-optimal performance based on defined metrics within a certain time budget \cite{cash1}. The existing optimization techniques for CASH problems is discussed in Section 4.4.

\subsection{Neural Architecture Search (NAS)}
Over the past several years, DL models have achieved remarkable progress on a variety of tasks, including IoT data analytics. Due to the success of many DL models, their architecture design has become a popular research topic. To ensure that a DL model performs well on a particular task, its network architecture must be well designed \cite{nas1}. Given that manually designing neural architectures requires deep domain knowledge and a significant amount of time, it is crucial to automate this process, named Neural Architecture Search (NAS) \cite{auto1}. With NAS, ML engineers can avoid the time-consuming process of designing neural architectures. 

NAS is the process of automating the design of DL architectures. NAS approaches are designed to automatically discover the best architecture and related hyperparameters for a given task. Given a search space for a neural architecture $S$, an input dataset $D$ segmented into $D_{train}$ and $D_{val}$, and a cost function $C$, the NAS technique aims to identify the optimal neural network $f$ with the lowest cost on the dataset $D$ \cite{nas2}:
\begin{equation}
\left\{\begin{array}{c}
f^{*}=\operatorname{argmin} \operatorname{Cost}\left(f\left(\theta^{*}\right), D_{v a l}\right), f \in F \\
\theta^{*}=\operatorname{argmin} L\left(f\left(\theta^{*}\right), D_{t r a i n}\right), \theta
\end{array}\right.
\end{equation}
where $Cost$ denotes the evaluation metric, such as accuracy or mean squared error, and $\theta$ denotes the hyperparameters. The search space $F$ encompasses all possible neural architectures derivable from the initial structures. 

The computational complexity of NAS is often described as $O(nt)$, where $n$ is the number of architecture designs to be evaluated, and $t$ denotes the average evaluation time for each architecture. Additionally, since DL models are trained using gradient-based optimization methods, their training process is computationally expensive. Hence, an effective NAS algorithm should return a high-performing architecture without increasing the task's complexity.

\subsection{Optimization Methods}
In this subsection, existing optimization methods for solving CASH and NAS problems are discussed, including Grid Search (GS), Random Search (RS), Bayesian Optimization (BO), gradient-based algorithms, Hyperband, Genetic Algorithm (GA), and Particle Swarm Optimization (PSO).

\subsubsection{Grid Search (GS)}
Grid search (GS) is a fundamental HPO method that uses a brute-force search strategy to detect the optimal CASH or NAS configurations \cite{cash1}. GS explores all candidate models and hyperparameter settings exhaustively and identifies the one with the best performance. The time complexity of GS is $O(n^k)$, where $k$ is the number of CASH configurations, and $n$ denotes the number of distinct values for each configuration \cite{gst}. 

GS is easy to implement and can benefit from parallelization.  However, GS is computationally expensive and subject to the curse of dimensionality, since the number of evaluations rises exponentially with the number of hyperparameters \cite{auto4}. 

\subsubsection{Random Search (RS)}
To reduce inefficient exhaustive evaluations, Random Search (RS) \cite{rs} was proposed. RS randomly selects configurations from the given CASH or NAS search space until a specified budget is exhausted \cite{cash1}. Thus, when a limited budget is given, RS often outperforms GS. 

RS is often much more efficient than GS through random searches, particularly when the search space is large. This is because RS can significantly reduce the amount of time spent on configurations that are unimportant. On the other hand, RS is capable of discovering the global or near-global optimum when given sufficient resources. RS and GS, on the other hand, also suffer from the curse of dimensionality, as the time complexity of RS is also $O(n^k)$ \cite{hpome}. Additionally, there are still a significant number of unnecessary evaluations when using RS since it does not determine the next evaluations based on prior experience. Hence, much effort still needs to be spent evaluating underperforming configurations. 

\subsubsection{Bayesian Optimization (BO)}
To overcome the limitation of GS and RS in that they do many unnecessary evaluations, BO \cite{bo} models have been developed for AutoML problems. BO is a state-of-the-art CASH approach that is well-suited for cost-sensitive objective functions. BO is composed of two primary components: surrogate models for modeling the objective function and an acquisition function for quantifying the value produced by the objective function's assessment at a new point \cite{cash1}. BO models aim to strike a balance between exploration and exploitation, with exploration referring to the process of traversing previously undiscovered regions, while exploitation to the process of examining samples in the present region where the global optimum is most likely to occur \cite{hpome}.

The Gaussian Process (GP) and the Tree Parzen Estimator (TPE) are common models used as BO surrogate models \cite{gp} \cite{tpe}. BO models can be classified into BO-GP and BO-TPE models based on the surrogate model utilized \cite{mjbo}.

GP has become a popular surrogate model for BO, as they can model non-convex functions effectively. Predictions in a GP follow Gaussian distributions \cite{bo2}: 
\begin{equation}
p(y | x, D)=N\left(y | \hat{\mu}, \hat{\sigma}^{2}\right),
\end{equation}
where $D$ is the CASH configuration search space, and $y=f(x)$ is the prediction for input data $x$, modeled by a function $f$ with a mean of $\mu$ and a covariance of $\sigma^2$. 

After evaluating each new data point, the current GP will be updated based on new evaluation results. Then, the next points to evaluate are chosen according to the confidence intervals produced by the GP. This process can be repeated to identify the global optimum.

One of the primary drawbacks of BO-GP is its cubic computational complexity, $O(n^3)$, which restricts its parallelizability. Moreover, GP was initially designed to process continuous variables, so it is often less effective for other types of variables.

Tree-structured Parzen Estimator (TPE) \cite{tpe} is another popular BO surrogate model. Instead of defining a predictive distribution for the objective function, BO-TPE generates two density functions, $l(x)$ and $g(x)$, that serve as generative models for all processed observations. In BO-TPE, the input data is partitioned into two sets (good and bad observations) based on a predefined threshold $^{*}$, which is modeled using basic Parzen windows:
\begin{equation}
p(x | y, D)=\left\{\begin{array}{ll}{l(x),} & {{ if \quad} y<y^{*}} \\ {g(x),} & { { if \quad} y>y^{*}}\end{array}\right..
\end{equation}

The ratio of the two density functions represents the anticipated improvement in the acquisition function and is used to determine new potential hyperparameter configurations. BO-TPE has shown superior performance when used in a variety of AutoML tasks, owing to its ability to optimize complicated CASH configurations with low computational complexity of $O(nlogn)$ \cite{cash1} \cite{hpome}. Additionally, TPE is capable of properly handling conditional variables, since it makes use of a tree structure to maintain conditional dependencies.

BO approaches are effective for AutoML applications, due to their ability to handle stochastic, non-convex, and non-continuous functions. Moreover, BO models are more efficient than GS and RS since they select future configurations based on the outcomes of previous evaluations. One limitation of BO models is that they are often difficult to parallelize, as they use a sequential process to identify optimums. Nevertheless, they can often discover well-performing solutions within a few iterations.

\subsubsection{Gradient-based Algorithms}
Gradient descent \cite{gra} is a conventional optimization method that utilizes gradient descent of variables to find the optimum direction and identify minimum values. Gradient descent begins at a random point and progresses in the opposite direction of the largest gradient to the next point until convergence occurs, signifying the detection of a local optimum. The local optimum is also the global optimum for convex functions. The time complexity of gradient descent is $O(n)$ \cite{gra2}. 

Gradient-based methods are faster than many other optimization methods for local minimum identification since they have a fast convergence speed for continuous variables. However, the local optimums detected by gradient-based methods are often not the global optimums for non-convex functions \cite{gra3}. As most real-world applications and ML models are non-convex, gradient-based methods often under-perform in AutoML problems.

\subsubsection{Hyperband}
High computational cost is a major issue for CASH problems, especially with large datasets \cite{hpome}. Multi-fidelity optimization is a technique for reducing the computational cost of evaluations using inexpensive low-fidelity evaluations and costly high-fidelity evaluations. While high-fidelity evaluations provide reliable answers throughout the full dataset, low-fidelity evaluations focus on small-sized subsets to reduce evaluation time. Using multi-fidelity optimization can greatly reduce the evaluation cost while still detecting the optimal solution \cite{cash1}. 

Hyperband \cite{hyperband} is a powerful multi-fidelity and bandit-based method that improves search space by choosing from randomly sampled configurations. The objective of Hyperband is to strike a balance between the evaluation cost and accuracy. Firstly, it divides a budget $B$ according to the number of configurations n, so that the budget allocated to each configuration is $b=B/n$. After that, for each configuration, the successive halving method is used for configuration selection. Successive halving is a multi-fidelity method that eliminates half of the possible configurations in each iteration until the ideal solution is found. Thus, by using successive halving, many poor-performing configurations can be eliminated, significantly reducing execution time. Hyperband has low computational complexity of $O(nlogn)$ \cite{hpome}. Hyperband outperforms several other optimization approaches, such as GS and RS, in AutoML applications, due to their high efficiency \cite{cash1}.

\subsubsection{Genetic Algorithm (GA)}
GA is a popular population-based algorithm that is based on the evolutionary theory that individuals with better survival capacity and adaptability are more likely to survive and pass on their genes to the next generations \cite{auto1}. To use GA in AutoML problems, each individual represents a CASH configuration, and the population includes all possible configurations in the CASH search space. Additionally, a fitness function is used as the evaluation metric of CASH configurations \cite{ga1}.

The fundamental concept behind GA is to apply numerous genetic operations on a population of configurations to identify optimal solutions \cite{cash1}. Selection, crossover, and mutation are three important operations in GA \cite{auto2}. Selection is the process of selecting a subset of the population with the objective of maintaining high-performing individuals while eliminating the poor ones. After selection, each pair of individuals is chosen to create a new child that receives half of each parent's genes, called crossover. Mutation randomly selects a chromosome and alters one or more of its genes or parameters, resulting in a completely new chromosome \cite{auto2}. Through these three operations, the best individual can eventually survive and be identified as the optimal solution. 

GA is easy to implement and can handle complex CASH configurations. Additionally, GA can usually identify global optimums through random initialization, so it does not require much ML expertise. However, GA itself has several hyperparameters to be set, such as the population size, the type of fitness function, and the crossover and mutation rates. The time complexity of GA is $O(n^2)$, which is higher than many other methods, like hyperband and BO-TPE  \cite{ga2}. Moreover, GA is often difficult to parallelize due to its sequential nature. 

\subsubsection{Particle Swarm Optimization (PSO)}
PSO \cite{pso1} is an optimization technique inspired by bird and fish social behaviors. PSO aims to detect the optimal solution by fostering collaboration and information exchange among members of a population. 

PSO begins with a population of randomly produced candidates or particles. Each particle's fitness value or score is computed using an optimal fitness function in each generation. The population is updated at each iteration by migrating toward the particles with the highest performance and searching for their neighbors. Each particle has two properties: velocity, which indicates its speed, and position, which indicates its direction of movement. Each particle separately searches the search space for the optimal solution and stores it as its current individual optimum. Additionally, information about the optimal individual optimum is shared with other particles in the whole particle swarm in order to identify the optimal individual optimum, which results in the global optimum. Each particle in the particle group adjusts its speed and location in line with its current individual optimal solution and the particle swarm's current global optimum. 

It is easier to implement PSO than GA, as PSO does not have additional hyperparameters to tune. PSO has a low time complexity of $O(nlogn)$ \cite{pso2}. Additionally, PSO can be easily parallelized, as individuals in PSO can perform actions independently of one another and simply need to share information. However, PSO requires proper population initialization, as it has a direct impact on the performance of PSO; otherwise, PSO may get trapped in local optimums rather than the global optimum.

\subsubsection{Optimization Method Conclusion}
The strengths and limitations of the optimization and CASH methods involved in this paper are summarized in Table \ref{t3}. GS and RS are simple to implement, but they are computationally expensive because they do not consider previous results, which causes many unnecessary evaluations \cite{hpome}. BO-GP and gradient-based models are suitable for ML algorithms with continuous hyperparameters, like NB with the hyperparameter alpha. However, they are unable to optimize other types of hyperparameters, like categorical hyperparameters. Hyperband is a fast optimization method with parallel execution support, but it requires the randomly-generated subsets of data to be representative because it evaluates ML models on those subsets instead of the original large dataset to improve efficiency. Thus, it is unsuitable for certain IoT datasets, like time-series datasets whose subsets may lose time correlations. GA is a robust optimization method, but is often slow due to its poor capacity for parallelization. Lastly, BO-TPE and PSO are two robust optimizations that are efficient with all types of hyperparameters. However, BO-TPE's capacity for parallelization is limited, and PSO requires good initialization to identify the global optimum. Nevertheless, BO-TPE and PSO are still the two most recommended 
optimization methods due to their strong capacity. Taking their limitations into account, BO-TPE is more suitable for IoT tasks or systems with high-performance requirements, and PSO is more suitable for IoT tasks or systems with high-efficiency requirements due to its parallelization ability \cite{hpome}.

\begin{table*}[!ht]
\caption{The comparison of common optimization methods for CASH and HPO problems.}
\setlength\extrarowheight{1pt}
\centering
\scriptsize
\begin{tabular}{p{1.35cm}|p{4.25cm}|p{4.4cm}|p{1.25cm}}
\Xhline{1.2pt}

\hline
\textbf{AutoML Method} & \textbf{Strengths }                                                                                            & \textbf{Limitations }                                                                                                                       & \textbf{Time Complexity}  \\ \Xhline{1.2pt}
\hline
GS                           & ·         Simple.                                                                                              & \begin{tabular}[c]{@{}p{4.4cm}}· Computationally expensive.\\ · Only efficient with categorical hyperparameters. \end{tabular}                                         & $O(n^k)$                        \\ 
\hline
RS                           & \begin{tabular}[c]{@{}p{4.25cm}}· Faster than GS.\\ · Support parallel execution.\end{tabular}                  & \begin{tabular}[c]{@{}p{4.4cm}}· Not consider previous results.\\ · Not efficient with conditional hyperparameters. \end{tabular}                           & $O(n)$                          \\ 
\hline
BO-GP                      &  \begin{tabular}[c]{@{}p{4.25cm}}·         Fast convergence speed for continuous hyperparameters.     \end{tabular}                                                         & \begin{tabular}[c]{@{}p{4.4cm}}· Not efficient with conditional HPs.\end{tabular}                         & $O(n^{3})$                      \\ 
\hline
BO-TPE                       & \begin{tabular}[c]{@{}p{4.25cm}}· Efficient with all types of hyperparameters.\\ · Keep conditional dependencies. \end{tabular}  & ·         Limited
  capacity for parallelization.                                                                                                & $O(nlogn)$                       \\ 
\hline
Gradient-based models        & \multicolumn{1}{l|}{\multirow{2}{*}{ \begin{tabular}[c]{@{}p{4.25cm}}·         Fast convergence speed for continuous hyperparameters.     \end{tabular} }}                                                 & \multicolumn{1}{l|}{\multirow{2}{*}{\begin{tabular}[c]{@{}p{4.4cm}}·  Only work with continuous continuous hyperparameters.\\ · May only detect local optimums. \end{tabular}    }}                             & \multicolumn{1}{l}{\multirow{2}{*}{$O(n^{k})$ }}                       \\ 
\hline
Hyperband                    & ·         Support parallel execution.                                                                              & \begin{tabular}[c]{@{}p{4.4cm}}· Not efficient with conditional hyperparameters.\\ · Need subsets with small budgets to be representative. \end{tabular} & $O(nlogn)$                       \\ 
\hline
GA                           & \begin{tabular}[c]{@{}p{4.25cm}}· Efficient with all types of hyperparameters.\\ · Not need good initialization.\end{tabular} & ·         Poor capacity for parallelization.                                                                                                & $O(n^{2})$                      \\ 
\hline
PSO                          & \begin{tabular}[c]{@{}p{4.25cm}}· Efficient with all types of hyperparameters.\\ · Enable parallel execution.\end{tabular}          & · Need proper initialization.                                                                                                            & $O(nlogn)$                       \\
\Xhline{1.2pt}

\end{tabular}
\label{t3}%
\end{table*}

\section{Data Pre-Processing}

\subsection{Overview}
Data pre-processing aims to improve the quality of data for ML model development. Common data quality issues include outliers, missing values, and class imbalance \cite{auto4}. Data pre-processing procedures guarantee that ML models can learn meaningful patterns from the quality data, but they are time-consuming and tedious. Therefore, Automated Data Pre-processing (AutoDP) is a critical component of AutoML \cite{auto4}.

Data pre-processing tasks can be divided into the following four categories \cite{dp1}:
\begin{enumerate}
\item \textbf{Transformation}: The process of transforming categorical features into continuous features using encoding techniques, or transforming continuous features into categorical features using discretization techniques.
\item \textbf{Imputation}: The process of handling missing values using imputation methods. 
\item \textbf{Balancing}: The process of balancing a dataset's class distribution through over-sampling or under-sampling methods.
\item \textbf{Normalization}: The process of converting continuous characteristics to a comparable or same range of values.
\end{enumerate}

\subsection{Data Transformation}
Data transformation indicates the transformation between numerical features and categorical features. Firstly, in real-world applications, many data values are generated as words or strings to make them human-readable. Data encoding is the process of converting string features to numerical features that machine learning models can understand and process \cite{auto4}. Common encoding techniques include label encoding, one-hot encoding, and target encoding. For better interpretation in ML models, label coding and one-hot encoding assign incremental values or a new column to each string value of categorical features, respectively. However, the transformed values only represent a unique label instead of containing meaning information \cite{transform}. Target encoding is to replace categorical values with the mean or median of the target variable. Target encoding can generate meaningful values, like the fraction of the samples in different classes, to replace string values \cite{auto4}.

Many AutoML tools have data transformation functionalities. For example, Auto-Sklearn \cite{auto-sklearn} uses one-hot encoding, and H2O.AI \cite{h2o} uses target encoding to encode data \cite{auto4}.

On the other hand, data discretization is the process of converting numerical features to categorical features by setting multiple intervals \cite{discretization}. Data discretization can better handle outliers and simplify the calculations. 

\subsection{Data Imputation}
Real-world datasets often have missing values as a result of data inaccessibility or collecting difficulties. Null values, whitespace, NaNs, and incorrect data types are all examples of missing values.  Most ML models are incapable of directly handling missing values or are adversely affected by them in the learning process \cite{auto4}.

While dropping the related features or observations with missing values is the easiest solution, it may result in the loss of significant information. As a result, missing values are often resolved using imputation techniques. The purpose of data imputation is to replace missing information with reasonable values. 
Several basic imputation methods replace missing values with the same value. For numerical features, basic imputation methods replace all missing values in this column with zero, the mean, or the median value of each column \cite{auto4}. For categorical features, the basic method is called mode imputation, which involves replacing missing values with the most common category in each feature.

However, since the sequential values in IoT time-series data often have strong correlations, basic imputation methods, such as zero, mean, and median imputation methods, are often ineffective in dealing with missing values of IoT time-series data \cite{miss1}. Thus, many advanced imputation methods for time-series data have been proposed, such as backward/forward filling and the moving window \cite{miss2}. Backward and forward filling methods replace each missing value with its most recent or next observation, respectively. As a result, they enable missing values to be distributed according to time series distributions. However, imputed values may be misleading for sudden changes in observations. The moving window is another time-series imputation method that replaces each missing value with the average of its previous n observations, indicating a moving window with size $n$. The primary difficulty with the moving window is determining the window size $n$. 

Model-based imputation techniques, like linear regression, KNN, and XGBoost imputation, can estimate the missing values as the target variable by learning other feature values \cite{auto4}. XGboost imputation method is used in several AutoML tools, such as Auto-WEKA \cite{weka} and TPOT \cite{tpot}. Additionally, Datawig \cite{datawig}, a DL-based method, is developed for data imputation. Model-based imputation techniques often outperform model-free methods as imputed values estimated by ML models are often closer to actual values. However, implementing machine learning models often takes much longer than other methods. Moreover, DL-based methods also require high computational power and a relatively large-sized dataset for accurate imputation. The pros and cons of common imputation methods are summarized in Table \ref{t4}.

\begin{table*}[htbp]
\caption{The comparison of common imputation methods.}
\setlength\extrarowheight{1pt}
\centering
\scriptsize
\begin{tabular}{p{1.7cm}|p{2.3cm}|p{4cm}|p{4cm}}
\Xhline{1.2pt}
\textbf{Type}                          & \textbf{Imputation Method}                                       & \textbf{Pros}                                                                                                 & \textbf{Cons}                                                                                                                               \\ 
\hline
\multirow{4}{1.7cm}{Model-free
  methods}  & Dropping                                                         & ·
  Easy to
  implement                                                                               & ·
  May lose
  important information                                                                                                \\ 
\cline{2-4}
                                       & Basic imputation methods (zero, mode, mean, median
  imputation) & \begin{tabular}[c]{@{}p{4cm}}· Easy to implement \\· Work well with small datasets\end{tabular} & \begin{tabular}[c]{@{}p{4cm}}· May generate misleading values \\· Not consider feature correlations\end{tabular}              \\ 
\cline{2-4}
                                       & Forward/ backward filling                                        & ·
  Work well with
  time-series datasets                                                             & ·
  Not
  effective for consecutive nulls and sudden changes                                                                        \\ 
\cline{2-4}
                                       & Moving window                                                    & ·
  Work well on
  time-series datasets                                                               & \begin{tabular}[c]{@{}p{4cm}}· Not effective for a large number of nulls  \\· Need to determine the window size\end{tabular}  \\ 
\hline
\multirow{2}{1.7cm}{Model-based
  methods} & Model-based imputation (linear regression, KNN,
  XGBoost, etc.) & ·
  Perform
  better than basic methods                                                               & \begin{tabular}[c]{@{}p{4cm}}· Time-consuming on large datasets \\· Need to tune hyperparameters\end{tabular}                 \\ 
\cline{2-4}
                                       & Datawig (DL imputation)                                         & ·
  Perform
  better than basic methods                                                               & \begin{tabular}[c]{@{}p{4cm}}· Require high computational power \\· Not work well with small datasets\end{tabular}            \\
\hline

\Xhline{1.2pt}
\end{tabular}
\label{t4}%
\end{table*}

The main procedures for automating the imputation process are as follows:
\begin{enumerate}
\item Calculate the total number of missing values and their percentage in a given dataset to evaluate whether data imputation is necessary;
\item Select a suitable imputation method to handle missing data according to the requirements of specific tasks. If execution speed is the top priority, model-free imputation methods are the most efficient choices. If the model performance is more important than the execution speed, model-based imputation methods can provide more accurate imputed data. On the other hand, if the given dataset is a time-series dataset, time-series imputation methods (e.g., forward/backward filling and moving window methods) are better choices. 
\item Optimize the parameters of imputation methods if it is necessary (e.g., the window size in the moving window method and the hyperparameters of ML algorithms in model-based imputation).
\end{enumerate}

\subsection{Data Balancing}
With the growth of IoT data streams, it is becoming more difficult to maintain uniform distributions of all classes for classification problems, resulting in class imbalance. Class imbalance indicates that the distributions of classes in a dataset are highly imbalanced, causing ML model degradation. Severe class imbalance occurs when certain classes have an extremely small number of instances.  Many ML algorithms, including SVM, DT-based algorithms, and neural network models, are very sensitive to class imbalance \cite{alex}. Learning imbalanced datasets often causes unjustified bias in majority classes, which has an adverse effect on the prediction accuracy of minority classes \cite{auto4}. Class imbalance problems can be solved by resampling techniques, including over-sampling and under-sampling \cite{sample1}.
\subsubsection{Under-Sampling Methods}
Under-sampling methods solve class imbalance by reducing the number of samples in the majority classes. Random Under-Sampling (RUS) is a basic under-sampling method for data balancing by randomly discarding samples from the majority classes \cite{sample2}. By reducing the size of the data, under-sampling techniques can increase model learning efficiency. However, by removing a fraction of data samples, critical information contained in the majority classes may be lost \cite{sample2}. 
\subsubsection{Over-Sampling Methods}
Since under-sampling algorithms may ignore certain critical instances in the majority class, resulting in model performance degradation \cite{sample3}, over-sampling methods are often utilized to resolve class imbalance. Random Over Sampling (ROS) and Synthetic Minority Oversampling TEchnique (SMOTE) \cite{smote} are the two common over-sampling methods used to create new instances in the minority classes. Unlike ROS, which simply replicates the instances, SMOTE analyzes the original instances and synthesizes new instances using the principle of KNN. For each instance X in the minority class, assuming its $k$ nearest neighbors are $X_1,X_2,\cdots,X_k$, and $X_i$ is a randomly selected sample from the $k$ nearest neighbors, the new synthetic instance is denoted by \cite{smote2},
\begin{equation}
X_{n}=X+rand(0,1) *\left(X_{i}-X\right), i=1,2, \cdots, k,
\end{equation}
where $rand(0,1)$ denotes a random number in the range of 0 to 1.

As SMOTE can solve the majority of class imbalance problems, it can be used as the default method on imbalanced datasets \cite{mth}.

\subsection{Data Normalization}
ML models often treat features with larger values as more important. If the scales of features are significantly different, data normalization should be used to prevent creating biased models. This is especially important for ML algorithms that use distance calculations, such as k-means, KNN, PCA, and SVM. Z-score and min-max normalization are two of the most often used normalization approaches in ML model learning. 

In Z-score normalization, the normalized value of each data point, $x_n$, is denoted by \cite{norm},
\begin{equation}
x_{n}=\frac{x-\mu}{\sigma},
\end{equation}
where $x$ is the original value, $\mu$ and $\sigma$ are the mean and standard deviation of the data points.

In min-max normalization, the normalized value of each data point, $x_n$, is denoted by \cite{norm},
\begin{equation}
x_{n}=\frac{x-min}{max-min},
\end{equation}
where $x$ is the original feature value, $min$ and $max$ are the minimum and maximum values of each original feature.

Min-max normalization ensures that all features have the same scale of 0-1, but it does not handle outliers well. In contrast, Z-score normalization can handle outliers, although the feature ranges may be slightly different. Thus, these two techniques can be automatically chosen depending on the occurrence and the percentage of outliers.

\section{Feature Engineering}

\subsection{Overview}
Although research into the automated model selection and HPO has made significant progress, Feature Engineering (FE), as a vital component of the ML pipeline, has been ignored in many AutoML applications. Using original feature values to train ML models often cannot obtain the best prediction results. In this case, new features need to be created, and misleading features should be removed to fit specific tasks \cite{auto4}. The objective of feature engineering is to provide data with optimal input features for ML models. The upper limit of ML applications is determined by FE \cite{auto2}.

Manual feature engineering is tedious and time-consuming, and often requires domain knowledge. Automated Feature Engineering (AutoFE) enables the automation of feature engineering by automatically generating and selecting relevant features using a generic framework applicable to different problems. AutoFE is more efficient and reproducible than manual feature engineering, enabling faster development of more accurate learning models. 

FE methods can be classified into three categories: feature generation, feature selection, and feature extraction. Feature generation is the process of creating new features through the combination or transformation of original features to expand the feature spaces. Feature selection is used to reduce feature redundancy by selecting relevant and significant features. Similar to feature generation, feature extraction can create new features, but its primary purpose is to reduce the dimensionality of original features through mapping functions. AutoFE is essentially a dynamic combination of these three components. 

\begin{figure}
     \centering
     \includegraphics[width=10cm]{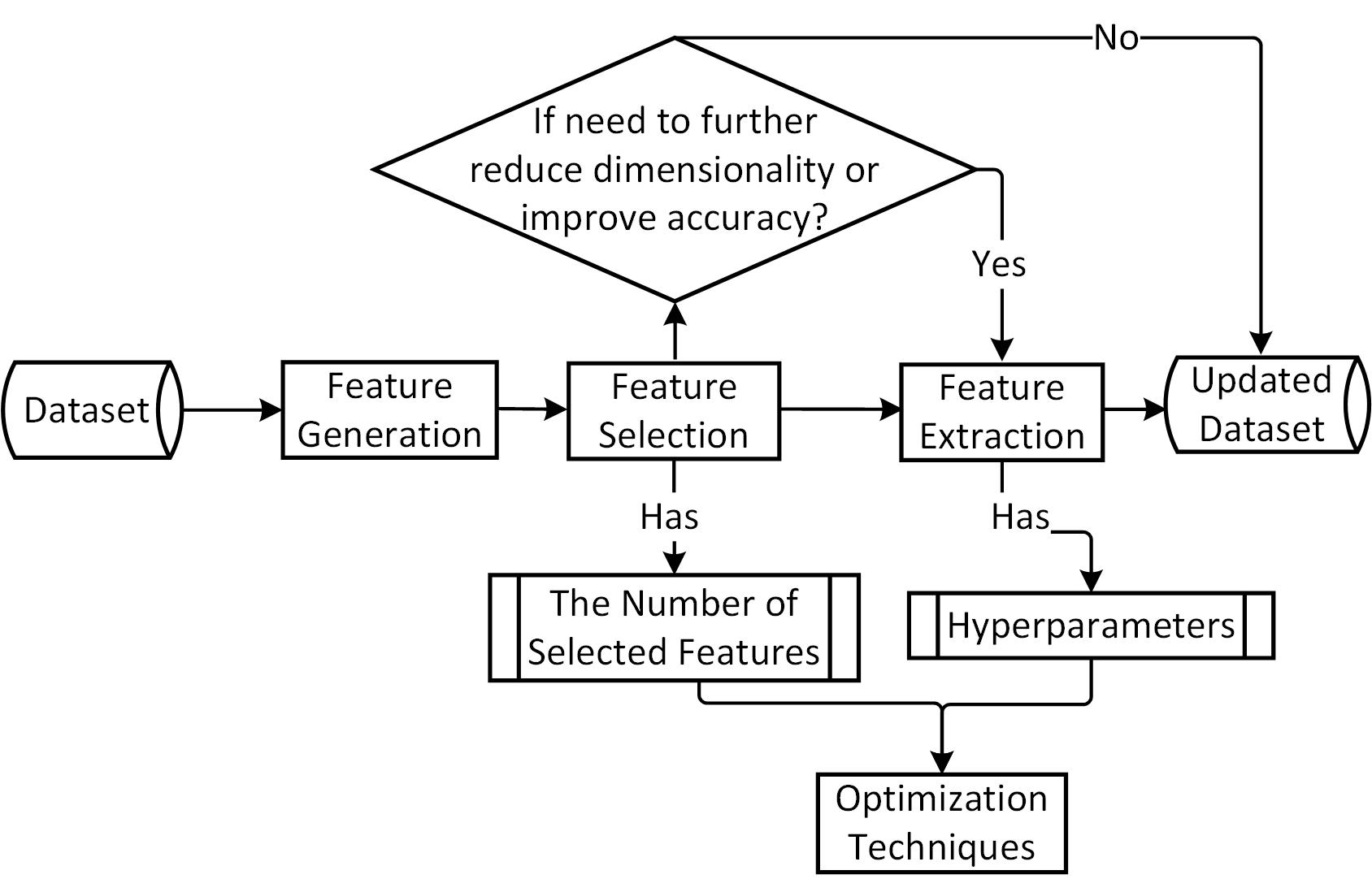}
     \caption{An automated feature engineering framework.} 
     \label{fe}
\end{figure}

Most state-of-the-art AutoFE approaches, like AutoFeat, use the generate-and-select strategy. In this strategy, an exhaustive feature pool is generated, and then valuable features are selected from it \cite{autofe1}. Certain procedures, like feature selection and extraction, may require the use of optimization techniques to identify appropriate parameters that can return the optimal model. As shown in Fig. \ref{fe}, the main procedures for the AutoFE framework with the generate-and-select strategy are as follows:
\begin{enumerate}
\item Generate a variety of candidate features using common operations;
\item Select important features using feature selection methods;
\item Determine the optimal number of features using optimization techniques;
\item Further extract features and reduce the data dimensionality if it is still high or the learning performance still needs to be further improved.
\end{enumerate}

\subsection{Feature Generation}
Feature generation is the process of generating new features by transforming or combining existing features to improve the generalizability and robustness of a ML model \cite{auto2}. In real-world IoT systems, data is often scattered over many devices and files and must be combined into a single database with rows for observations and columns for features \cite{auto4}. Although the feature generation step often requires domain knowledge from experts, certain features can still be created automatically to extract useful information. 

Common feature generation operations can be classified as follows \cite{gra3}:
\begin{enumerate}
\item \textbf{Unary operations}: Numerical feature discretization or normalization, time expansion, or mathematical operations like a logarithm.
\item \textbf{Binary operations}: The combination of two features using feature correlations or mathematical operations (e.g., addition, subtraction, multiplication, division, etc.). 
\item \textbf{High-order operations}: The calculation of multiple records for one feature, like the maximum, minimum, average, or median values.
\end{enumerate}

It is challenging to manually produce all meaningful and useful features. The process of generating valuable features usually requires human expertise. For automated feature generation, the general process is to automatically generate a large number of features using various operations and use feature selection techniques to select the relevant and useful features. Although generating numerous features is often time-consuming, automated feature generation can largely reduce human efforts and overhead by getting rid of the dependence on human expertise. Decision tree-based and GA-based methods described in \cite{fg2} can be used to simplify the feature generation process by defining and exploring the feature space.

\subsection{Feature Selection}
While feature generation may create a large number of features, some of them may be irrelevant or redundant. For specific tasks, certain features have a great impact on the target variable prediction, while other features may have a minimal or negative effect on the prediction \cite{auto4}. Therefore, Feature Selection (FS) should be implemented to identify the most appropriate features for use in constructing a more efficient and accurate learning model \cite{alex}. FS is a time-consuming and challenging procedure in ML pipelines, especially for high-dimensional datasets. 

Automated feature selection is the process of automatically selecting a subset of the original feature set to improve ML model performance and training speed by removing irrelevant and redundant features \cite{gra3}. To achieve AutoFS, the FS problem can be framed as an optimization problem \cite{fs1}. For a small number of input features, all combinations of the features can be evaluated to detect the best-performing feature set. On the other hand, for a large number of features, optimization techniques can be utilized to explore the feature search space and identify the optimal feature set.  

Existing FS methods can be divided into three categories: filter methods, wrapper methods, and embedded methods.

Filter methods assign a score to each feature by calculating its importance, and then select a subset of features based on a given threshold (e.g., the number of selected features or the accumulated importance). Each feature's score can be estimated using a variety of measures, including Information Gain (IG), the chi-square test, Pearson correlation coefficient, variance, etc. For example, the Fast Correlation Based Filter (FCBF) is a popular filter method that measures the correlation of features and selects features by calculating the symmetrical uncertainty (SU) \cite{mth}.

Wrapper methods make predictions based on a selected subset of features, and then evaluate the feature set according to prediction accuracy. Recursive Feature Elimination (RFE) is a wrapper method that recursively evaluates subsets of features to remove irrelevant features until a desired number of features are selected \cite{fs2}. 

Embedded methods indicate the FS process included in the learning process of ML models, like Lasso regularization, DT-based algorithms, and DL models. Thus, using those ML algorithms with embedded FS functionality often does not need additional FS procedures.

These three types of FS methods have different advantages and limitations. The major advantage of filter methods is that they can be completed prior to model training, which results in a relatively fast execution time \cite{fs2}. However, since filter approaches derive their features only via statistical measurements, they may not be optimal for all ML models. Embedded methods can select the relevant features for a specific ML model in its construction process \cite{fs2}. Thus, it can often achieve optimal performance on this specific ML model. However, the selected features are often only beneficial for the same type of model. Additionally, applying embedded methods to select features for another ML model is often time-consuming. Therefore, using an embedded method in the ML model where it is embedded is often the most appropriate choice. Wrapper techniques can be used for various ML models to select the most relevant features \cite{fs2}. However, wrapper methods are often more time-intensive than filter and embedded methods, as they need to continually train a ML algorithm on different subsets of features until the termination conditions are met.

In conclusion, when choosing feature selection approaches, a trade-off between the time complexity and learning performance must be made. Different FS approaches should be selected in different situations or tasks. Filter methods are often utilized in tasks with strict time constraints, while wrapper techniques often work better for tasks that demand great performance. Embedded methods are often used in certain ML algorithms that already have these embedded FS functionalities.

\subsection{Feature Extraction}

Feature extraction is the process of reducing dimensionality using mapping functions. Unlike feature generation, which preserves the original features, feature extraction alters the original features to extract more informative features that can replace the original features \cite{alex}. Through feature extraction, a more concise representation of the original dataset can be obtained. Additionally, model learning efficiency may be increased by dimensionality reduction \cite{fg2}. Common feature extraction methods include PCA, Linear Discriminant Analysis (LDA), and AE \cite{auto2}. Feature extraction is not a required procedure in the general feature engineering process. It is often utilized only when the feature set produced after feature generation and selection is still high dimensional or under-performing, since feature extraction can further reduce dimensionality and misleading feature components.  

\section{Automated Model Updating by Handling Concept Drift}

\subsection{Model Drift in IoT Systems}
Because of the dynamic IoT environments, IoT online data analysis often encounters concept drift issues when data distributions shift over time. Concept drift often impairs the performance of IoT data analytics models, posing significant threats to IoT services. To deal with concept drift, a successful data analytics model must reliably identify and respond to detected drifts in order to retain high prediction accuracy.

Concept drift refers to the conceptual and unpredictable changes in data streams \cite{dr1}. The presence of concept drift has brought significant challenges to the development of ML models. As the majority of ML models are built with the premise that the data is collected in a static environment, they lack the adaptability to learn streaming data with concept drift \cite{dr2}. Thus, in an ever-changing environment with concept drift issues, ML models’ performance may gradually degrade. When concept drift occurs, it is necessary to upgrade the current ML model in order to preserve or enhance model performance \cite{dr3}. This process is also referred to as automated model updating. To ensure high performance in a non-stationary environment, an IoT data analytics model should be updated automatically when concept drift occurs. 

\subsection{Concept Drift Definition}
In non-stationary and dynamically changing environments, the distribution of input data often changes over time, causing concept drift. Given an instance $(\boldsymbol{X},y)$ with the input features $\boldsymbol{X}$ and the target variable $y$, concept drift that occurs between the time points $t_0$ and $t_1$ can be denoted by \cite{dr4}:
\begin{equation}
\exists X: P_{t_{0}}(\boldsymbol{X}, y) \neq P_{t_{1}}(\boldsymbol{X}, y)
\end{equation}
where $P_{t_{0}}$ represents the joint distribution between $\boldsymbol{X}$ and $y$ at time $t_0$.

The joint probability $P_{t}(\boldsymbol{X}, y)$ can be calculated by \cite{dr1}:
\begin{equation}
P_{t}(\boldsymbol{X}, y)=P_{t}(\boldsymbol{X}) \times P_{t}(y \mid \boldsymbol{X})
\end{equation}
where $P_{t}(\boldsymbol{X})$ denotes the marginal probability and $P_{t}(y \mid \boldsymbol{X})$ represents the posterior probability.

Although changes in different probabilities can result in concept drift, only the distribution changes that affect the performance of learning models should be dealt with for data learning purposes \cite{dr4}.  Thus, the changes in the posterior probability, $P_{t}(y \mid \boldsymbol{X})$, are referred to as real concept drift because they cause model decision boundary changes and model performance degradation; other types of drift, like changes in $P_{t}(\boldsymbol{X})$, are referred to as virtual concept drift and are not taken into account in the learning system adaptation procedures \cite{dr1}. 

\begin{figure}
     \centering
     \includegraphics[width=13cm]{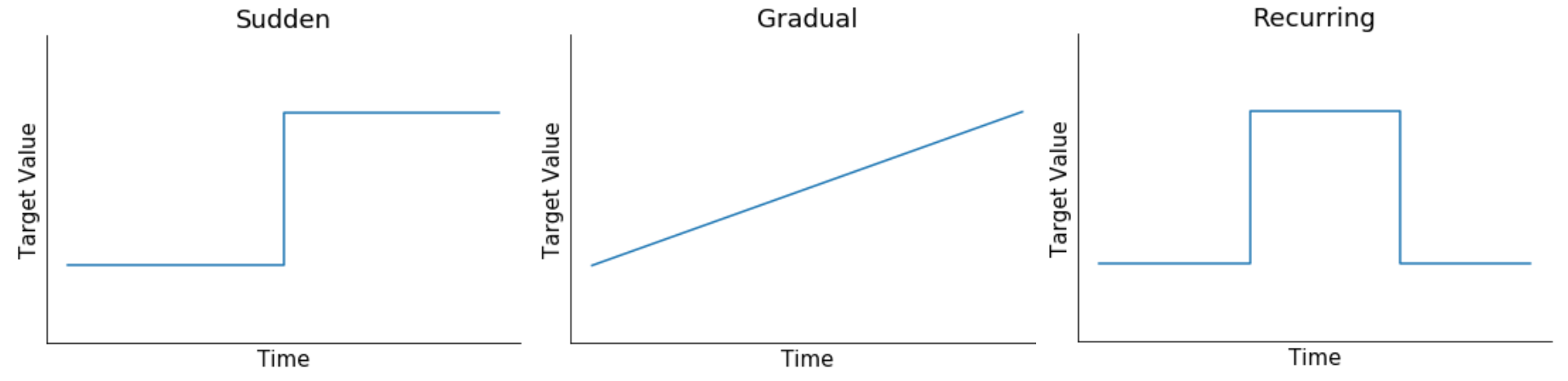}
     \caption{Concept drift types.} 
     \label{drift_types}
\end{figure}

As illustrated in Fig. \ref{drift_types}, there are three major types of data distribution changes that can cause concept drift: sudden, gradual, and recurring drift.
\begin{enumerate}
\item Sudden drift is the term used to describe the rapid and irreversible changes that occur in a short period of time.
\item Gradual drift occurs when a new data distribution gradually replaces an older one over time.
\item Recurring drift is a temporary change in the distribution of data. The distribution will return to its previous state within a certain period of time.
\end{enumerate}

Due to the occurrence of concept drift, the learning system must detect drift in time and update itself by adapting to the detected drift; hence, accurate predictions can be made on the continuously arriving data streams. Therefore, in addition to the training and prediction procedures in traditional ML models, there are two additional procedures for analyzing streaming data with concept drift: drift detection (detect the occurrence and the time of drift) and drift adaptation (handle the detected drift) \cite{dr1}. In this Section, the state-of-the-art drift detection and adaptation methods are described. Additionally, the strengths and limitations of the drift detection and adaptation methods discussed in this paper are summarized in Table \ref{t5}.

\begin{table*}[htbp]
\caption{The comparison of concept drift methods for automated model updating.}
\setlength\extrarowheight{1pt}
\centering
\scriptsize
\begin{tabular}{p{1.4cm}|p{1.6cm}|p{1.3cm}|p{3.6cm}|p{3.6cm}}
\Xhline{1.2pt}
\textbf{Task}                       & \textbf{Category}                             & \textbf{Methods}      & \textbf{Strengths}                                                                                                                                                     & \textbf{Limitations}                                                                                                                                                                                            \\ 
\hline
\multirow{4}{1.5cm}{Drift
  Detection}  & \multirow{2}{1.5cm}{Distribution -based
  Methods} & ADWIN                 & \begin{tabular}[c]{@{}p{3.6cm}}· Work well with gradual drifts. \\· Good interpretability.\end{tabular}                                                    & \begin{tabular}[c]{@{}p{3.6cm}}· A single ADWIN model is limited to one-dimensional data. \\· Characteristic values used by ADWIN are not always effective.\end{tabular}                            \\ 
\cline{3-5}
                                    &                                               & IE, KL Divergence     & \begin{tabular}[c]{@{}p{3.6cm}}· Good interpretability. \\· Can work with unlabeled data.\end{tabular}                                                     & \begin{tabular}[c]{@{}p{3.6cm}}· High computational cost. \\· May detect virtual drifts. \\· Require pre-defined time periods.\end{tabular}                                                 \\ 
\cline{2-5}
                                    & \multirow{2}{1.5cm}{Performance -based
  Methods}  & DDM                   & \begin{tabular}[c]{@{}p{3.6cm}}· Work well with sudden drifts. \\· Can ensure all detected drifts are real drifts.\end{tabular}                            & \begin{tabular}[c]{@{}p{3.6cm}}· Slow reaction time.  \\· Ineffective for gradual drifts. \\· Need to tune the drift and warning thresholds.\end{tabular}                                   \\ 
\cline{3-5}
                                    &                                               & EDDM                  & ·
  Work
  better with gradual drifts than DDM.                                                                                                                & \begin{tabular}[c]{@{}p{3.6cm}}· Sensitive to noise. \\· Need to tune the drift and warning thresholds.\end{tabular}                                                                                \\ 
\hline
\multirow{7}{1.5cm}{Drift
  Adaptation} & \multirow{3}{1.5cm}{Model
  Retraining}           & Full Retraining       & \begin{tabular}[c]{@{}p{3.6cm}}· Easy to understand and implement. \\· Can retain all the existing concepts.\end{tabular}                                  & \begin{tabular}[c]{@{}p{3.6cm}}· Time-consuming due to unnecessary retrainings. \\· May become extremely slow as data increases.\end{tabular}                                                       \\ 
\cline{3-5}
                                    &                                               & Partial
  Retraining  & \begin{tabular}[c]{@{}p{3.6cm}}· Easy to understand and implement. \\· Can remove outdated samples.  \\· Faster than full retraining.\end{tabular} & \begin{tabular}[c]{@{}p{3.6cm}}· May lose historical patterns. \\· Unnecessary retrainings.\end{tabular}                                                                                            \\ 
\cline{3-5}
                                    &                                               & Instance Weighting    & \begin{tabular}[c]{@{}p{3.6cm}}· Can retain all the existing concepts. \\· Can better adapt to drifts than full and partical retrainings.\end{tabular}     & \begin{tabular}[c]{@{}p{3.6cm}}· Time-consuming due to unnecessary retrainings. \\· Need to choose an updatable learner capable of weighted learning.\end{tabular}                                 \\ 
\cline{2-5}
                                    & \multirow{2}{1.5cm}{Incremental
  Learning}       & HT, VFDT,
  CVFDT     & \begin{tabular}[c]{@{}p{3.6cm}}· Can be continuously updated.  \\· Fast training speed due to partial updating.\end{tabular}                               & \begin{tabular}[c]{@{}p{3.6cm}}· Incapable of directly addressing concept drift. \\· Limited ML algorithms support incremental learning.\end{tabular}                                               \\ 
\cline{3-5}
                                    &                                               & AONN                  & ·
  Strong
  adaptability to drifts.                                                                                                                           & \begin{tabular}[c]{@{}p{3.6cm}}· Ineffective for sudden drifts. \\· Time-consuming.\end{tabular}                                                                                                    \\ 
\cline{2-5}
                                    & \multirow{2}{1.5cm}{Ensemble
  Learning}          & SEA, AWE,
  ACE       & \begin{tabular}[c]{@{}p{3.6cm}}· Can retain historical concepts.  \\· Strong adaptability to drifts. \\· Good generalizability.\end{tabular}       & \begin{tabular}[c]{@{}p{3.6cm}}· Need to determine a proper chunk size.  \\· Time and memory consuming.  \\· Outdated concepts may be misleading.\end{tabular}  \\ 
\cline{3-5}
                                    &                                               & ARF, SRP, LB,
  PWPAE & \begin{tabular}[c]{@{}p{3.6cm}}· Strong adaptability to drifts. \\· Good generalizability.\end{tabular}                                                    & \begin{tabular}[c]{@{}p{3.6cm}}· Time-consuming.  \\· Require high memory space.\end{tabular}                                                                                                       \\

\Xhline{1.2pt}

\end{tabular}
\label{t5}%
\end{table*}

\subsection{Drift Detection}
To design a model capable of dealing with concept drift, it should be able to effectively detect drift nodes and address the drift rapidly. Thus, drift detection is critical functionality for adaptive ML models capable of resolving concept drift problems.  

Drift detection methods are generally classified into two main categories: distribution-based methods and performance-based methods. Distribution-based methods identify concept drift by detecting the changes in data distributions. Statistical variables, such as the mean, variance, and class imbalance, can be used to quantify data distribution changes. In model-based methods, concept drift is measured based on the changes in the metrics used to assess model performance. For example, accuracy degradation and error rate increase are common indicators of concept drift. The severity of concept drift can be measured by the degree of model performance degradation.

\subsubsection{Distribution-based methods}
Distribution-based drift detection techniques are developed by measuring and comparing the data distributions of old and new data in time windows. Significant data distribution changes often cause concept drift and trigger model updates \cite{dr1}. There are several approaches to measure the data distributions of different time windows, like mean, variance, information entropy, Kullback-Leibler (KL) divergence, etc. 

ADaptive WINdowing (ADWIN) \cite{adwin} is a distribution-based approach for detecting concept drift using variable-size sliding windows and characteristic values (e.g., mean, variance). If no noticeable drift or distribution change is identified when the streaming data enters the model, the window size is dynamically enlarged, whereas the window size is reduced when concept drift is identified \cite{dr3}. The main procedures of ADWIN are as follows:

\begin{enumerate}
\item For a sliding window W, the characteristic values (e.g., mean, variance) between its two sub-windows, $W_1$ and $W_2$, is computed and compared. $W_1$ and $W_2$ represent earlier and more recent data, respectively.
\item A concept drift alarm will be triggered if the characteristic values of $W_1$ and $W_2$ diverge significantly enough (i.e., the difference exceeds a specific threshold).
\item Once a drift has been detected, the sliding window size is adjusted to the newer sub-window, $W_2$, while the older subwindow, $W_1$, is dropped. 
\end{enumerate}

ADWIN is well-suited for data streams with gradual drift because the sliding window can be enlarged to a large size window for detecting long-term changes. On the other hand, a single ADWIN model can only handle one-dimensional data. As a result, multiple ADWIN base models with discrete windows for each dimension are required for multi-dimensional data \cite{iot3}. Furthermore, the mean value is not always a suitable method to define changes. 

Distribution-based approaches can be used to detect concept drift using other metrics. The Information Entropy (IE) \cite{ie} is a widely-used distance metric to quantify how much information is included in a data distribution. The entropy of a data distribution $X$ can be calculated by: 
\begin{equation}
H(X)=-\sum_{x \in X} p(x) \log p(x)
\end{equation}

Assuming the two probability distributions are $p$ and $q$, the IE-based method calculates their distance based on the difference between their entropy values:
\begin{equation}
D_{IE}(p \| q)=|H(p)-H(q)|
\end{equation}

KL divergence \cite{kl} is another common distant metric to compute the distance between two distributions.

For the two probability distributions $p$ and $q$, KL divergence estimates the distance by:
\begin{equation}
D_{KL}(p \| q)=\sum_{x \in X} p(x) \log \frac{p(x)}{q(x)}
\end{equation}

In IE or KL divergence-based methods, concept drift occurs when $D_{IE}(p \| q)$ or $D_{KL}(p \| q)$ is greater than a threshold $\alpha$. 

Distribution-based methods are often used in systems with limited memory, like IoT devices, since they need only the most recent samples to be stored. Additionally, distribution-based approaches can provide a high level of interpretability by illustrating how the data distribution varies over time. They can also generally specify the exact time and place of the drift, which is beneficial for drift adaptation. However, distribution-based approaches often incur higher computational costs than performance-based approaches and often require the use of predefined historical and new time periods.

\subsubsection{Performance-based methods}
Model performance-based methods track changes in learners' prediction error rates to identify concept drift \cite{pwpae}. If the data distribution is stationary and without drift, the error rate of a learner should gradually decrease or remain constant as more data samples are learned. Conversely, if a learner's error rate rises dramatically as more data is processed, this often reveals the occurrence of concept drift.

Drift Detection Method (DDM) is a popular performance-based method that measures model error rate and standard deviation changes using two pre-defined thresholds, the warning threshold and the drift threshold \cite{ddm}. With an estimating error rate at time $t$ is $p_t$, the standard deviation at time $t$ can be calculated based on the Binomial distribution: 
\begin{equation}
S_{t}=\sqrt{p_{t}\left(1-p_{t}\right) / t}
\end{equation}

The error rate and standard deviation can be used to determine whether the warning level or drift level has been exceeded: 
\begin{equation}
\left\{\begin{array}{c}
\text { if } p_{t}+s_{t} \geq p_{\min }+2 * s_{\min } \rightarrow \text { Warning Level } \\
\text { if } p_{t}+s_{t} \geq p_{\min }+3 * s_{\min } \rightarrow \text { Drift Level }
\end{array}\right.
\end{equation}
where $p_{\min }$ and $s_{\min }$ are the current error rate's minimum and standard deviation's minimum, respectively. If the warning threshold is exceeded, newly arrived samples will be archived for potential drift adaption. If the drift level is exceeded, the learning model will be updated with the newly collected data.

DDM often performs well on data streams with sudden drift, but its reaction time is often too slow for detecting gradual drift \cite{drp1}. 

Early Drift Detection Method (EDDM) is an improved version of DDM that uses the same warning and drift mechanism as DDM to identify concept drift \cite{eddm}. Unlike DDM, EDDM detects drift by calculating the change rate of the learner’s error rate, rather than the error rate itself. Although EDDM often outperforms DDM, it is still inferior to distribution-based approaches on the gradual drift. Additionally, since it is sensitive to noise, it may misclassify noise as drift, resulting in false alarms.

Both DDM and EDDM have three primary hyperparameters that have a direct impact on the accuracy of drift detection: 1) the warning threshold; 2) the drift threshold; 3) the minimum number of incoming samples before detecting the first drift \cite{ddmhp}.

\subsection{Drift Adaptation}
After detecting concept drift, it is necessary to update or reconstruct the existing models to handle the drift using proper drift adaptation methods. Drift adaptation techniques can be classified into three categories: 1) Model retraining; 2) Incremental learning methods; 3) Ensemble learning methods.
\subsubsection{Model Retraining}
Model retraining is a simple and straightforward method for reacting to concept drift \cite{dr1}. Owing to the fact that pre-trained or offline models cannot always precisely predict the future incoming streaming data due to concept drift, they can be retrained on the newly arrived data streams to maintain high performance. A conventional online learning strategy for analyzing data streams without concept drift detection is to update the learner regularly to fit the most recent data. However, using this method can result in unnecessary model retrainings or drift adaptation delays. Therefore, an appropriate drift detection method should be used together with learning models to determine when to retrain the learning model for timely and necessary updates. 

Model retraining strategies include full retraining, partial retraining, and instant weighting. Full retraining is the process of retraining the learning model on the entire dataset involving all available samples. Full retraining is easy to implement but often time-consuming. 

Partial retraining is developed by retraining the model on only certain parts of data to improve model updating efficiency. Using window-based strategies that retain the model only on the recent data can reduce the training time but may result in the loss of historical data patterns. Thus, it is critical to choose a proper window size. ADWIN \cite{adwin} is a well-performing drift detection method for model retraining since it uses a dynamic window to fit new data. Optimized Adaptive and Sliding Windowing (OASW) \cite{oasw} is another partial retraining model for IoT data stream analytics. It uses adaptive and sliding windows to detect concept drift and collect samples of a new concept. Thus, the learning model can be partially retrained on only the new concept samples to save training time.

Instead of directly retraining the learning model on new data, the instance weighting method is another popular model retraining technique \cite{dr2} \cite{retraining}. It adjusts the weights of data samples according to their timestamp or retention time. Recent samples will be assigned a higher weight, while old data samples will be assigned a reduced weight or even deleted from the training set. This method is based on the assumption that as time passes, outdated data samples become less relevant, and new data samples become more critical. As a result, an existing learning model can adjust to concept drift by retraining on weighted samples. For weighted model retraining, an updatable learner capable of weighted learning should be used.

\subsubsection{Incremental Learning methods}
In data stream analytics, new data samples are continuously being added to the learning system. Rather than learning offline on static data, an effective model needs to be continuously updated to adapt to the changing data distributions. Therefore, incremental learning has become a widely-used strategy in data stream analytics research. Incremental learning is the process of learning data samples sequentially and updating the learning model with each instance is processed \cite{inc1}. 

When new samples arrive, incremental learning approaches often partially update the learning model to fit the new samples \cite{drel1}. Due to the progressive learning ability of incremental learning approaches, they do not need a sufficient amount of data prior to the training process. However, only a small number of ML algorithms enable partial updates, including MLP, multinomial NB, etc. Thus, several new incremental learning methods for concept drift adaptation, such as Hoeffding Trees (HT) \cite{ht1} based methods and Adaptive Online Neural Network (AONN) \cite{aonn}, were proposed. HT algorithms based on Hoeffding's inequality are one of the most common incremental learning methods for data stream analytics. By using the Hoeffding bound to calculate the number of samples required to determine the split node, the nodes in HTs can be partially updated as new samples arrive \cite{pwpae}. There are several variants of HTs, including the Very Fast Decision Tree (VFDT), Concept-adapting Very Fast Decision Tree (CVFDT), Extremely Fast Decision Tree (EFDT), etc. \cite{dr4}.

VFDT is a technique for creating classification decision trees in a data stream mining environment by using Hoeffding inequality \cite{vfdt}. It is constructed by continuously replacing leaf nodes with branch nodes to preserve an essential statistic at each decision node, and the splitting test is performed when the statistic of the node reaches a certain threshold. VFDT is an efficient method since it only has to process the data stream once. Additionally, it can often achieve high performance comparable to typical ML techniques. The primary drawback is that it is incapable of effectively addressing the concept drift issue. 

CVFDT extends VFDT to rapidly tackle the concept drift issue associated with data streams \cite{ht1}. CVFDT's basic principle is to replace the historical subtree with a new subtree that has a lower error rate. It uses a sliding window to choose test data samples and updates the resulting decision tree as data flows into and out of the time frame.

EFDT \cite{efdt}, also known as the Hoeffding Anytime Tree (HATT), is a modified version of the HT that divides nodes as soon as the confidence level is reached, rather than identifying the optimal split in the HT. This splitting method enables the EFDT to adjust more precisely to concept drifts than the HT, although its performance still has much room for improvement.

AONN is another incremental learning method based on neural network models \cite{aonn}. In AONN, a model update is triggered when the model's error increases. The AONN network is updated by either increasing the number of neurons in the output layers or by changing the weights of neurons using a batch of online data samples.
Incremental methods can often adapt to new data patterns by continually learning from newly received data samples. However, they are not specifically designed to address concept drift, as the old concepts and model components are still retained. Thus, they are ineffective in addressing certain types of drifts, like sudden drifts, which often need a completely new learner. 

\subsubsection{Ensemble Learning methods}
Ensemble learning techniques have been developed to generate powerful learners for data stream analytics in order to achieve greater concept drift adaptation. Ensemble learning is a ML technique that combines multiple base learners to tackle the same problem \cite{drel1}. In ensemble learning, base learners can be constructed using different algorithms, different hyperparameter configurations, or different subsets. As ensemble learning models aggregate the outputs of multiple base learners, they often have better generalizability than single models. For concept drift adaptation, reusing existing models in an ensemble is much more efficient than training new models on data streams with recurring concept drift \cite{dr1}. Ensemble methods for data stream analytics can be further classified as block-based ensembles and online ensembles \cite{drel2}. 

Block-based ensembles divide the data streams into discrete blocks with defined sizes and train a base learner on each block. When a new block is added, the existing base learners are evaluated and upgraded. Many block-based ensemble learning methods have been designed for concept drift adaptation, including Streaming Ensemble Algorithm (SEA), Accuracy Weighted Ensemble (AWE), Adaptive Classifier Ensemble (ACE), Learn++.NSE, Dynamic Weighted Majority (DWM), Diversity and Transfer-based Ensemble Learning (DTEL), etc.

SEA \cite{sea} is an ensemble learning model that adapts to concept changes by changing its structure. It constructs an ensemble of $N$ base learners, each trained on a batch of data samples. The final result is computed using the majority voting technique, which combines the prediction outcomes of base learners with the same weight. SEA limits the maximum number of base learners by the use of a threshold. Once the threshold is reached, the newly trained base learner will replace the worst-performing base learner according to the error rate and diversity. Experimental studies show that SEA is effective when the ensemble has no more than 25 base learners \cite{drp1}. 

AWE \cite{awe} is another ensemble learning approach that trains a base learner on each data chunk and combines the base learners, but it improves the technique of base learner replacement. Each incoming data chunk will be used to train a new base learner and evaluate the other existing base learners. The top $n$ best-performing base learners will be chosen to create a new ensemble model. Thus, the outdated base learners will be removed from the ensemble, leaving only those capable of effectively predicting the data with the new concept. The AWE method outperforms other methods when dealing with streaming data that contains recurring concept drift, and its performance on large streaming data will continue to improve \cite{drp1}. However, AWE's chunk-size selection remains a concern. Additionally, a noisy new data block may result in a biased ensemble \cite{drel3}. 

ACE \cite{ace} is a variant of AWE that is designed to deal with gradual drift. It continuously monitors the error rate change of each base learner in response to new input data, and removes the base models with degrading performance. ACE is effective at handling gradual drift, but struggles with sudden and recurring concept drifts. 

DWM \cite{dwm} is another ensemble model that trains multiple base learners but weights them differently based on their prediction performance. When a base learner makes an inaccurate prediction, its weight is slightly reduced. Additionally, if the ensemble model makes an incorrect prediction, a new base learner will be trained and given the highest weight among the base learners. The primary advantage of DWM is that it is capable of preserving historical models built on existing concepts. However, it may be resource-intensive, particularly when dealing with huge volumes of streaming data \cite{drel3}.

Learn++.NSE \cite{learn++} is an ensemble learning model that consists of multiple incrementally trained neural network models. Each base learner is trained on a single batch of incoming data. The Learn++.NSE model dynamically weights base neural network models depending on their error rates on the most recent batch of data. Additionally, the incorrectly predicted instances will be assigned a higher weight, allowing learners to concentrate on the challenging instances. When the ensemble model's prediction error rate exceeds a predefined threshold, a new base learner is trained and added to the ensemble \cite{drel3}. Learn++.NSE can handle sudden, gradual, and recurring drifts, because the base learners can be deactivated and reactivated by adjusting their weights \cite{dr1}. 

DTEL \cite{inc1} is an ensemble learning model that trains and stores each historical model from the initial models and then uses a transfer learning strategy to transfer the initial or historical models to new incoming data. To maintain model diversity in DTEL, it is crucial to train base models on a number of diverse data distributions or concepts. DTEL works effectively in the presence of recurring drift because historical models can be preserved and directly transferred to a new drift. On the other hand, owing to the utilization of transfer learning, DTEL often has a high learning efficiency and a quick response time to drift. 

Paired Learner (PL) technique \cite{pl} is an effective drift adapter that pairs a steady online learner with a reactive one to deal with concept drift. A stable learner makes predictions based on its entire experience, while an active learner makes predictions based on its most recent experience. Thus, the proper learner can be chosen for different scenarios. Comparative studies have revealed that the PL approach outperformed a wide variety of other ensemble methods or achieved equivalent performance at a much lower computational cost.

Online ensembles can enhance learning performance by integrating multiple incremental learning models, such as HTs. Gomes et al. \cite{arf} introduced the Adaptive Random Forest (ARF) technique, which makes use of HTs as base learners and ADWIN as the default drift detector for each tree. The drift detection process substitutes new trees that fit the new concept for underperforming base trees. ARF often outperforms a wide variety of other techniques, since the random forest method is also a well-performing ML technique. Additionally, ARF makes optimal use of resampling and is adaptable to a wide range of drift types. 

Gomes et al. \cite{srp} have presented a unique adaptive ensemble approach for streaming data analytics called Streaming Random Patches (SRP). SRP makes predictions using a combination of random subspace and online bagging techniques. SRP is similar to ARF in principle but employs a global subspace randomization mechanism rather than ARF's local subspace randomization. Global subspace randomization is a more flexible method of boosting the diversity of base learners. While SRP's prediction accuracy is often slightly higher than that of ARF, its execution time is frequently longer. The number of base learners and the embedded drift detector (e.g., ADWIN, DDM, EDDM, etc.) are the two significant hyperparameters of SRP and ARF models.

Leverage bagging (LB) \cite{lb} is another popular online ensemble that uses bootstrap samples to construct base learners. It employs the Poisson distribution to increase the diversity of input data and maximize bagging performance. While LB is simple to build, it often performs worse than SRP and ARF.

Performance Weighted Probability Averaging Ensemble (PWPAE) \cite{pwpae} is a novel online ensemble framework for concept drift adaptation. It uses the weighted prediction probabilities to integrate four base online learners: ARF-ADWIN, ARF-DDM, SRP-ADWIN, and SRP-DDM. PWPAE outperforms other compared drift adaptation methods as it uses dynamic weights to take advantage of other online learning models. However, the computational complexity of PWPAE is also higher than other methods. 

Although ensemble learning models often perform well when dealing with gradual and recurring drifts, they are incapable of coping with abrupt drifts owing to the ensemble learner's limited impact on a new base learner. In comparison to a single learner, however, using an ensemble learning model often increases computing complexity and costs. Thus, ensemble models that use the local learning strategy to train each base learner on a small local subset are more efficient in streaming data analytics \cite{drel1}.

\section{Selection of Evaluation Metrics and Validation Methods}

\subsection{Evaluation Metrics Selection}

To evaluate the learning model on a given IoT dataset, appropriate metrics should be selected in the AutoML pipeline, as they have a significant impact on model selection and HPO procedures. 

The performance metrics are mainly chosen according to the types of problems (e.g., accuracy, precision, recall, and F1-score for classification problems; Mean Squared Error (MSE), Mean Absolute Error (MAE), and Root Mean Squared Error (RMSE) for regression problems) \cite{mth} \cite{fadi1} \cite{metricR}.

\subsubsection{Classification Metrics}
Accuracy is the most basic metric, defined as the proportion of correctly categorized test instances to the total number of test instances \cite{preq1}. It is applicable to the majority of classification problems but is less useful when dealing with imbalanced datasets. Accuracy can be calculated by using True Positives (TPs), True Negatives (TNs), False Positives (FPs), and False Negatives (FNs):
\begin{equation}
Acc= \frac{T P+T N}{T P+T N+F P+F N} 
\end{equation}

Precision is the metric used to quantify the correctness of classification. Precision indicates the ratio of correct positive classifications to expected positive classifications. The larger the proportion, the more accurate the model, indicating that it is more capable of correctly identifying the positive class.
\begin{equation}
Precision=\frac{T P}{T P+FP} 
\end{equation}

Recall is a measure of the percentage of accurately recognized positive instances to the total number of positive instances.
\begin{equation}
Recall=\frac{T P}{T P+F N} 
\end{equation}

The F1 score is calculated as the harmonic mean of the Recall and Precision scores, therefore balancing their respective strengths.
\begin{equation}
F 1=\frac{2 \times T P}{2 \times T P+F P+F N} 
\end{equation}

The Receiver Operating Characteristic curve (ROC curve) plots the true positive rate against the false positive rate. AUC-ROC stands for Area Under Receiver Operating Characteristics, and a larger area indicates a more accurate model.

If class imbalance occurs, the F1-score or AUC-ROC should be used instead of accuracy to determine the optimal solution. Otherwise, a biased model may be returned.

\subsubsection{Regression Metrics}
In contrast to classification models, which produce discrete output variables, regression models aim to predict continuous output variables \cite{metricR}. As a result, relevant measures for evaluating regression models are appropriately established.

MSE is a straightforward measure that computes the difference between the actual and anticipated values (error), squares it, and then delivers the mean of all errors. MSE is very sensitive to outliers and will display a very large error rate even if a few outliers exist in otherwise well-fitted model predictions. Assuming $y$ is the real value and $\hat{y}$ is the estimated value, the MSE for a dataset of size $n$ can be denoted by: 
\begin{equation}
M S E=\frac{1}{n} \sum_{i=1}^{n}\left(y_{i}-\hat{y}_{i}\right)^{2}
\end{equation}

RMSE is the root of MSE. The advantage of RMSE is that it assists in reducing the magnitude of the mistakes to more interpretable numbers.
\begin{equation}
R M S E=\sqrt{\frac{1}{n} \sum_{i=1}^{n}\left(y_{i}-\hat{y}_{i}\right)^{2}}
\end{equation}

MAE is the average of the absolute error numbers (actuals – expectations).
\begin{equation}
M A E=\frac{1}{n} \sum_{i=1}^{n}\left|y_{i}-\hat{y}_{i}\right|
\end{equation}

MAE is the preferred method when outlier values need to be ignored, since it considerably reduces the penalty associated with outliers by deleting the square terms.

\subsubsection{Unsupervised Learning Metrics}
The Silhouette Coefficient quantifies how close or distant each point in one cluster is to each point in the other clusters \cite{sil}. Higher Silhouette values (closer to $+1$) indicate a strong separation of sample points from two different clusters. While a value of 0 indicates that the points are close to the decision boundary, values closer to $-1$ indicate that the points were incorrectly assigned to the cluster.

\subsubsection{Execution Time \& Memory}
Due to the fact that IoT systems often face strict time and memory limits, the execution time and memory usage of the AutoML model should also be considered \cite{vfdt}. The execution time comprises the time spent training and updating the model, as well as the time spent testing each instance. This can be used to determine if the learning model meets the requirements for real-time processing. Memory consumption can be used to determine if the size of the learning model and the memory used by it are smaller than the system memory on the IoT device, which is often used for edge computing; otherwise, the model must be implemented on a cloud server to ensure enough computational power and resources.

\subsection{Validation Method Selection}

\subsubsection{Hold-out Evaluation}
Hold-out evaluation is a frequently used evaluation method for ML algorithms \cite{dr4}. In hold-out evaluation, a hold-out subset is separated from the original dataset before the model training process. After training a model, its generalizability on the previously unseen dataset will be validated using the hold-out subset. For streaming data with concept drift, a hold-out evaluation will assess a learner at time $t$ by generating a hold-out subset that has the same concept at $t$. The current learning model is evaluated on the test sets at regular time intervals. Thus, for dynamic IoT data stream analytics, hold-out evaluation is only able to evaluate the learner’s performance on synthetic data with predefined drift times \cite{dr1}.

\subsubsection{Cross-Validation}
Cross-validation is an effective and popular evaluation method \cite{cv}. It uses a resampling strategy to evaluate ML models and assess how a model performs on different partitions of a given dataset. The k-fold cross-validation method is conducted by dividing the original dataset into $k$ equal subsets for $k$ different experiments; in each experiment, each of the subsets is selected as the validation set, while the other $(k-1)$ subsets are used as the training set. The average performance of the learning model in $k$ experiments is calculated as the final prediction performance. Unlike hold-out validation, which only assesses ML models on a subset of data and may produce biassed models that only perform well on a subset of data, using cross-validation can evaluate ML models on all subsets of data to avoid over-fitting. Cross-validation in IoT time-series data enables the evaluation of a learner on different time periods, assisting in the development of a complete and robust online learning model. 

The order of the data is critical for time-series-related problems. For time-related datasets, random split or k-fold split of data into train and validation may not yield good results. For the time-series dataset, the split of data into train and validation sets is according to the time, also referred to as the forwarding chaining method or rolling cross-validation. For a particular iteration, the future instances of train data can be treated as validation data.

\subsubsection{Prequential Evaluation}
Prequential evaluation, also named test-and-train validation, is one of the most appropriate methods to evaluate model learning performance on data streams generated in dynamic environments \cite{preq1} \cite{preq2}. In prequential evaluation, each incoming instance is firstly predicted by the learning model to update the metrics, and then learned by the model for model updating \cite{dr1}. The prequential error $E$ can be calculated by the sum of a loss function:
\begin{equation}
E=\sum_{t=1}^{n} f\left(y_{t}, \hat{y}_{t}\right)
\end{equation}
where $n$ is the total number of incoming samples, $y_t$ and $\hat{y}_{t}$ are the true and predicted values of the $t_{th}$ sample, the loss function $f$ can be selected from the metrics introduced in Section 8.1, based on the problem type. 

The prequential error is dynamically updated as new data samples arrive. Thus, using prequential evaluation can monitor the real-time performance of a learning model using metrics that change dynamically with each new data sample. Prequential evaluation can often be used to evaluate real-time model performance and take the most advantage of streaming data.

\section{Tools and Libraries}

\subsection{AutoML Tools}
Auto-Weka \cite{weka} is recognized as the first framework for AutoML. It is built on top of Weka, a well-known Java library package that contains a large number of ML methods. Bayesian optimization methods are the core strategies of Auto-Weka, including Sequential Model-based Algorithm Configuration (SMAC) and BO-TPEs, for both model selection and HPO procedures. 

Auto-Sklearn \cite{auto-sklearn} is a Python package for AutoML and CASH that is developed on top of Scikit-Learn. Auto-Sklearn introduced the concept of meta-learning for the model selection and HPO procedures. BO and ensemble approaches are employed in Auto-Sklearn to optimize the output models' performance. Both meta-learning and ensemble approaches can enhance the performance of model optimization. 

Hyperopt-Sklearn \cite{hyperopt} is AutoML framework built on the Scikit-learn library. Hyperopt-Sklearn utilizes Hyperopt to establish the search space for possible Scikit-Learn core components, such as the HPO and preprocessing approaches. Hyperopt supports a variety of optimization techniques for CASH, including random search and Bayesian optimization, for exploring search spaces of different types of variables.

Auto-Keras \cite{auto-keras} is an open-source AutoML library. It is developed on top of Keras, a well-known DL library. Auto-Keras implements NAS and HPO methods to design optimal DL models.

TPOT \cite{tpot} is a tree-based optimization framework for AutoML applications built on top of Scikit-Learn. It uses genetic algorithms to explore potential configurations by feature engineering and CASH procedures, thus finding the best solution.

H2O \cite{h2o} is an AutoML platform that supports both Python and R languages. H2O is capable of automating a wide variety of complex ML tasks, including feature engineering, model selection, model tuning, model visualization, and model validation. 

Amazon SageMaker \cite{aws} is an AutoML tool built on Amazon Web Services (AWS). It involves automated model tuning as a major module. In Amazon SageMaker, RS and BO methods are used to optimize ML models. It enables large-scale parallel optimization of complicated models and datasets.

\subsection{Online Learning and Concept Drift Adaptation Tools}
Several tools and frameworks are available for analyzing streaming data and resolving concept drift issues.

Massive Online Analysis (MOA) \cite{MOA} is an open-source tool for streaming data analysis. It is developed in Java and is based on the Waikato Environment for Knowledge Analysis (WEKA) platform. MOA is capable of detecting and adapting to concept drift using a number of strategies, including DDM, EDDM, and Hoeffding tree. Additionally, MOA contains various classes for creating streaming data, such as those for the Agrawal, Hyperplane, and Waveform datasets.

Scikit-multiflow (Skmultiflow) \cite{Skmultiflow} is a Python package for streaming data learning and concept drift adaptation. It provides many state-of-the-art streaming data learning algorithms, data generators, concept drift detection methods, and algorithms. The included drift detection methods are ADWIN, DDM, EDDM, and Page Hinkley. Streaming data learners for concept drift adaptation include KNN+ADWIN, Hoeffding adaptive tree, ARF, Oze bagging, etc. Stream data generators include Agrawal, Hyperplane, Led, Mixed, Random Tree, Waveform, etc. Skmultiflow supports both prequential and hold-out evaluations of models and all regularly used machine learning measures, such as accuracy, Kappa, and MSE. 

River \cite{River} is a Python library for data stream analytics and addressing concept drift through online ML models. All accessible learning models in River can be updated with a single incoming instance, allowing these methods to learn from data streams. It also includes a variety of streaming datasets, such as AirlinePassengers, Bananas, Bikes, ChickWeights, CreditCard, and Elec2. Additionally, it incorporates several well-known ML algorithms that support incremental learning, such as KNN, NB, and MLP. 

Scikit-learn (Sklearn) \cite{sklearn} is a popular ML library written in Python. Although Sklearn is primarily used for batch learning problems, it also provides several incremental learning methods for online learning and streaming data analytics, including multinomial NB, stochastic gradient descent (SGD), MLP, incremental PCA, etc.

\section{Case Study}
With the introduction of IoT data analytics and AutoML techniques, a case study is presented in this Section to illustrate the capabilities and advantages of AutoML techniques. A comprehensive AutoML pipeline is used in this case study to solve IoT anomaly detection problems.

This Section provides the experimental results of applying the complete AutoML pipeline to an IoT anomaly detection use case using real-world IoT datasets. The first subsection discusses the use case.  In the second part of this section, the experimental setup of the AutoML pipeline is described. In the last part, the results of offline IoT data analytics using traditional ML algorithms and dynamic IoT data analytics utilizing online adaptive algorithms are presented and analyzed. 

\subsection{Use Case}

IoT anomaly detection problems are selected as the case study for AutoML framework evaluation. Other alternative IoT data analytics use cases include smart healthcare medical diagnosis \cite{knna1} \cite{nba2}, smart citizen behavior classification \cite{ml1}, smart home monitoring \cite{xga1}, human gesture recognition \cite{xga2}, smart city analysis \cite{ml3}, intelligent transportation systems \cite{svmme} \cite{thesisme}, smart agriculture \cite{r191}, and Twitter sentiment analysis \cite{senti}. The reasons for choosing the IoT anomaly detection problems as the use case are as follows:
\begin{enumerate}
\item Compared with other IoT data analytics applications, IoT anomaly detection is a use case with many existing publications: \cite{oasw} \cite{knna3} \cite{nba1} \cite{mjdt} \cite{treeme} \cite{lighta1} \cite{pcaa1} \cite{rnna2} \cite{cnna2} \cite{cnnme} \cite{aea1}. Thus, IoT anomaly detection is one of the most popular and representative use cases for IoT data analytics. 
\item Unlike many other IoT data analytics use cases, IoT anomaly detection problems usually have concept drift issues. This is because IoT anomaly detection tasks usually aim to identify cyber-attacks with various patterns, and zero-day or previously-unseen types of attack data samples often have essentially different patterns, causing concept drift. The occurrence of concept drift enables the evaluation of the proposed AutoML framework in terms of online learning and automated model updating. It is difficult to evaluate the AutoML frameworks on other use cases without concept drift issues, because the performance of the ML models with or without automated updating would make little difference. 
\end{enumerate}

With the rapid development of IoT systems, numerous cyber-threats have extended from the Internet to people’s everyday devices. Current IoT systems are vulnerable to most existing cyber-threats, due to the limited IoT device capability, gigantic scale, and vulnerable environments \cite{iotad1}. Due to the paucity of IoT security mechanisms capable of dealing with IoT threats, it is critical for IoT system protection to develop advanced approaches for detecting and identifying abnormal IoT devices and events. Thus, IoT anomaly detection has become an important use case in recent IoT systems for detecting compromised IoT devices and malicious IoT attacks \cite{tnsm}.

For the purpose of enhancing IoT security, supervised ML algorithms can be used as effective mechanisms to distinguish malicious attack traffic from normal traffic. As discussed in Section 2.3, IoT anomaly detection problems can be classified into batch learning problems and online learning problems based on whether their environment is static or dynamic. In static IoT environments, traditional ML algorithms can be used to construct a conventional AutoML pipeline for static IoT data analytics. In dynamic IoT environments, online learning techniques can be used to construct a drift-adaptive AutoML pipeline for IoT streaming data analytics.

Two public IoT anomaly detection datasets are used in this work to evaluate the proposed AutoML pipeline. The first dataset is the IoT Intrusion Dataset 2020 (IoTID20) dataset proposed in \cite{iotid20}. This dataset was created by using normal and attack virtual machines as network platforms, simulating IoT services with the node-red tool, and extracting features with the Information Security Center of Excellence (ISCX) flow meter program. A typical smart home environment was established for generating this dataset using five IoT devices or services: a smart fridge, a smart thermostat, motion-activated lights, a weather station, and a remotely-activated garage door. Thus, the traffic data samples of normal and abnormal IoT devices are collected in Pcap files. 

The second dataset utilized in this paper is the Canadian Institute for Cybersecurity Intrusion Detection System 2017 (CICIDS2017) dataset \cite{cic}, which has the most updated network threats. The CICIDS2017 dataset is close to real-world network data since it has a large amount of network traffic data, a variety of network features (80), various types of attacks (14), and highly imbalanced classes. 

IoTID20 and CICIDS2017 datasets are both generic IoT network traffic datasets in a tabular format. They include various types of data, such as numerical, discrete, and string/text data. Thus, they are also representative IoT datasets for IoT applications. Other IoT anomaly detection datasets include KDD-99 \cite{kdd}, Kyoto 2006+ \cite{kyoto}, NSL-KDD \cite{nsl}, ISCXIDS2012 \cite{iscx}, and Bot-IoT \cite{botiot} datasets. 
The reasons for selecting CICIDS2017 and IoTID20 datasets are as follows:
\begin{enumerate}
\item Compared with other feasible alternative datasets, such as KDD-99 (proposed in 1999), Kyoto 2006+ (proposed in 2006), NSL-KDD (proposed in 2009), and ISCXIDS2012 (proposed in 2012), the two selected datasets are the most recent benchmark IoT anomaly detection datasets proposed in 2017 and 2020. Thus, CICIDS2017 and IoTID20 have state-of-the-art cyber-attack scenarios, which enables the case study to make more contributions to the IoT anomaly detection field. 
\item Compared with other IoT anomaly detection datasets, such as Bot-IoT, in which the percentage of attack data is much higher than the normal data, the two selected datasets are more representative and closer to real-world IoT traffic datasets, because only a small percentage of data samples are attack or abnormal data in the two datasets and real-world IoT systems usually maintain the normal state most of the time.
\item The two selected datasets, especially the CICIDS2017 dataset, were created by launching many new types of attacks on different days. These zero-day attacks made changes in the data distributions and patterns over time and caused concept drift issues, which enables the evaluation of the drift-adaptive online learning models in the proposed AutoML framework. 
\end{enumerate}

To conclude, selecting the CICIDS2017 and IoTID20 datasets can make the experimental results closer to the results of real-world IoT anomaly detection tasks, as they have the most updated attack scenarios and concept drift issues.

The proposed AutoML pipeline is evaluated using a reduced IoTID20 dataset with 62,578 entries and a reduced CICIDS2017 dataset with 28,307 records for the purpose of this work.

\subsection{Experimental Setup}
The studies use a comprehensive AutoML pipeline to solve the IoT anomaly detection problem, including AutoDP, AutoFE, automated model selection, and HPO procedures. The specifications of each procedure in the AutoML pipeline are presented in Table \ref{t6}.

\begin{table*} [htbp]
\caption{The specifications of the proposed AutoML pipeline.}
\setlength\extrarowheight{1pt}
\centering
\scriptsize
\begin{tabular}{p{1.5cm}|p{2.2cm}|p{2.2cm}|p{6cm}}
\Xhline{1.2pt}

\textbf{Category}                         & \textbf{Procedure}                        & \textbf{Method}                  & \textbf{Aim/Operation}                                                                                                                                                                   \\ 
\hline
\multirow{4}{*}{AutoDP}                   & Encoding                                  & Label Encoding                   & Identify and
  transform string features into numerical features to make the data more
  readable by ML models                                                                           \\ 
\cline{2-4}
                                          & Imputation                                & Mean Imputation                  & Detect and
  impute missing values to improve data quality                                                                                                                               \\ 
\cline{2-4}
                                          & Normalization                             & Z-Score or Min-Max Normalization & Normalize
  the range of features to a similar scale to improve data quality                                                                                                             \\ 
\cline{2-4}
                                          & Balancing                                 & SMOTE                            & Generate minority
  class samples to solve class-imbalance and improve data quality                                                                                                      \\ 
\hline
\multirow{2}{*}{AutoFE}                   & \multirow{2}{*}{Feature Selection}        & IG                               & Remove
  irrelevant features to improve model efficiency                                                                                                                                 \\ 
\cline{3-4}
                                          &                                           & Pearson Correlation              & Remove
  redundant features to improve model efficiency and accuracy                                                                                                                     \\ 
\hline
\multirow{6}{1.5cm}{Automated Model Learning} & \multirow{5}{*}{Model Selection}          & NB                               & \multirow{5}{6cm}{Select the
  best-performing model among five common ML models by evaluating their
  learning performance}                                                               \\ 
\cline{3-3}
                                          &                                           & MLP                              &                                                                                                                                                                                          \\ 
\cline{3-3}
                                          &                                           & KNN                              &                                                                                                                                                                                          \\ 
\cline{3-3}
                                          &                                           & RF                               &                                                                                                                                                                                          \\ 
\cline{3-3}
                                          &                                           & LightGBM                         &                                                                                                                                                                                          \\ 
\cline{2-4}
                                          & Hyperparameter Optimization               & BO-TPE                           & Tune the hyperparameters
  of the learning models to obtain the optimized models                                                                                                         \\ 
\hline
\multirow{4}{1.5cm}{Automated Model Updating} & \multirow{4}{2.2cm}{Adaptive Model Selection} & HT                               & \multirow{4}{6cm}{Select the
  best-performing model among four online adaptive models to adapt to dynamic
  data streams with concept drift issues for learning performance enhancement}  \\ 
\cline{3-3}
                                          &                                           & EFDT                             &                                                                                                                                                                                          \\ 
\cline{3-3}
                                          &                                           & ARF                              &                                                                                                                                                                                          \\ 
\cline{3-3}
                                          &                                           & SRP                              &                                                                                                                                                                                          \\

\Xhline{1.2pt}

\end{tabular}
\label{t6}%
\end{table*}

AutoDP involves automated encoding, imputation, normalization, and balancing procedures. The automated encoding procedure identifies and converts string features to numerical features to make the data more understandable for ML models. The automated imputation procedure includes detecting missing values and imputing missing values using the mean imputation method introduced in Section 5.3. The automated normalization process automatically chooses an appropriate normalization method from Z-score and min-max normalization methods based on their performance in anomaly detection.

As the CICIDS2017 and IoTID20 datasets are both highly-imbalanced datasets, with an abnormal/normal ratio of 19\%/81\% and 6\%/94\%, respectively, an automated data balancing technique is also implemented in the proposed AutoML pipeline to balance the datasets. The system will evaluate whether the incoming dataset is imbalanced (the abnormal/normal ratio is smaller than a threshold (e.g., 50\%)); and if it is, the SMOTE technique introduced in Section 5.4.2 will be automatically implemented to synthesize new samples for the minority class to balance the data. As described in Section 5.4.2, apart from SMOTE, there are other data balancing methods, such as Random Under-Sampling (RUS) and Random Over-Sampling (ROS). 
SMOTE is selected over other data balancing methods in the proposed AutoML framework due to the following reasons:
\begin{enumerate}
\item Unlike RUS, which may cause the loss of critical information by using the under-sampling strategy to remove majority class samples, SMOTE is an over-sampling method that synthesizes new samples for the minority class to balance data without discarding any existing samples. 
\item Compared with other over-sampling methods, like ROS, which simply replicates the instances, SMOTE uses the principle of KNN to create high-quality new samples.
\end{enumerate}

Thus, using SMOTE can balance the datasets without losing critical samples or adding high-quality samples, which can improve data quality and avoid obtaining biased models.

As both datasets used in the experiments have a large number of features totaling more than 80, the AutoFE technique in the proposed AutoML pipeline focuses primarily on feature selection in order to obtain a sanitized and optimal feature subset. In the first step of AutoFE, an IG-based method is used to remove irrelevant or unimportant features by measuring the importance of each feature. It is selected in the proposed AutoML pipeline mainly due to the following reasons:
\begin{enumerate}
\item Compared with certain wrapper FS methods, like RFE, which recursively evaluates subsets of features to remove irrelevant features, the IG-based method has better interpretability because it can generate importance scores for each feature from the dataset.
\item Compared with embedded FS methods, like Lasso regularization and DT-based algorithms, which train ML models to calculate feature importance scores, the IG-based method works directly on the correlations between the target variable and input features without additional ML model training. It has a fast speed due to its low computational complexity of $O(n)$ \cite{igfs}. 
\end{enumerate}

Using the IG-based FS method can remove irrelevant features and reduce data complexity, thus improving model training and testing efficiency.

In the second step of AutoFE, a Pearson correlation-based method is used to remove redundant and noising features by calculating the correlation between different features. By removing unnecessary and redundant information using AutoFE, the learning model becomes more efficient and accurate at detecting anomalies. Among correlation-based FS methods introduced in Section 6.3, such as the chi-square test, Pearson correlation coefficient, and variance, the Pearson correlation-based method is selected as the second FS method to remove redundant and noising features, mainly due to the following reasons: 
\begin{enumerate}
\item Pearson correlation-based FS method can determine the exact degree in the range of 0 to 1 to which every two features are correlated using its Pearson formula.
\item This method can determine the direction of the correlations according to whether the correlation values between features are positive or negative. 
\item This method is a fast FS method with low computational complexity of $O(nlogn)$ \cite{pearson}.
\end{enumerate}

Using the Pearson correlation-based FS method can remove redundant and noisy features to avoid using disturbing data and biased models, thus improving model learning efficiency and accuracy.

Automated model selection is an essential procedure in the development of AutoML pipelines. The experiments are divided into two parts, one for batch learning in static IoT environments and the other one for online learning in dynamic IoT environments. 

For batch learning in static IoT environments, the learning models are chosen from five representative candidate ML algorithms (NB \cite{nba1}, MLP \cite{dl2}, KNN \cite{knna3}, RF \cite{ml2}, and LightGBM \cite{lighta1}). ML algorithms have proven to be the most effective solution for general data analytics problems \cite{Inj_ML}. For IoT data analytics problems, all regular ML algorithms demonstrated in Section 3 are other feasible alternatives, including Support Vector Machine (SVM), Decision Tree (DT), XGBoost, LightGBM, MLP, CNN, RNN, etc. The reasons for selecting the NB, MLP, KNN, RF, and LightGBM algorithms for batch learning in static IoT environments are as follows:
\begin{enumerate}
\item NB, MLP, KNN, RF, and LightGBM are all popular IoT data analytics algorithms that have been widely used in many IoT applications, such as IoT anomaly detection \cite{oasw} \cite{knna3} \cite{nba1} \cite{lighta1}, medical diagnosis \cite{knna1} \cite{nba2}, smart citizen behavior classification \cite{ml1}, etc. 
\item NB and KNN are two basic and representative ML algorithms with low complexity. Thus, they can usually learn simple datasets without over-fitting at a fast speed. They are selected for low-complexity IoT data analytics. 
\item As Deep Learning (DL) is an essential type of ML algorithm, MLP, a basic and representative DL algorithm for IoT anomaly detection use cases, is selected in the proposed AutoML framework. Other DL algorithms, like CNNs and RNNs, are primarily used for image processing and Natural Language Processing (NLP) applications.
\item RF and LightGBM are two robust ensemble ML models built on multiple DTs. They have shown success in many data analytics tasks because they have great generalizability. Thus, they are selected as representative ensemble ML models for complex IoT data analytics. Moreover, as they have a large number of hyperparameters that require tuning, they often benefit more from HPO and AutoML than other ML algorithms. 

In conclusion, we selected two representative low-complexity ML algorithms (NB and KNN), a representative DL algorithm (MLP), and two representative robust ensemble algorithms (RF and LightGBM) to represent common ML algorithms for IoT data analytics.
\end{enumerate}

For online learning in dynamic IoT environments, the learning model is chosen from four drift-adaptive online learning algorithms (HT \cite{ht1}, EFDT \cite{efdt}, ARF \cite{arf}, and SRP \cite{srp}), as it needs to be updated automatically based on data distribution changes (concept drift). As described in Section 7.4, incremental learning and ensemble learning are two primary types of drift-adaptive online learning algorithms. Other feasible alternative incremental learning methods include VFDT, CVFDT, EFDT, etc., and other alternative ensemble methods include SEA, AWE, ACE, LB, PWPAE, etc. The reasons for selecting HT, EFDT, ARF, and SRP, for online learning are as follows: 
\begin{enumerate}
\item HT and EFDT are two representative incremental learning methods for drift adaptation. HT is selected because it is the most basic incremental online learning method that is the base model of many other online learning methods, such as VFDT, CVFDT, EFDT, LB, ARF, and SRP. It is used as the baseline model for comparison with other advanced models. EFDT is selected because it is the most state-of-the-art and stable tree-based incremental learning method that often achieves better performance than other incremental learning methods \cite{efdt}.
\item ARF and SRP are selected because they are the most state-of-the-art and stable ensemble learning methods. They have strong data stream analysis capability and often outperform other drift adaptation methods \cite{arf} \cite{srp}. Additionally, unlike block-based ensemble methods (e.g., SEA, AWE, ACE) that require the formation of data chunks, ARF and SRP can work directly on individual data samples to avoid result delays.
\item Using these four online learning algorithms allows us to comprehensively compare incremental learning and ensemble learning methods, the two primary types of drift-adaptive online learning algorithms.
\end{enumerate}

After evaluating the performance of each learning model based on accuracy and F1-scores, the best-performing and the second best-performing models using default hyperparameters are selected for further evaluations using HPO. By selecting not only the best-performing model but also the second-best-performing model, the probability of missing the real optimal model can be decreased. 

After selecting the top-two learning models, their hyperparameters are tuned by the HPO technique to obtain the two optimized models and then select the final optimal model. As shown in Tables \ref{t7} - \ref{t11}, the two best-performing batch learning algorithms on both datasets are RF and LightGBM, while the two best-performing online learning methods are ARF and SRP. Thus, the hyperparameters of these four algorithms are optimized. Table \ref{t7} illustrates the search space and the detected optimal values for the hyperparameters of these learning algorithms. Continuous hyperparameters are assigned a search range, while categorical hyperparameters are assigned all possible values/choices. 

BO-TPE is used as the optimization method for automated model selection and HPO. As discussed in Section 4.4, other optimization methods include GS, RS, BO-GP, BO-TPE, gradient-based models, hyperband, PSO, and RL. BO-TPE is selected for ML model optimization due to the following reasons:
\begin{enumerate}
\item It works well with large hyperparameter space and all types of hyperparameters. As the proposed AutoML framework uses several ML models with a large number of hyperparameters, such as RF, LightGBM, and MLP, certain other optimization methods that do not work well with large hyperparameter space and different types of hyperparameters, like GS, RS, BO-GP, Hyperband, and gradient-based models, are unsuitable for the proposed work. 
\item Compared with other complex optimization methods, like GA and RL, BO-TPE has a low time complexity of $O(nlogn)$ \cite{hpome}. 
\end{enumerate}

To conclude, using BO-TPE enables the proposed AutoML framework to return the optimized models with the best performance at a fast speed.

\begin{table*} [htbp]
\caption{The HPO configuration of well-performing learning models.}
\setlength\extrarowheight{1pt}
\centering
\scriptsize
\begin{tabular}{p{1.5cm}|p{3cm}|p{2.2cm}|p{2.4cm}|p{2.2cm}}
\Xhline{1.2pt}
\textbf{Model}            & \textbf{Hyperparameter Name} & \textbf{Configuration Space} & \textbf{Optimal Value on CICIDS2017} & \textbf{Optimal Value on IoTID20}  \\ 
\hline
\multirow{5}{*}{LightGBM} & n\_estimators                & {[}50,500]                   & 360                                  & 440                                \\ 
\cline{2-5}
                          & max\_depth                   & {[}5,50]                     & 36                                   & 38                                 \\ 
\cline{2-5}
                          & learning\_rate               & (0, 1)                       & 0.957                                & 0.456                              \\ 
\cline{2-5}
                          & num\_leaves                  & {[}100,2000]                 & 1100                                 & 1200                               \\ 
\cline{2-5}
                          & min\_child\_samples          & {[}10,50]                    & 50                                   & 25                                 \\ 
\hline
\multirow{5}{*}{RF}       & n\_estimators                & {[}50,500]                   & 460                                  & 220                                \\ 
\cline{2-5}
                          & max\_depth                   & {[}5,50]                     & 26                                   & 14                                 \\ 
\cline{2-5}
                          & min\_samples\_split          & {[}2,11]                     & 8                                    & 2                                  \\ 
\cline{2-5}
                          & min\_samples\_leaf           & {[}1,11]                     & 1                                    & 4                                  \\ 
\cline{2-5}
                          & criterion                    & {[}’gini’, ’entropy’]        & ’entropy’                            & ’entropy’                          \\ 
\hline
\multirow{2}{*}{ARF}      & n\_models                    & {[}3, 20]                    & 18                                   & 15                                 \\ 
\cline{2-5}
                          & drift\_detector              & {[}‘ADWIN’, ‘DDM’]           & ‘DDM’                                & ‘DDM’                              \\ 
\hline
\multirow{2}{*}{SRP}      & n\_models                    & {[}3, 20]                    & 14                                   & 10                                 \\ 
\cline{2-5}
                          & drift\_detector              & {[}‘ADWIN’, ‘DDM’]           & ‘DDM’                                & ‘DDM’                              \\

\Xhline{1.2pt}

\end{tabular}
\label{t7}%
\end{table*}

The evaluation method for the learning models is determined by the tasks and environments. For offline learning in static environments, 5-fold cross-validation is used in the experiments since it can help develop a generic and robust learning model. As described in Section 8.2, hold-out validation and cross-validation are the two commonly used validation methods for offline/static learning. Compared with hold-out validation, which only evaluates ML models on a certain part of data and may obtain biased models that only perform well on a specific area of the data, 5-fold cross-validation splits the dataset into five equal-sized folds and evaluates the ML models on each fold to evaluate their generalization capabilities. Obtaining the ML models with optimal cross-validation performance can prevent them from over-fitting the dataset. The number of folds, five, is selected by balancing the evaluation time and model generalizability \cite{cross}. If the number of folds is too large, it will take much additional time; if it is too small, it would be insufficient to evaluate the model's generalizability.
For online learning in dynamic environments, prequential evaluation introduced in Section 8.2.3 is used in the experiments to evaluate the long-term learning performance of the drift-adaptive models on IoT time-series data, as it is the standard evaluation method for online learning.

Lastly, the four common classification metrics, accuracy, precision, recall, and F1-scores, as well as a model efficiency metric, model learning time, are adopted to evaluate the AutoML framework's performance. The selected four classification performance metrics are the most common metrics that are used in most papers and can reflect the ML model performance sufficiently \cite{fadi1} \cite{f1}. As the IoT anomaly detection datasets are usually imbalanced datasets, many individual performance metrics, such as accuracy, precision, and recall, cannot reflect the model performance on imbalanced datasets alone. Thus, the four common classification metrics, accuracy, precision, recall, and F1-scores, are considered together to comprehensively compare the ML models and avoid biased evaluation results. The F1-score considers data distributions and uses the harmonic mean of the Recall and Precision scores to give a fair view of anomaly detection results and minimize bias. This is because, in F1-scores, both false negatives (measured by recall) and false positives (measured by precision) are taken into account \cite{f1}. Thus, it is chosen as the primary performance metric for evaluating the proposed AutoML pipeline.

On the other hand, due to the processing time and efficiency requirements of IoT systems, the model learning time is also calculated to compare the learning speed of different ML models. The final ML model that is suitable for IoT data analytics should strike a balance between the model's effectiveness and efficiency.

The experiments were conducted on a machine with an i7-8700 processor and 16 GB of memory, representing an IoT server machine that supports large IoT data analytics. The techniques and methods utilized in the studies are implemented using the Python packages: Auto-Sklearn \cite{auto-sklearn}, Hyperopt \cite{hyperopt}, Skmultiflow \cite{Skmultiflow}, and River \cite{River}.

\subsection{Experimental Results and Analysis}

In this work, two series of experiments were conducted to validate the effectiveness of the AutoML framework. The first series of experiments were conducted to assess an offline AutoML pipeline in static IoT environments, while the second series of experiments were conducted to evaluate an online AutoML pipeline in dynamic IoT environments. The primary difference between the offline and online AutoML pipelines is the learning models used in the framework (traditional ML models versus adaptive online models).

To assess the AutoML framework's performance, we evaluated the accuracy, precision, recall, F1-score, and model learning time when using AutoML versus when not using AutoML in both offline and online learning tests. The experimental results for offline learning on the CICIDS2017 and IoTID20 datasets are shown in Tables \ref{t8} and \ref{t9}, while the experimental results for online learning on the CICIDS2017 and IoTID20 datasets are presented in Tables \ref{t10} and \ref{t11}.  

Specifically, in each Table, three different sets of results are shown to demonstrate the performance of the proposed AutoML pipeline. The first set of results compares the performance of original ML algorithms with default hyperparameter configurations (without AutoML) as baseline models for comparison purposes. The second set of results shows the performance of ML algorithms after implementing the proposed AutoDP \& AutoFE procedures to illustrate the impact of data quality improvement by using AutoML. The third set of results presents the performance of a complete AutoML pipeline, which comprises AutoDP, AutoFE, automated model selection of the top-2 ML algorithms, and HPO. The proposed AutoML pipeline starts by implementing AutoDP and AutoFE, and then automatically selects the two best-performing learning models based on their F1-scores shown in the second set of results as the automated model selection procedure. After that, the hyperparameters of the two selected models are optimized to obtain a final optimal model with the best F1-score, as shown in the third set of results. The well-performing configurations in each set experiment are highlighted with boldface in Tables \ref{t8} - \ref{t11}.

\begin{table*} [htbp]
\caption{The experimental results of offline learning on the CICIDS2017 dataset using 5-fold cross-validation.}
\setlength\extrarowheight{1pt}
\centering
\scriptsize
\begin{tabular}{p{1.5cm}|p{1.7cm}|p{1.5cm}|p{1.5cm}|p{1.4cm}|p{1.4cm}|p{1.4cm}}
\Xhline{1.2pt}
\textbf{AutoML Procedures}        & \textbf{Learning Algorithm} & \textbf{Accuracy (\%)} & \textbf{Precision (\%)} & \textbf{Recall (\%)} & \textbf{F1 (\%)} & \textbf{Model Learning Time (s)}  \\ 
\hline
\multirow{5}{*}{No}               & NB \cite{nba1}                        & 72.545                 & 37.837                  & 62.253               & 47.896           & 0.2                               \\ 
\cline{2-7}
                                  & MLP \cite{dl2}                        & 88.536                 & 94.277                  & 43.701               & 58.830           & 63.5                              \\ 
\cline{2-7}
                                  & KNN \cite{knna3}                        & 97.238                 & 92.081                  & 93.782               & 92.923           & 9.1                               \\ 
\cline{2-7}
                                  & \textbf{RF} \cite{ml2}                & \textbf{99.703}        & \textbf{99.577}         & \textbf{98.830}      & \textbf{99.248}  & \textbf{17.2}                     \\ 
\cline{2-7}
                                  & \textbf{LightGBM} \cite{lighta1}          & \textbf{99.816}        & \textbf{99.543}         & \textbf{99.506}      & \textbf{99.525}  & \textbf{1.4}                      \\ 
\hline
\multirow{5}{*}{\shortstack{AutoDP \& \\AutoFE}} & NB                        & 73.316                 & 39.690                  & 73.080               & 51.435           & 0.1                               \\ 
\cline{2-7}
                                  & MLP                       & 85.968                 & 92.069                  & 26.563               & 44.831           & 55.2                              \\ 
\cline{2-7}
                                  & KNN                       & 97.058                 & 92.024                  & 92.831               & 92.423           & 9.1                               \\ 
\cline{2-7}
                                  & \textbf{RF}               & \textbf{99.735}        & \textbf{99.632}         & \textbf{99.012}      & \textbf{99.294}  & \textbf{13.3}                     \\ 
\cline{2-7}
                                  & \textbf{LightGBM}         & \textbf{99.844}        & \textbf{99.616}         & \textbf{99.579}      & \textbf{99.598}  & \textbf{0.9}                      \\ 
\hline
\multirow{2}{*}{All}              & RF                        & 99.760                 & 99.578                  & 99.141               & 99.368           & 62.3                              \\ 
\cline{2-7}
                                  & \textbf{LightGBM}          & \textbf{99.866}        & \textbf{99.670}         & \textbf{99.634}      & \textbf{99.653}  & \textbf{1.0}                      \\

\Xhline{1.2pt}

\end{tabular}
\label{t8}%
\end{table*}

\begin{table*} [htbp]
\caption{The experimental results of offline learning on the IoTID20 dataset using 5-fold cross-validation.}
\setlength\extrarowheight{1pt}
\centering
\scriptsize
\begin{tabular}{p{1.5cm}|p{1.7cm}|p{1.5cm}|p{1.5cm}|p{1.4cm}|p{1.4cm}|p{1.4cm}}
\Xhline{1.2pt}
\textbf{AutoML Procedures}        & \textbf{Learning Algorithm} & \textbf{Accuracy (\%)} & \textbf{Precision (\%)} & \textbf{Recall (\%)} & \textbf{F1 (\%)} & \textbf{Model Learning Time (s)}  \\ 
\hline
\multirow{5}{*}{No}               & NB \cite{nba1}                & 89.603          & 95.886          & 92.882          & 94.359          & 0.5            \\ 
\cline{2-7}
                                  & MLP \cite{dl2}               & 99.202          & 96.800          & 98.319          & 97.742          & 125.7          \\ 
\cline{2-7}
                                  & KNN \cite{knna3}               & 97.445          & 98.258          & 99.027          & 98.641          & 43.3           \\ 
\cline{2-7}
                                  & \textbf{RF} \cite{ml2}       & \textbf{99.920} & \textbf{99.913} & \textbf{100.0}  & \textbf{99.953} & \textbf{25.5}  \\ 
\cline{2-7}
                                  & \textbf{LightGBM} \cite{lighta1} & \textbf{99.984} & \textbf{99.985} & \textbf{99.998} & \textbf{99.991} & \textbf{2.5}   \\ 
\hline
\multirow{5}{*}{\shortstack{AutoDP \& \\AutoFE}} & NB                & 93.628          & 93.629          & 99.998          & 96.709          & 0.2            \\ 
\cline{2-7}
                                  & MLP                & 95.009          & 95.099          & 98.943          & 97.781          & 113.6          \\ 
\cline{2-7}
                                  & KNN                & 97.865          & 98.534          & 99.196          & 98.864          & 44.3           \\ 
\cline{2-7}
                                  & \textbf{RF}       & \textbf{99.976} & \textbf{99.973} & \textbf{100.0}  & \textbf{99.989} & \textbf{16.3}  \\ 
\cline{2-7}
                                  & \textbf{LightGBM} & \textbf{99.986} & \textbf{99.986} & \textbf{99.998} & \textbf{99.992} & \textbf{1.2}   \\ 
\hline
\multirow{2}{*}{All}              & RF               & 99.984          & 99.980          & 100.0           & 99.991          & 29.5           \\ 
\cline{2-7}
                                  & \textbf{LightGBM}  & \textbf{99.992} & \textbf{99.993} & \textbf{99.998} & \textbf{99.996} & \textbf{2.3}   \\

\Xhline{1.2pt}

\end{tabular}
\label{t9}%
\end{table*}

Table \ref{t8} summarizes the experimental results from the first series of experiments on offline learning using the CICIDS2017 dataset. For original ML models without using AutoML, five ML models (NB, MLP, KNN, RF, and LightGBM) demonstrate largely different performances. The F1-scores of NB and MLP models are at a low level (47.896\% and 58.830\%), because they are simple models and under-fitting on the complex CICIDS2017 dataset, as discussed in Table \ref{t1}. RF and LightGBM models achieve high F1-scores of 99.248\% and 99.525\% due to their strong capacity to process complex and imbalanced datasets. After implementing the proposed AutoDP and AutoFE procedures, the RF and LightGBM models are still the two best-performing models. Their F1-scores are more than 6\% higher than those of the other three compared ML models. Compared with the ML models without AutoDP and AutoFE, the F1-scores of the RF and LightGBM models have improved from 99.248\% to 99.294\% and from 99.525\% to 99.598\%, respectively. This is because the data quality has been improved by using SMOTE to balance the dataset and using FS methods to remove noisy features. Additionally, the learning time for RF and LightGBM has been reduced from 17.2s to 13.3s and from 1.4s to 0.9s, respectively. This is because the number of features of the CICIDS2017 dataset has been reduced from 80 to 19 after implementing the AutoFE technique. Furthermore, after implementing the HPO procedure to optimize the RF and LightGBM models to complete the entire AutoML pipeline, the performance of the learning models has been further improved, and the optimal LightGBM model with the highest F1-score of 99.653\% is returned as the final model. 

Similarly, as shown in Table \ref{t9}, the RF and LightGBM models outperform the other three compared ML models on the IoTID20 dataset due to their ability to analyze complex and imbalanced IoT anomaly detection data. Their F1-scores have slightly improved from 99.953\% to 99.989\% and 99.991\% to 99.992\% after implementing AutoDP and AutoFE, as the SMOTE method has been implemented to balance the dataset and the FS methods have been implemented to remove irrelevant and noisy features. As the number of features has been reduced from 83 to 31, the learning time has also been reduced for each ML model. After conducting the HPO procedures on the two best-performing models, RF and LightGBM, the optimized LightGBM model achieves the highest F1-score of 99.996\% on the IoTID20 dataset and is selected as the final optimal model. 

To summarize, implementing the AutoML pipeline can obtain a better offline learning model with 0.128\% and 0.005\% F1-score improvement as well as 28.6\% and 8.0\% reduction in learning time, when compared to the best-performing learning model obtained without AutoML on the CICIDS2017 and IoTID20 datasets, respectively.

\begin{table*} [htbp]
\caption{The experimental results of online learning on the CICIDS2017 dataset using prequential evaluation.}
\setlength\extrarowheight{1pt}
\centering
\scriptsize
\begin{tabular}{p{1.5cm}|p{1.7cm}|p{1.5cm}|p{1.5cm}|p{1.4cm}|p{1.4cm}|p{1.4cm}}
\Xhline{1.2pt}
\textbf{AutoML Procedures}        & \textbf{Learning Algorithm} & \textbf{Accuracy (\%)} & \textbf{Precision (\%)} & \textbf{Recall (\%)} & \textbf{F1 (\%)} & \textbf{Model Learning Time (s)}  \\ 
\hline
\multirow{5}{*}{No}               & Offline LightGBM \cite{lighta1}                         & 88.033               & 90.538                & 36.794              & 50.510           & 7.4                              \\

\cline{2-7}             & HT \cite{ht1}                         & 88.676                 & 95.266                  & 43.451               & 59.681           & 17.1                              \\ 
\cline{2-7}
                                & EFDT \cite{efdt}                       & 95.132                 & 86.438                  & 88.674               & 87.541           & 17.4                              \\ 
\cline{2-7}
                                & \textbf{ARF} \cite{arf}               & \textbf{96.228}        & \textbf{92.784}         & \textbf{87.228}      & \textbf{89.920}  & \textbf{30.8}                     \\ 
\cline{2-7}
                                & \textbf{SRP} \cite{srp}               & \textbf{95.772}        & \textbf{92.204}         & \textbf{85.292}      & \textbf{88.614}  & \textbf{231.7}                    \\ 
\hline
\multirow{5}{*}{\shortstack{AutoDP \& \\AutoFE}}               & Offline LightGBM                       & 85.023               & 76.967                 & 18.323               & 29.599           & 6.2                            \\

\cline{2-7} & HT                          & 88.496                 & 76.749                  & 57.894               & 66.001           & 7.7                               \\ 
\cline{2-7}
                                & EFDT                        & 94.181                 & 84.880                  & 84.966               & 84.923           & 7.9                               \\ 
\cline{2-7}
                                & \textbf{ARF}              & \textbf{94.912}        & \textbf{89.045}         & \textbf{83.948}      & \textbf{86.421}  & \textbf{26.6}                     \\ 
\cline{2-7}
                                & \textbf{SRP}               & \textbf{94.547}        & \textbf{90.314}         & \textbf{80.342}      & \textbf{85.037}  & \textbf{101.8}                    \\ 
\hline
\multirow{2}{*}{All}            & ARF                        & 98.593                 & 96.259                  & 96.455               & 96.357           & 29.3                              \\ 
\cline{2-7}
                                & \textbf{SRP}              & \textbf{98.990}        & \textbf{97.801}         & \textbf{96.944}      & \textbf{97.371}  & \textbf{139.4}                    \\

\Xhline{1.2pt}

\end{tabular}
\label{t10}%
\end{table*}

\begin{table*} [htbp]
\caption{The experimental results of online learning on the IoTID20 dataset using prequential evaluation.}
\setlength\extrarowheight{1pt}
\centering
\scriptsize
\begin{tabular}{p{1.5cm}|p{1.7cm}|p{1.5cm}|p{1.5cm}|p{1.4cm}|p{1.4cm}|p{1.4cm}}
\Xhline{1.2pt}
\textbf{AutoML Procedures}        & \textbf{Learning Algorithm} & \textbf{Accuracy (\%)} & \textbf{Precision (\%)} & \textbf{Recall (\%)} & \textbf{F1 (\%)} & \textbf{Model Learning Time (s)}  \\ 
\hline
\multirow{5}{*}{No}              & Offline LightGBM \cite{lighta1}                         & 98.500               & 98.793                 & 99.618               & 99.204           & 11.6                              \\

\cline{2-7} & HT \cite{ht1}                         & 98.220                 & 98.981                  & 99.120               & 99.051           & 37.2                              \\ \cline{2-7} 

                                  & \textbf{EFDT} \cite{efdt}              & \textbf{99.478}        & \textbf{99.571}         & \textbf{99.873}      & \textbf{99.722}  & \textbf{37.3}                     \\ 
\cline{2-7}
                                  & ARF \cite{arf}                        & 98.195                 & 98.434                  & 99.659               & 99.043           & 65.3                              \\ 
\cline{2-7}
                                  & \textbf{SRP} \cite{srp}               & \textbf{99.208}        & \textbf{99.367}         & \textbf{99.790}      & \textbf{99.578}  & \textbf{492.1}                    \\ 
\hline
\multirow{5}{*}{\shortstack{AutoDP \& \\AutoFE}} 
 & Offline LightGBM                        & 99.037          & 99.005                 & 99.983               & 99.492           & 9.4                              \\
\cline{2-7} & HT                         & 98.922                 & 99.092                  & 99.763               & 99.426           & 12.1                              \\ 
\cline{2-7}
                                  & EFDT                     & 99.280                 & 99.359                  & 99.877               & 99.617           & 12.2                              \\ 
\cline{2-7}
                                  & \textbf{ARF}             & \textbf{99.501}        & \textbf{99.541}         & \textbf{99.928}      & \textbf{99.734}  & \textbf{51.1}                     \\ 
\cline{2-7}
                                  & \textbf{SRP}               & \textbf{99.494}        & \textbf{99.539}         & \textbf{99.922}      & \textbf{99.730}  & \textbf{159.2}                    \\ 
\hline
\multirow{2}{*}{All}              & ARF                        & 99.664                 & 99.658                  & 99.985               & 99.821           & 58.5                              \\ 
\cline{2-7}
                                  & \textbf{SRP}               & \textbf{99.705}        & \textbf{99.726}         & \textbf{99.960}      & \textbf{99.843}  & \textbf{199.7}                    \\

\Xhline{1.2pt}

\end{tabular}
\label{t11}%
\end{table*}

In the second series of experiments for online learning, the results on the CICIDS2017 and IoTID20 datasets are shown in Tables \ref{t10} and \ref{t11}, respectively. To justify the necessity of online adaptive learning, the best-performing static ML model in the offline learning experiments, offline LightGBM, is also evaluated for comparison purposes. As shown in Table \ref{t10}, the offline LightGBM model shows the worst F1-scores of 50.510\% among the five evaluated models (offline LightGBM, HT, EFDT, ARF, and SRP) on the CICIDS2017 dataset. This is because many new or zero-day attacks were launched in the creation process of the CICIDS2017 dataset, causing several concept drift points. As offline LightGBM cannot adapt to concept drift and can only detect existing types of attacks, its performance has been gradually degrading over time. On the other hand, the four online learning methods (HT, EFDT, ARF, and SRP), especially ARF and SRP, can adapt to concept drift and maintain high performance. Among the four online learning algorithms,  ARF and SRP are the two best-performing learning models, with F1-scores of 86.421\% and 85.037\% after implementing the AutoDP and AutoFE procedures. This is because they are strong online ensemble models with high drift adaptability, as discussed in Section 7.4.3. The two incremental learning methods, HT and EFDT, are not as robust as ensemble models due to their relatively low model complexity. 

Compared with the learning models without using AutoDP and AutoFE, although F1-scores of the learning models with AutoDP and AutoFE are slightly lower, their model learning time has largely reduced from 30.8s to 26.6s and from 231.7s to 101.8s, respectively. The F1-scores are slightly lower because the number of features of the CICIDS2017 dataset has been largely reduced from 80 to 19 after implementing the AutoFE method, which eliminates many less important features. As online learning usually starts with a small number of samples, the features have a more significant impact on online learning performance than offline learning. Although removing less important features may ignore certain data patterns and slightly reduce the learning performance, it can largely reduce the model learning time as the complexity and size of the dataset have been significantly reduced. To achieve real-time online learning in IoT data analytics, it is crucial to take learning speed into account. Moreover, after implementing the HPO procedure to complete the entire AutoML pipeline, the F1-scores of the ARF and SRP models have significantly improved to 96.357\% and 97.371\%, respectively. Thus, the proposed AutoML pipeline can return the optimal SRP model with the highest F1-score of 97.371\%. This justifies that the overall AutoML procedures can still improve both model learning effectiveness and efficiency. 

For the IoTID20 dataset, as shown in Table \ref{t11}, the ARF and SRP models have also achieved higher F1-scores than the other two learning models after executing the AutoDP and AutoFE procedures, of 99.734\% and 99.730\%, respectively, although the best-performing original learning model without AutoML is EFDT (99.722\%). This shows that different online learning models will perform differently in specific IoT data analytics tasks. Additionally, the offline LightGBM still achieves a relatively low F1-score of 99.204\% on the IoTID20 dataset, which is much lower than the F1-scores of most online learning methods. 

After implementing AutoDP and AutoFE, the F1-scores of ARF and SRP have increased by 0.691\% and 0.152\%, while their model learning time has reduced by 21.8\% and 67.7\%, respectively. This is because the dataset remains balanced during the learning process after implementing the SMOTE method, and the feature size has been largely reduced from 83 to 31 after implementing the IG and Pearson-based FS methods. Finally, after implementing the HPO procedure, the optimal SRP model with the highest F1-score of 99.843\% can be returned. Therefore, implementing the AutoML pipeline can obtain a better online learning model with 8.757\% and 0.265\% F1-score improvement as well as 39.8\% and 59.4\% learning time reduction than the same learning model obtained without AutoML on the CICIDS2017 and IoTID20 datasets, respectively.

In conclusion, the proposed AutoML pipeline enables us to obtain an optimal learning model with high effectiveness and efficiency for IoT anomaly detection tasks in both offline and online environments using the IoTID20 and CICIDS2017 datasets. Furthermore, the experimental results in Tables \ref{t8} and \ref{t11} have justified the following assumptions and theoretical analysis in previous Sections:
\begin{enumerate}
\item Different ML methods have different performances in specific tasks, as demonstrated by the performance of the five offline ML models and four online learning models in the experimental results. This supports the necessity of selecting appropriate models.
\item Hyperparameter tuning and optimization have a direct impact on the model performance, as shown in the third set of results in each of Tables \ref{t8} - \ref{t11}. The performance of learning models has been improved significantly by implementing the HPO method (“All” versus “AutoDP \& AutoFE”). This supports the necessity of hyperparameter optimization or automated model tuning.
\item Data pre-processing and feature engineering methods affect learning performance. The performance of most learning models has been improved significantly by implementing AutoDP and AutoFE (“AutoDP \& AutoFE” versus “No”). This supports the necessity of AutoDP and AutoFE.
\item Concept drift issues will cause model performance degradation, but automated model updating and concept drift adaptation methods can address model performance degradation. As shown in Tables \ref{t10} and \ref{t11}, the best-performing offline models (offline LightGBM) still perform the worst when compared to other online adaptive learning models due to the occurrence of concept drift. This supports the necessity of automated model updating and concept drift adaptation.
\end{enumerate}

\section{Open Challenges and Research Directions}
To effectively apply AutoML methods to IoT streaming data analytics problems, many challenges need to be addressed. In this Section, we discuss the open challenges and research directions in this domain. These challenges are classified into three major categories: IoT data analytics challenges, AutoML application challenges, and concept drift challenges, as summarized in Table \ref{t12}. 

\begin{table*}[htbp]
\caption{The challenges and research directions of applying AutoML to IoT data analytics}
\setlength\extrarowheight{1pt}
\centering
\scriptsize
\begin{tabular}{p{2cm}|p{3cm}|p{7cm}}
\Xhline{1.2pt}
\textbf{Category}                                        & \textbf{Challenge} & \textbf{Brief Description}                                                                                                                                             \\ \Xhline{1.2pt}

\multirow{3}{*}{\shortstack{IoT Data\\ Analytics\\ Challenges}} & IoT Data Quality                              & Data quality
  has a direct effect on data
  analytics performance. However, IoT data is often collected from different
  data sources, it is often challenging to ensure data quality.                                    \\ 
\cline{2-3}
                                               & IoT Data Privacy                              & The
  collection process of IoT data streams often faces privacy issues, as they
  are often from different IoT devices/systems. Federated
  Learning (FL) techniques can be used to protect data privacy.                 \\ 
\cline{2-3}
                                               & IoT Data Analytics Speed                      & IoT data
  analytics models often require fast processing speed to achieve real-time
  processing in IoT systems.                                                                                                          \\ 
\hline
\multirow{6}{*}{\shortstack{AutoML\\ Challenges} }            & Automated Model Updating                      & The
  automated model updating process is often ignored
  in many AutoML systems. This step is important in real-world IoT data
  analytics applications, as IoT data is often dynamic streaming data.                     \\ 
\cline{2-3}
                                               & Data Pre-Processing and Feature
  Engineering & Most AutoML
  systems only focus on automated model selection and HPO
  procedures. Thus, data pre-processing and feature engineering need more attention,
  as they also have a significant effect on model performance.  \\ 
\cline{2-3}
                                               & Large Scale AutoML                            & It is
  challenging to apply AutoML models on large-scale data, such as ImageNet, as
  the learning models often need to be trained many times to identify the
  optimal solution.                                         \\ 
\cline{2-3}
                                               & Explainability                                & AutoML
  solutions are often black boxes, so their explainability needs more research.                                                                                                                                     \\ 
\cline{2-3}
                                               & Transfer Learning in AutoML                   & High
  complexity is a common issue in AutoML systems. Transfer learning techniques
  can be used to save model learning time.                                                                                             \\ 
\cline{2-3}
                                               & Benchmarking and Comparability                & A benchmark
  should be agreed upon by the community for a fair comparison of different
  AutoML techniques.                                                                                                               \\ 
\hline
\multirow{4}{*}{\shortstack{Concept Drift\\ Challenges} }     & Unsupervised Learning                         & More
  research should be conducted on unsupervised or semi-supervised
  drift detection and adaptation, as most existing methods are developed for
  supervised learning.                                                 \\ 
\cline{2-3}
                                               & Drift Analysis                                & A
  comprehensive analysis should be conducted on the detected drifts, such as
  the timing and severity of each drift.                                                                                                    \\ 
\cline{2-3}
                                               & ML Model Integration                          & Appropriate
  drift methods should be selected and integrated with specific
  ML algorithms to develop effective automated drift adaptation functionality.                                                                 \\ 
\cline{2-3}
                                               & Accurate Drift Detection                      & Drift detection methods should be capable of accurately detecting different types of drift (e.g., abrupt and gradual drift.)                                                                                      \\
\Xhline{1.2pt}

\end{tabular}
\label{t12}%
\end{table*}

\subsection{IoT Data Analytics Challenges}
Although data analytics contributes significantly to IoT applications, it is still in its early stages \cite{iot4}. Numerous challenges must be addressed before IoT data can be properly used in IoT applications \cite{iotdata}, such as the quality, privacy, and analytics speed of IoT data.
\subsubsection{IoT Data Quality}
Firstly, the quality of data has a direct effect on the performance of data analytics models. Thus, it is critical to have high-quality data \cite{iotdata}. However, as IoT data is often collected from different data sources and is highly variable, maintaining data quality is usually challenging. The high generation speed and volume of IoT data are also significant problems \cite{iot4}. Effective data integration has also become a challenge for creating high-quality datasets from different IoT devices.

\subsubsection{IoT Data Analytics Speed}
Due to the massive amount of data generated by IoT devices, time constraints are a primary challenge of IoT data analytics. Many IoT applications have real-time requirements, such as autonomous vehicles and e-health systems \cite{rl1}. In these applications, real-time feedback on environmental changes is required. However, many factors, like transmission delays and model learning time, have increased the reaction time of IoT systems. Therefore, it is essential for data analytics methods to achieve real-time analytics on large amounts of IoT data generated at high speeds \cite{iotdata}. Real-time analytics allows IoT devices to make real-time decisions and provide services. As described in Section 2.2, edge computing and collaborative computing techniques are promising solutions to achieve real-time analytics, but better architecture should be designed to balance data analytics efficiency and accuracy.

On the other hand, to avoid unfeasible IoT data analytics due to time or memory constraints, distributed ML is a promising solution that allocates the learning process over multiple workstations \cite{disml}. Critical methods of distributed ML include data parallelism, model parallelism, task parallelism, and hybrid parallelism \cite{disml2}.

In data parallelism, the training dataset is partitioned into multiple subsets and then distributed to multiple computing entities. On the computing entities, each subset is trained by the same model in parallel. Data parallelism can adapt to increasing volumes of training data, but it is difficult to handle complex ML models (e.g., DL models with large numbers of layers) due to their high memory footprints and transmission delays. To handle complex ML models that are difficult to be loaded into single computing entities, model parallelism is another distributed ML method that splits a ML model into multiple parts and then places them in different computing entities. Model parallelism can address the memory constraints of computing devices and improve data processing speed. Task parallelism is the process of executing computing tasks or programs on different processors on the same or multiple computing devices. For example, multiple threads can be created for a specific task to enable parallel execution, and each thread is in charge of carrying out a different action. Apache Storm is a popular task parallelism framework for big data analytics \cite{disml2}. Hybrid parallelism refers to the combination of different parallelism techniques. For example, data and model parallelism methods can be implemented simultaneously to save both execution time and memory. Task parallelism can also be combined with data parallelism to make use of both multi-threading and subsets. 

Unlike traditional data processing systems that collect data from multiple sources for central processing, in distributed ML, every IoT end device or edge server can store and process its own data in itself or only share data with trusted devices to avoid the leakage of private and sensitive data. The above distributed ML methods enable each computing device to process data locally without leakage. Distributed ML can also improve data analytics speed by enabling parallel execution and reducing data transmission time. This is because the computational time of analyzing a large central database is much higher than dividing it into multiple sub-tasks that analyze data in parallel \cite{disml}. Thus, both data security and processing efficiency can be improved by using distributed ML techniques.

\subsubsection{IoT Data Security and Privacy}
With the advent of data analytics techniques for IoT data, data security has emerged as a critical concern \cite{iot4}. As a comprehensive IoT dataset is often generated from different data sources, certain personal or sensitive business data may be derived during the data collection process \cite{iotdata}. Thus, it is crucial for IoT systems to solve data privacy issues.

Cybersecurity mechanisms, like data encryption and device authentication, can improve the privacy of IoT data. However, these techniques introduce additional overhead to IoT systems. On the other hand, distributed ML and Federated Learning (FL) techniques can be a potential solution for IoT data privacy \cite{fldm}. By distributed ML approaches, the data is stored only in local IoT devices to make data unavailable to unauthorized devices, thus maintaining data security and privacy. FL techniques can train ML models without direct access to local data by exchanging model parameter values between edge and central servers \cite{fldm2}. Thus, by the employment of FL approaches, the privacy of IoT data can be protected without compromising learning performance. 

\subsubsection{IoT Data Analytics Application Benchmarks}

\subsection{AutoML Challenges}
AutoML has made considerable strides in the previous decade in automating model construction and development, especially for supervised learning tasks. However, to be widely applied to real-world IoT applications, AutoML still faces many challenges \cite{challenge1}.
\subsubsection{Automated Model Updating}
Despite the development of AutoML, most AutoML solutions are offline models designed for static datasets. However, many real-world applications, like IoT systems, face concept drift issues throughout the data analytics process, and most existing AutoML solutions only update models using new data samples \cite{oasw}. Therefore, this paper considers automated model updates by addressing concept drift issues in the AutoML pipeline, which is a novel contribution to AutoML applications. Considering automated model updates can help construct robust AutoML models that maintain effectiveness over time. 
\subsubsection{Data Pre-Processing and Feature Engineering}
Although there are many existing AutoML solutions, the majority of them focus on automated model selection and HPO. Researchers have paid little attention to automated data pre-processing and feature engineering \cite{challenge1}. However, data pre-processing and feature engineering are critical components of the AutoML pipeline and have a direct influence on system performance. It is often challenging to generalize and automate the feature engineering process because it is very task- and dataset-dependent \cite{gra3}. Appropriate feature engineering often requires specialized domain knowledge or a significant amount of effort. Therefore, automated feature engineering is a critical but challenging subject that needs further research.  
\subsubsection{Large Scale AutoML}
Applying AutoML to large-scale data is still an unsolved issue. Due to the fact that AutoML pipelines often need a significant number of model trainings to identify the optimum final learner, the majority of AutoML solutions are developed on small datasets, with just a few capable of large-scale data learning. For instance, research on AutoML solutions for the ImageNet problem is currently rather limited, owing to the dataset's massive size \cite{hpome}. 
\subsubsection{Explainability}
In general, AutoML solutions are black boxes that attempt to explore the space of possible models and discover the optimal solution. Despite AutoML's advances, the community has not explored the prospect of transparent AutoML systems. AutoML models should have mechanisms for explaining and understanding them, since this would considerably improve AutoML's accessibility. On the other hand, data visualization techniques can also be considered in HITL-based applications. Through effective data visualization, humans can better interpret and analyze intermediate data analytics results to further enhance prediction performance.

\subsubsection{Transfer Learning in AutoML}
Due to the high complexity of existing AutoML solutions in terms of time and space, transfer learning methods can be utilized to increase AutoML's efficiency. This is because transfer learning enables the reduction of unnecessary model retrainings via the usage of existing models. While meta-learning processes are a subtype of transfer learning, transfer learning can also refer to the transfer of knowledge about the optimization process (e.g., transferring information on the dynamics of the optimization process from task to task) \cite{cnnme}. 
\subsubsection{Benchmarking and Comparability}
As various AutoML systems have distinct benefits and drawbacks in different IoT applications, the community should agree on a set of common benchmarks that allow a fair comparison of different techniques \cite{hpome}. Similarly, code sharing and processes that facilitate the replication of AutoML discoveries may have a substantial impact on the field's maturity.

\section{Conclusion}
Machine Learning (ML) and Deep Learning (DL) algorithms have achieved great success in data analytics tasks for IoT applications, such as intelligent transportation systems, smart homes, e-health, and IoT security. However, developing effective ML models for specific IoT tasks requires a high level of human expertise, which limits their applicability. Thus, Automated ML (AutoML) has become a promising solution for constructing ML models without or with minimal human intervention. In this paper, we have comprehensively discussed the procedures of the standard AutoML pipeline, including automated data pre-processing, automated feature engineering, automated model selection, Hyper-Parameter Optimization (HPO), and automated model updating with concept drift adaptation. Moreover, we have explored the IoT data analytics tasks, as well as the ML and DL models that are often employed in IoT data analytics. Existing tools and libraries for implementing AutoML and IoT data analytics are also presented in this paper. Additionally, a case study of IoT anomaly detection is conducted in this work to demonstrate the procedures of AutoML applications. Experimental results have shown the benefits of using AutoML frameworks in IoT data analytics problems. Finally, we discuss the open challenges and research directions related to the existing AutoML and IoT data analytics tasks. Although this paper provides a comprehensive discussion and a case study of applying AutoML technology to IoT data analytics, due to time and resource constraints, a more comprehensive and fair benchmark of AutoML techniques should be developed for the community in the future. Additionally, the automation of data pre-processing and feature engineering still has much room for improvement, so advanced automation techniques should be proposed to better automate these two ML procedures. In future work, we will explore the application of AutoML in other areas like sentiment analysis and conduct research on other techniques to further improve IoT data analytics performance, like distributed ML.

\newpage
\begin{wrapfigure}{l}{30mm} 
\includegraphics[width=1.25in,height=1.5in,clip,keepaspectratio]{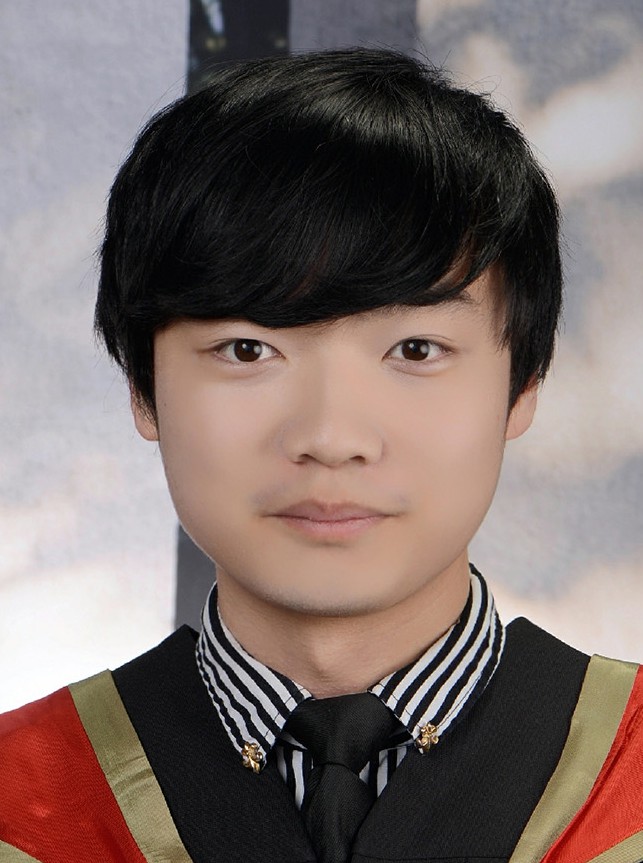}
\end{wrapfigure}\par
\textbf{Li Yang} received his Ph.D. in Electrical and Computer Engineering from Western University, London, Canada, in August 2022, his MASc. degree in Engineering from the University of Guelph, Guelph, Canada, 2018, and his B.E. degree in Computer Science from Wuhan University of Science and Technology, Wuhan, China, in 2016. Currently, he is a Postdoctoral Associate in the Optimized Computing and Communications (OC2) Lab at Western University.  His research interests include cybersecurity, machine learning, AutoML, deep learning, network data analytics, Internet of Things (IoT), anomaly detection, online learning, concept drift, and time series data analytics.\par
  
~\\

\begin{wrapfigure}{l}{30mm} 
\includegraphics[width=1.25in,height=1.45in,clip,keepaspectratio]{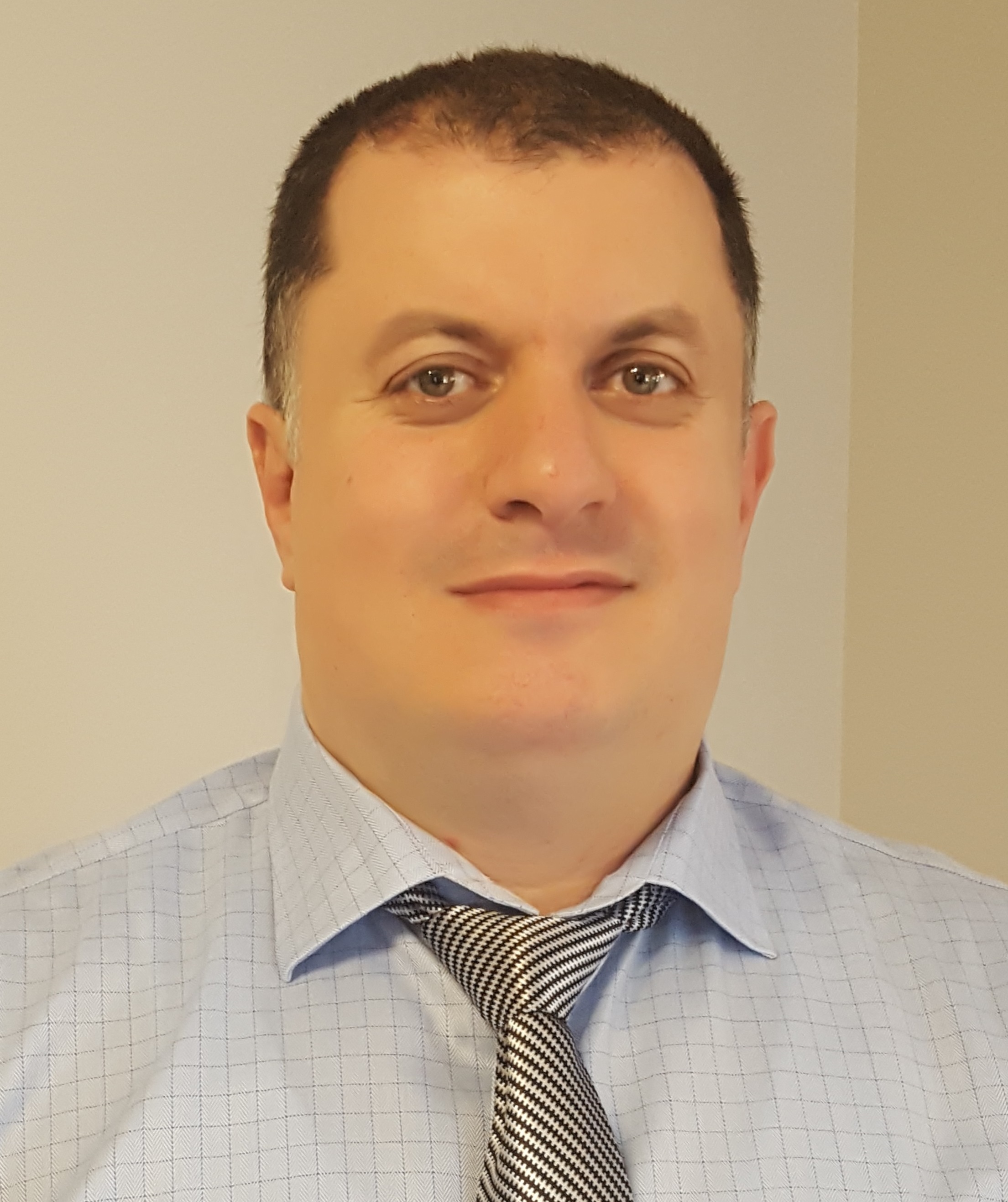}
\end{wrapfigure}\par
\textbf{Abdallah Shami} is currently a Professor in the Electrical and
Computer Engineering Department and the Acting Associate Dean (Research) of the Faculty of Engineering, Western University, London, ON, Canada, where he is also the Director of the Optimized Computing and Communications Laboratory. Dr. Shami has chaired key symposia for the IEEE GLOBECOM, IEEE International Conference on Communications, and IEEE International Conference on Computing, Networking and Communications. He was the elected Chair for the IEEE Communications Society Technical Committee on Communications Software from 2016 to 2017 and the IEEE London Ontario Section Chair from 2016 to 2018. He is currently an Associate Editor of the IEEE Transactions on Mobile Computing, IEEE Network, and IEEE Communications Surveys and Tutorials.\par

\end{document}